\documentclass[smallcondensed,final]{article}     
\usepackage[utf8]{inputenc} 

\usepackage{authblk} 

\usepackage{graphicx}
\usepackage{amsmath, amssymb, amsfonts, bm}
\usepackage{nicefrac}       	
\usepackage{bbm}                
\usepackage{ulem}

\usepackage{url}            
\usepackage{microtype}      
\usepackage{lipsum}

\usepackage[ruled, vlined]{algorithm2e}
\usepackage{algorithmic}
\usepackage{xcolor} 
\usepackage{color}
\usepackage{mathtools}
\usepackage[colorlinks,linkcolor=blue,citecolor=blue, anchorcolor = blue]{hyperref}
\usepackage{natbib}
\usepackage[labelfont=bf, labelsep=space]{subcaption}
\usepackage[labelfont=bf, labelsep=space]{caption}

\usepackage{booktabs}   
\usepackage{multirow} 
\usepackage{scalefnt}
\usepackage{colortbl}   
\usepackage{diagbox}

\usepackage{appendix}

\usepackage[left=2.9cm,right=2.9cm,top=2.5cm,bottom=3.5cm]{geometry}
\begin{document}

\title{Time Series Analysis and Modeling to Forecast: a Survey
}

\makeatletter
\newcommand\email[2][]%
   {\newaffiltrue\let\AB@blk@and\AB@pand
      \if\relax#1\relax\def\AB@note{\AB@thenote}\else\def\AB@note{\relax}%
        \setcounter{Maxaffil}{0}\fi
      \begingroup
        \let\protect\@unexpandable@protect
        \def\thanks{\protect\thanks}\def\footnote{\protect\footnote}%
        \@temptokena=\expandafter{\AB@authors}%
        {\def\\{\protect\\\protect\Affilfont}\xdef\AB@temp{#2}}%
         \xdef\AB@authors{\the\@temptokena\AB@las\AB@au@str
         \protect\\[\affilsep]\protect\Affilfont\AB@temp}%
         \gdef\AB@las{}\gdef\AB@au@str{}%
        {\def\\{, \ignorespaces}\xdef\AB@temp{#2}}%
        \@temptokena=\expandafter{\AB@affillist}%
        \xdef\AB@affillist{\the\@temptokena \AB@affilsep
          \AB@affilnote{}\protect\Affilfont\AB@temp}%
      \endgroup
       \let\AB@affilsep\AB@affilsepx
}
\makeatother

\author[$\dag$]{Fatoumata Dama}
\author[$\dag$]{Christine Sinoquet}
\affil[$\dag$]{LS2N / UMR CNRS 6004, Nantes University, France}
\email{\url{{fatoumata.dama,christine.sinoquet}@univ-nantes.fr}} 

\date{}

\maketitle

\begin{abstract}
Time series modeling for predictive purpose has been an active research area of machine learning for many years. However, no sufficiently comprehensive and meanwhile substantive survey was offered so far. This survey strives to meet this need. A unified presentation has been adopted for entire parts of this compilation. 
 
A red thread guides the reader from time series preprocessing to forecasting. Time series decomposition is a major preprocessing task, to separate nonstationary effects (the deterministic components) from the remaining stochastic constituent, assumed to be stationary. The deterministic components are predictable and contribute to the
prediction through estimations or extrapolation. Fitting the most appropriate model to the remaining stochastic component aims at capturing the relationship between past and future values, to allow prediction. 

We cover a sufficiently broad spectrum of models while nonetheless offering substantial methodological developments. We describe three major linear parametric models, together with two nonlinear extensions, and present five categories of nonlinear parametric models. Beyond conventional statistical models, we highlight six categories of deep neural networks appropriate for time series forecasting in nonlinear framework. 

Finally, we enlighten new avenues of research for time series modeling and forecasting. 
We also report software made publicly available for the models presented. 
\end{abstract}


{\bf Keywords} time series modeling, forecasting, time series decomposition, autocorrelation, stationarity, linearity, nonlinearity

\section{Introduction}

Time series data are amongst the most ubiquitous data types that capture information and record activity in most aeras. In any domain involving temporal measurements {\it via} sensors, censuses, transaction records, the capture of a sequence of observations indexed by time stamps first allows to provide insights on the past evolution of some measurable quantity. Beyond this goal, the pervasiveness of time series has generated an increasing demand for performing various tasks on time series data (visualization, discovery  of recurrent  patterns, correlation discovery, classification, clustering, outlier detection, segmentation, forecasting, data simulation).

Temporal {\it visualization} is one of the simplest and quickest ways to represent time series data. More importantly, graphical analysis is a crucial step of time series analysis, whatever the downstream data processing line. Notably, it should be the first step before starting with time series modeling.

In online recommendation systems, {\it correlation mining} aims at identifying Internet users sharing similar shopping patterns, for instance. The final objective is to provide dynamic recommendations based on the correlation between a given customer and other customers' behaviors. In the stock market sector, identifying correlations amongst stock prices can lead stockbrokers to spot investment opportunities. 

Spatio-temporal {\it classification} may open up new possibilities, such as management of land use and of its evolution through geographical information systems. However, time series classification requires the collection of reference data: in the absence of label information, this deep retrospective analysis represents a tough task generally hampered by the complexity of thematic classes and their lack of formal description in time series. Unsupervised {\it clustering} is often used, which offers a solution based upon the data alone. 

{\it Outlier detection} represents another prominent field of time series analysis. Outlier detection is central to understand the normal behavior of data streams and detect situations that deviate from the norm. 

The aim of time series {\it segmentation} is to identify the boundary points of segments in the data flow, to characterize the dynamical properties associated with each segment. 

{\it Forecasting} in time series is an important area of machine learning. Time series analysis helps in analyzing the past, which comes in handy to forecast the future. Forecasting is of prime importance in many domains: in finance and business, to plan policies by various organizations; in industrial quality process control and digital transactions, to detect abnormal or fraudulent situations; in electric power distribution, to control loads {\it via} advanced monitoring and cope with disturbances in power flows; in meteorology, to guide informed decision-making for agriculture, air and maritime navigation; in biology, to gain knowledge on the activity of genes given affected/unaffected status; in medicine, to predict the spread of a disease, estimate mortality rates or assess time-dependent risk. 

Classification in time series has motivated the writing of several surveys \citep{susto_cenedese_terzi_2018_book-chapter_time-series-classif-review-power-systems-appli,fawaz_forestier_weber_et_al_2019_dmkd_time-series-classif-deep-learning-review-experim,abanda_mori_lozano_2019_dmkd_review-dist-based-time-series-classif_without_experim,ruiz_flynn_large_et_al_2021_dmkd_survey-multivar-time-series-classif-experim}. To our knowledge, the most recent review addressing clustering in stream data dates back to 2015 \citep{aghabozorgi_shirkhorshidi_wah_2015_inform-syst_time-series-cluster-review-without-exper}. The recent book by \citet{maharaj_d-urso_caiado_2019_book_time-series-cluster-and-classif} is dedicated to both classification and clustering in time series. \citet{ali_alqahtani_jones_et_al_2019_ieee-access_time-series-cluster-classif-for-visual-analytics-review} review both classification and clustering methods devoted to visual analytics in time series. Two recent state-of-the-art reviews focus on anomaly detection \citep{braei_wagner_2020_arxiv_time-series-anomaly-detect-review,blazquez-garcia_conde_mori_et_al_2021_acm-comput-surv_review-outlier-detec-time-series}. As regards forecasting in time series, we identified a few recent substantial surveys, which are moreover all entirely devoted to deep learning \citep{lim_zohren_2020_arxiv_time-series-forecasting-deep-learning-review,sezer_gudelek_ozbayoglu_2020_applied-soft-comput_time-series-forecasting-finance-review,torres_hadjout_sebaa_et_al_2021_big-data_deep-learn-time-series-foresc-review,lara-benitez_carranza-garcia_riquelme_2021_int-j-neural-syst_experi-review-deep-learn-time-series-forec}. Even less could we identify a survey meeting the objectives sought through the writing of the present document. Drafting this survey was motivated by three goals:\newline

\noindent{\bf $\bullet$ A compilation bringing an informed choice for modeling and forecasting purposes}
\newline
Multiple media encompassing journal articles, tutorials and blog posts allow to catch valuable though fragmented information on prediction in time series. With a focus on operationality (code source examples are generally provided) and illustration, the two latter media can only tackle specific aspects of time series (for instance, introduction to the fundamentals of time series data and analysis: \citealp{aptech_2020_blog_intro-times-series-analysis}; how to decompose time series data into trend and seasonality: \citealp{brownlee_2020_blog-machine-learning-mastery_time-series-decomposition-trend-seasonality}; using Python and Auto {\sc arima} to forecast seasonal time series: \citealp{portilla_2018_blog_python-autoarima-forecast-seasonality}; Holt-Winters forecasting simplified: \citealp{solarwinds_2019_blog_holt-winters-forecasting}). On the other hand, several books and e-books dealing with time series were written. In this category, one finds large textbooks that offer a generalist approach for time series analysis and forecasting \citep[{\it e.g.},][]{montgomery_jennings_kulahci_2016_book_intro-time-series-analy-forecast}. Other books and e-books target hands-on approaches \citep[{\it e.g.},][]{nielsen_2019_book_practical-time-series-analys}. Finally, several writings are entirely dedicated to a few methods or models \citep[{\it e.g.},][]{brownlee_2020_blog-machine-learning-mastery_time-series-forecast-deep-learning-mlp-cnn-lstm-python}, to a specific application field \citep[{\it e.g.},][]{brooks_2019_book_intro-time-series-finance} or to one language \citep[{\it e.g.},][]{avishek_prakash_2017_ebook_practical-time-series-analys}.

However, the exponential increase in the volume of data, including time series data in all sectors of industry, finance, economy, biology, medicine, together with the growing availability of software programs, announces in particular the appropriation of time series prediction tools by scientists that are nonspecialists of time series. In complement to these media targetting "how to" aspects, and to the aforecited books and e-books, our aim is to provide readers a quick access to sufficiently detailed information, over a large methodological spectrum. In particular, we wished to offer a compilation allowing nonspecialists to step back and make an informed choice to process their stream data for modeling and forecasting purposes.\newline

\noindent{\bf $\bullet$ A sufficiently broad panorama covered by substantial developments}
\newline
Time series modeling, stationarity analysis and forecasting ability are intimately related. Numerous publications target cutting-edge theoretical developments on stationarity. Other publications are entirely dedicated to the description of some specific variant of a linear or nonlinear framework, to model time series data. Currently, insights on linear and nonlinear time series modeling are generally presented separately in books \citep{paolella_2018_book_linear-time-series-analysis,tsay_chen_2018_book_nonlinear-time-series-analysis}. In addition, no recent survey could be identified that substantially develops a sufficiently broad panorama on time series modeling upstream forecasting. Moreover, {\it a compilation accessible to scientists with no sharp expertise on time series was lacking, that would put in perspective stream data preprocessing, various facets of time series modeling and forecasting}. In a word, one of the goals of this survey is to provide a self-contained explication of the state-of-the-art we cover here.\newline

\noindent{\bf $\bullet$ A dictactical and unified presentation}
\newline
Finally, strong didactical concerns have motivated the design of this survey. Besides the guideline from time series preprocessing to forecasting, a unified presentation has been adopted as far as possible for entire parts of this compilation.\newline 

A time series is a sequence of data points (observations) ordered in  time. Time is supposed discrete here. A time series represents the temporal evolution of a dynamic system that one searches to describe, explain and predict. Correlation analysis allows to model how a time series is related to its delayed values. This analysis is crucial to design a forecasting procedure, a major aim of time series data processing. Forecasting is based on the following principle: knowing the past behaviors of a system, it is possible to make previsions on its nearby or long-term behaviors.

Let $\mathbf{x} = x_1, x_2, ..., x_T$ be a time series describing a dynamic system of interest; $\mathbf{x}$ is a realization of the stochastic process $\mathbf{X} = \{X_1, X_2, ...\} = \{X_t\}_{t=1}^{\infty}$. Prediction requires the modeling of the relationship between past and future values of the system
\begin{equation}
X_t = f(t, X_{t-1}, X_{t-2}, ...) + g(t, X_{t-1}, X_{t-2},...)\epsilon_t,
\label{time.series.model.specification}
\end{equation}{}
where $\{\epsilon_t\}$ is a series of noises such that the $\epsilon_t$'s are independent and identically distributed (i.i.d) with $0$ mean and unit variance, and are also independent from the past values of $X_t$, and where $f$ and $g$ are respectively the conditional mean and variance of $X_t$ (given the past values). The conditional mean and variance are possibly time-dependent. The key to prediction in time series is the ability to model the dependencies between current and lagged values (defined as autocorrelation).

\textbf{In what follows $h$ designates $f$ or $g$.} To model $h$, the approaches described in the literature break down into two main categories: parametric and nonparametric models.

Nonparametric models suppose that $h$ belongs to some flexible class of functions. For instance, the review by \citet*{hardle_lutkepohl_chen_1997_int_stat_review_review-nonparametric-time-series} put forward approaches relying on functions that belong to the $C_1$ class (that is functions that are differentiable over some interval $I$, and whose derivative function $f\prime$ is continuous on $I$). In contrast, parametric models specify $h$ within a class of parametric functions, such as the class of polynomial functions \citep{tong_1990_book_non-linear-time_series,hamilton_1994_book_time-series}. A nonparametric model can be seen as a parametric framework with a high number of parameters to instantiate, thus requiring a large volume of data to learn the model. Flexibility and interpretability are respective advantages of nonparametric and parametric models. Nonetheless, flexibility may be increased in parametric models. Notably, this survey will show that it is possible to increase flexibility by enabling more complex relationships between the value at time step $t$ and its lagged values. To note, it is possible to use a parametric model for $f$ and a nonparametric model for $g$, and {\it vice-versa}.

This survey aims at presenting a state-of-the-art of parametric models dedicated to time series analysis, with the purpose of prediction. These models are divided into two categories: \textbf{linear} and \textbf{nonlinear} models. In this survey, we will decrypt in particular the links between linearity and stationarity, and under which conditions they hold for some of the linear models presented. We will also see that stationarity and nonlinearity may be compatible.
 
This survey is organized as follows. Section 2 deals with the stationarity concept and reviews the main classes of methods to test for weak stationarity. The main frameworks for times series decomposition are presented in Section 3 in a unified way. Section 4 is devoted to the unified presentation of three popular linear models used for time series modeling, each time detailing autocovariance, parameter learning algorithms and forecasting.  This section ends with the description of more flexible models, to escape the limits of the previous models. A step further, but still in a unified manner, Section 5 depicts five major nonlinear models used for time series. Amongst nonlinear models, artificial neural networks hold a place apart, as deep learning has recently gained considerable attention. Section 6 therefore brings methodological insights in time series forecasting with deep learning. This section describes a selection of five models. Section 7 briefly overviews time series model evaluation in two aspects, model diagnosis and forecast performance evaluation. Section 8 provides a list of R and Python implementations for the methods, models and tests presented throughout this review. Finally, Section 9 identifies research prospects in the field of time series forecasting. A glossary is provided in Section 10. 

To recapitulate, we highlight the following contributions:
\begin{itemize}
 \item{To the best of our knowledge, this is the first comprehensive survey dedicated to forecasting in time series, with the concern to take a comprehensive view of the full process flow, from decomposition to forecasting {\it via} stationary tests, modeling and model evaluation.}

 \item{This survey offers, as far as possible, a unified presentation of decomposition frameworks, on the one hand, and of linear and nonlinear time series models on the other hand.}

 \item{The relationships between stationarity and linearity are decrypted, with the aim of bringing them within the reach of scientists that are nonspecialists of time series.}

 \item{In this document, our intention is to bring sufficient in-depth knowledge, while covering a broad range of models and forecasting methods: this compilation spans from well-established conventional approaches to more recent adaptations of deep learning to time series forecasting.}

 \item{This compilation opens up in the enlightment of new paths for future works around time series modeling.}
\end{itemize}

\section{Stationarity}

This section will first recall the definitions of strong and weak stationarity. Then it will briefly describe the main categories of methods employed to test for stationarity. Finally, a recapitulation of methods falling in these categories will be provided. 

\subsection{Stationary Stochastic Process}
\label{stationarity.def}

A stochastic process $\mathbf{X}$ is \textit{stationary} if its statistical properties are time-independent. A distinction is made between \textit{strong} stationarity and \textit{weak} stationarity.

\subsubsection{Strong Stationarity}
$\mathbf{X}$ is a strongly stationary process if its distribution satisfies the following property:


\begin{equation}
\label{strict.stationarity}
    \mathbb{P}(X_1, X_2,..., X_{T}) = \mathbb{P}(X_{1+\tau}, X_{2+\tau},..., X_{T+\tau}), \;\ \forall T, \tau  \in \mathbb{N}^*.
\end{equation}
Stationarity means that the distribution of $\mathbf{X}$ is the same over time or, in other words, that it is invariant to any time shift.
In particular, any {\it i.i.d} stochastic process is strongly stationary.
In practice, it is difficult to test property (\ref{strict.stationarity}). Thus, a weak version, named \textit{weak stationarity}, was introduced.

\subsubsection{Weak Stationarity}
\label{weak.stationarity.def}

$\mathbf{X}$ is a weakly stationary (or mean-covariance stationary) process if its mean and covariance verify the following properties:
\begin{equation}
\label{weak.stationarity}
\begin{split}
    \mathbb{E}[X_t] &= \mu \quad \text{(mean stationarity)},\\
    \text{Cov}(X_t, X_{t+h}) &= \gamma(h) \implies \text{Var}(X_t) = \gamma(0) < \infty \;\;\;\ \text{(covariance stationarity)},
\end{split}
\end{equation}
where $\text{Cov}$ and $\text{Var}$ respectively denote covariance and variance, and $\gamma$ is symmetric ($\gamma(-h) = \gamma(h)$) and bounded ($|\gamma(h)| \le \gamma(0) < \infty$). Formula (\ref{weak.stationarity}) expresses that the mean and autocovariance of $\mathbf{X}$ are time-independent and that its variance is finite.

\begin{figure}[t]
	\begin{subfigure}[b]{0.5\textwidth}
        \includegraphics[width=\textwidth]{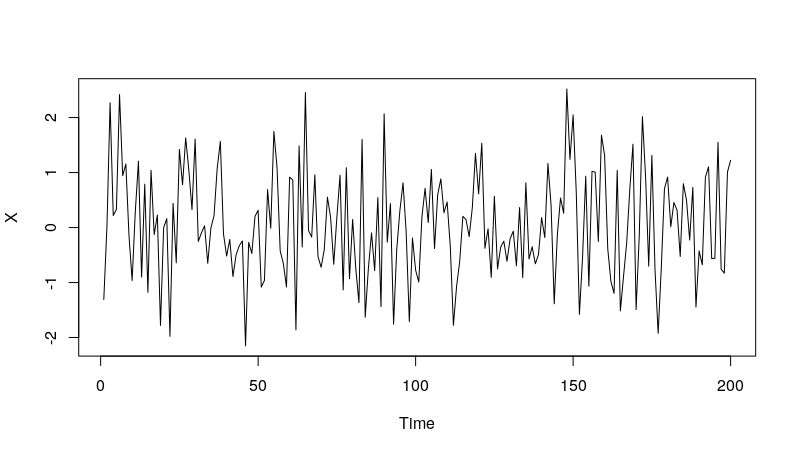}
        \caption{Weak stationarity}
        \label{stationary.white.noise}
    \end{subfigure}
    \begin{subfigure}[b]{0.5\textwidth}
        \includegraphics[width=\textwidth]{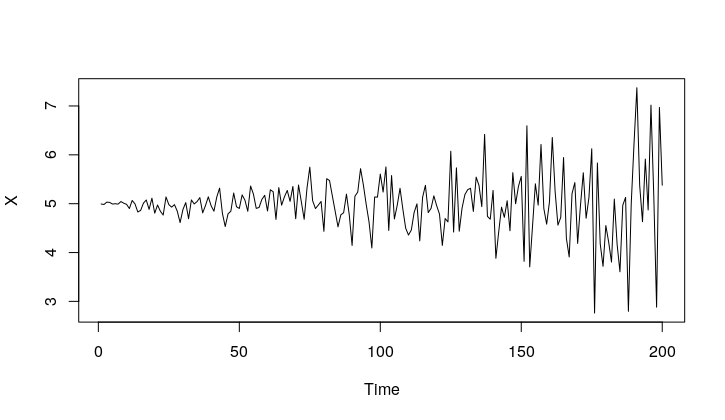}
        \caption{Mean stationarity and variance nonstationarity}
        \label{weak_stationarity.def}
    \end{subfigure}
    \begin{subfigure}[b]{0.5\textwidth}
        \includegraphics[width=\textwidth]{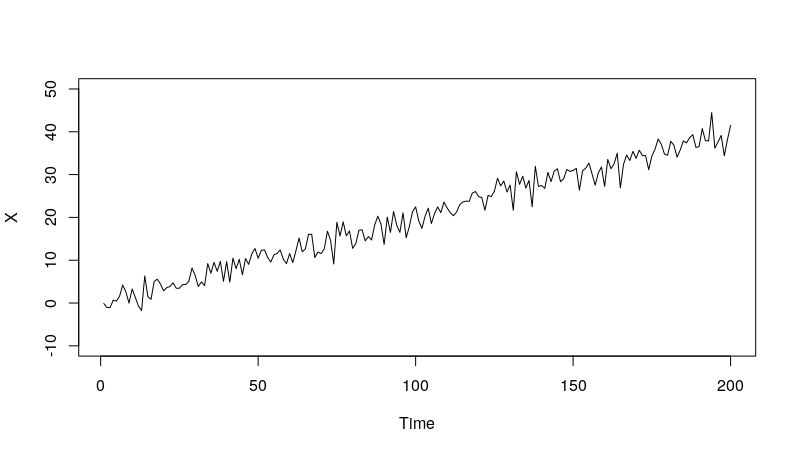}
        \caption{Mean nonstationarity and variance stationarity}
        \label{}
    \end{subfigure}
    \begin{subfigure}[b]{0.5\textwidth}
        \includegraphics[width=\textwidth]{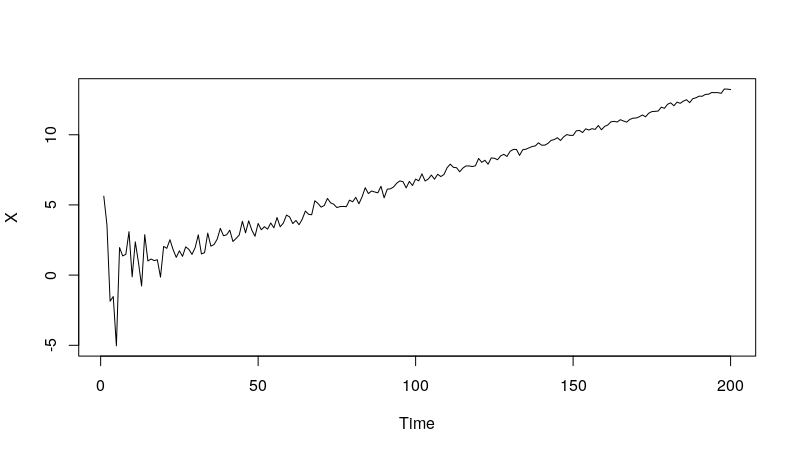}
        \caption{Mean and variance nonstationarities}
        \label{}
    \end{subfigure}
	\caption[Square plots]{Illustration of stationarity and nonstationarity on simulated time series. $\mathcal{N}(\mu,\sigma)$ represents the \textit{normal} law with mean $\mu$ and variance $\sigma^2$. 
	{\bf (a)} Weak stationarity: uncorrelated white noise simulated through $X_t \sim \mathcal{N}(0, 1)$; $\mathbb{E}[X_t] = 0$; $\text{Var}[X_t] = 1$. 
	{\bf (b)} Mean stationarity: $X_t \sim \mathcal{N}(5, \sigma_t)$ with $\sigma_t = \frac{t+1}{200}$, $\mathbb{E}[X_t] = 5$ and $\text{Var}[X_t] = \sigma_t^2$ increases with time. 
	{\bf (c)} Variance stationarity: $X_t = t/5 + \mathcal{N}(0, 2)$, $\mathbb{E}[X_t] = t/5$ increases with time and $\text{Var}(X_t) = 4$. 
	{\bf (d)} Mean and variance nonstationarities: $X_t = t/15 + \mathcal{N}(0, \sigma_t)$ with $\sigma_t = \frac{20}{t}$, $\mathbb{E}[X_t] = t/15$ and $\text{Var}[X_t] = \sigma_t^2$. Mean increases with time whereas variance decreases.}
\label{time.series.stationarity.type.examples}
\end{figure}

Weak stationarity relaxes the strong stationarity property by only constraining the first moment and autocovariance function of $\mathbf{X}$ instead of its whole distribution. Thus, \textit{strong stationarity} implies \textit{weak stationarity}. Figure \ref{time.series.stationarity.type.examples} shows an example for \textit{weak stationarity}, \textit{mean stationarity}, \textit{variance stationarity} and \textit{mean and variance nonstationarity}. Weak stationarity is also called \textit{second-order stationarity}. \textit{First-order stationarity} refers to the case where only the mean is constant.

From now on, the term "weak" will be omitted for the sake of simplicity, that is \textbf{weak stationarity will be referred to as stationarity}. 

\subsection{How to Test for Stationarity?}
\label{stationarity.test}
Testing the stationarity for a time series is of great importance for its modeling.
To assess the stationarity of a stochastic process $\mathbf{X}$ based on a realization $\{x_t\}_{t=1}^{T}$, we can use either \textit{graphical methods} or \textit{statistical tests}.

\subsubsection{Graphical Methods}
\label{stationary_test_graphical_methods}

\noindent{\bf $\bullet$ Analysis of time series plot}
\newline
Stationarity can be visually assessed from time series plots on which prominent seasonality, trend and changes in variances are investigated. For instance, the time series of atmospheric concentration of $CO_2$ in Hawaii (Figure \ref{co2.concentration.hawaii}) has an increasing trend and a yearly seasonal pattern.

\noindent{\bf $\bullet$ Plotting rolling statistics}
\newline    
\textit{Rolling statistics} graphs plot means and variances computed within sliding time windows. Such graphs can provide insights on stationarity. Indeed, if these graphs exhibit time-invariance, then the studied time series will probably be stationary. 

\begin{figure}[t]
    \begin{subfigure}[b]{0.5\textwidth}
        \includegraphics[width=\textwidth]{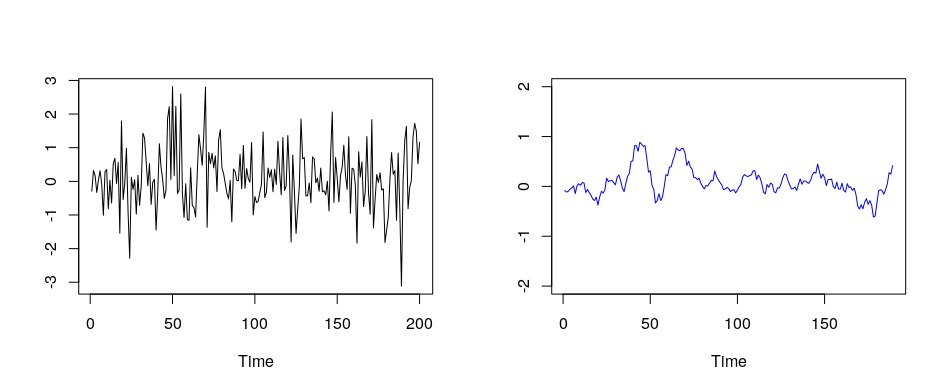}
        \caption{Stationary time series.}
        \label{rolling.mean.12.white.noise}
    \end{subfigure}
    \begin{subfigure}[b]{0.5\textwidth}
        \includegraphics[width=\textwidth]{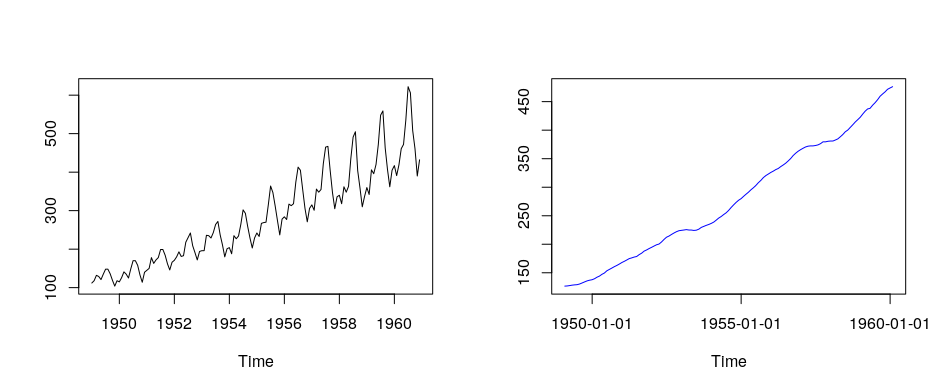}
        \caption{Nonstationary time series.}
        \label{rolling.mean.12.usa.airline.numbers.of.passenger}
	\end{subfigure}
\caption{Rolling mean method applied to examples of stationary and nonstationary time series. Window size equals $12$. 
{\bf (a)} White noise process $\mathcal{N}(0,1)$ where $\mathcal{N}$ stands for Gaussian law (left), rolling mean graph (right). {\bf (b)} USA monthly airline numbers of passengers (left), rolling mean graph (right). 
}
\label{rolling.mean}
\end{figure}

Figure \ref{rolling.mean} displays an application of the rolling mean method to a stationary time series and a nonstationary time series respectively. Note that the rolling graph is time-invariant for the (stationary) white noise process (Figure \ref{rolling.mean.12.white.noise}) while it shows an increase for the nonstationary time series (Figure \ref{rolling.mean.12.usa.airline.numbers.of.passenger}).

\noindent{\bf $\bullet$ Correlogram {\it versus} covariogram}
\newline
In time series analysis, a \textit{correlogram} and a \textit{covariogram} are respectively the graphs of the autocorrelation function ({\sc acf}) and of the scaled autocovariance function. These functions are defined as 

\begin{equation}
\label{autocorrelation_formula}
\begin{split}
    \rho(h) &= \frac{\text{Cov}(X_t, X_{t+h})}{\sqrt{\text{Var}(X_t) \text{Var}(X_{t+h})}} \;\;\ (\text{autocorrelation function}),
\end{split}
\end{equation}{}
\begin{equation}
\label{scaled_autocovariance_formula}
\begin{split}
    \rho_s(h) &= \frac{\text{Cov}(X_t, X_{t+h})}{\text{Var}(X_t)} \;\;\ (\text{scaled autocovariance}).
\end{split}
\end{equation}{}

\begin{figure}[t]
    \begin{subfigure}[b]{0.5\textwidth}
        \includegraphics[width=\textwidth]{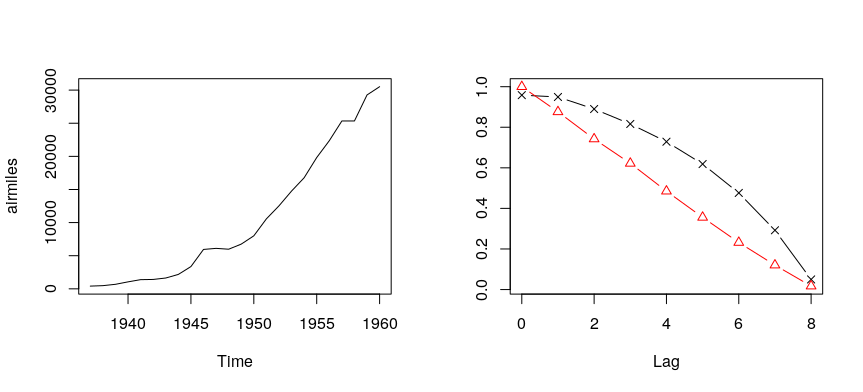}
        \caption{Revenue passenger miles flown by commercial airlines in the United States.}
        \label{corr.cov.airmiles}
    \end{subfigure}
    \begin{subfigure}[b]{0.5\textwidth}
        \includegraphics[width=\textwidth]{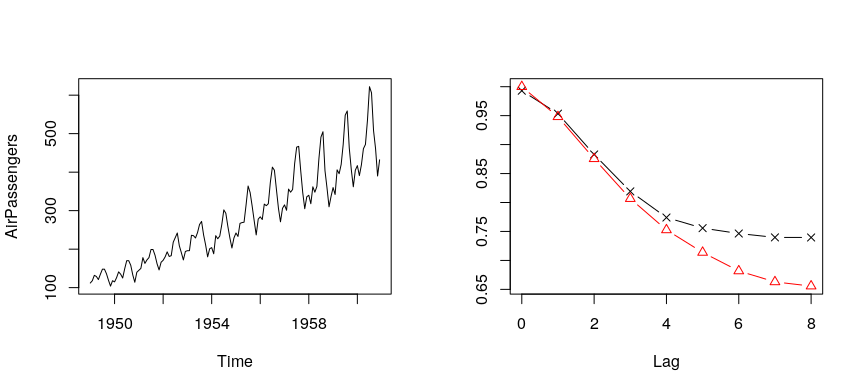}
        \caption{Monthly airline numbers of passengers in the United States.}
        \label{corr.cov.air.passengers}
	\end{subfigure}
\caption{Correlogram (cross) \textit{vs} covariogram (triangle) obtained for two nonstationary time series.}
\label{correlogram.vs.covariogram}
\end{figure}

Correlograms and covariograms allow to graphically describe temporal dependencies existing within the observations. A correlogram (respectively a covariogram) plots the value of the {\sc acf} (respectively the scaled autocovariance) for increasing lags. 
It has to be highlighted that correlograms and covariograms are identical for stationary time series (since $\text{Var}(X_t) = \text{Var}(X_{t+h})$).
Besides, it was proven that correlograms tend to better discriminate between stationarity and nonstationarity than covariograms \citep{nielsen_2006_journ-royal-stat-soc_correlograms-vs-covariograms}.
In particular, correlograms show a quick decrease down to zero for stationary processes, unlike nonstationary ones for which the decay is slower. Figure \ref{correlogram.vs.covariogram} illustrates this fact. 

However, visual investigations of stationarity do not always allow to conclude. In this case, more rigorous \textit{statistical tests} have to be used.

\subsubsection{Statistical Tests}
\noindent{\bf $\bullet$ A short reminder about statistical tests}
\newline
A statistical test is used to decide between two hypotheses, the null hypothesis $H_0$ and the alternative hypothesis $H_1$, where $H_0$ is the hypothesis considered true if no evidence is brought against it. The risk to wrongly accept $H_1$, $\mathbb{P}_{H_0}(H_1) = \alpha$, is referred to as \textit{type I error}. Symmetrically, the risk $\mathbb{P}_{H_1}(H_0) = \beta$ is called \textit{type II error}. $1-\alpha$ and $1 - \beta$ are respectively called \textbf{test confidence} and \textbf{test power}. $\alpha$ and $\beta$ were proven to vary inversely.

Formally, a statistical test relies on a \textit{test statistic}, $S$. The \textit{observed statistic} $S_{obs}$ is a numerical summary of the information contained in the observed dataset, whereas $S$ is the random variable summarizing the information for any dataset in the whole distribution of possible datasets. Provided we can compute or approximate the law of $S$ under $H_0$, $\mathbb{P}_{H_0}(S)$, we can evaluate the probability to obtain a \textit{test statistic} more atypical than $S_{obs}$ in the direction of the alternative. This probability is called the $p\text{-}value$. For example, in a right (respectively left) uni-lateral test, $H_0$ is rejected when $S_{obs}$ is too high (respectively too low), that is if $p\text{-}value = \mathbb{P}_{H_0}(S > S_{obs}) \le \theta $ (respectively $p\text{-}value = \mathbb{P}_{H_0}(S < S_{obs}) \le \theta$), with some user-defined threshold $\theta$. $H_0$ rejection entails $H_1$ acceptance. Finally, the higher the difference $|p\text{-}value - \theta|$, the safer the decision.
Moreover, the lower the threshold $\theta$, the higher the test confidence.  

\noindent{\bf $\bullet$ Stationarity tests}
\newline
When relying on statistical tests to analyze time series, we must pay attention to the choice of the hypotheses. If we test $H_0$ (nonstationarity) against $H_1$ (stationarity), the lower the $\alpha$ type I error (high confidence), the safer it is to conclude to stationarity. In contrast, if we test $H_0$ (stationarity) against $H_1$ (nonstationarity), the lower the $\beta$ type II error (high power), the safer it is to decide stationarity.
The statistical tests proposed to test whether a stochastic time series is stationary or not can roughly be grouped into two categories: tests that mostly work in the frequency domain and those that work in the time domain.

\noindent{\bf $\bullet$ Frequency domain methods} 
\newline
A natural way to analyze a signal (for instance a time series) consists in plotting its behavior against time. However, this temporal representation does not offer insight on the frequencies contained in the signal. For a brief reminder, just consider the simple cosine signal $x_t = A \cos(2 \pi f t + \phi)$, where amplitude $A$ determines the maximum absolute height of the signal, frequency $f$ controls how rapidly the signal oscillates and $\phi$ is the phase at origine. This simple signal contains only one frequency, but more complex functions such as times series can be viewed as a sum of cosine waves with varying amplitudes, frequencies and phases at origine, for example $X_t = 3 \cos(\frac{\pi}{4} t) - 2 \cos(\frac{\pi}{3} t + \frac{\pi}{2}) + \cos(\frac{\pi}{2} t + \frac{\pi}{6})$. The spectrum of a signal plots the amplitudes of the various components existing in the signal against the frequencies. Moreover, it is recalled that the Fourier series expansion of a signal writes as a sum of trigonometric function of all frequencies, for instance $A_i \sin(\omega i t + \phi_i)$, $i = 1, 2, ...$, where $\omega = 2 \pi f$. Short-time Fourier transform uses a sliding window, which provides information of both time and frequency. To increase the resolution in frequency, Wavelet transform is another standard function based on small wavelets with limited duration.

Several methods testing for (non)stationarity rely on some spectral decomposition of the time series under analysis. Therein, we find approaches testing the constancy of spectral characteristics \citep{priestley_subba_1969_journ-royal-stat-soc_test_non-statio-time-series,paparoditis_2010_journ-americ-stat-assoc_statio-time-series-periodograms}, methods comparing a time-varying spectral density estimate with its stationary approximation \citep{dette_preuss_vetter_2011_journ-americ-stat-assoc_locally-stationary-processes,preuss_vetter_dette_2013_bernoulli_statio-test-empiric-process,puchstein_preuss_2016_journ-time-series-analys_multivar-locally-statio-proces} and approaches based on Fourier transform \citep{basu_rudoy_wolfe_2009_ieee-int_conf-acou-speech-signal-proces_nonparam-test-statio-fourier,dwivedi_subba_2011_journ-time-series-analys_second-order-statio-fourier} or on wavelets \citep{nason_2013_journ-royal-stat-soc_second-order-locally-statio-time-series,cardinali_nason_2013_journ-stat-softwa_locally-statio-time-series}.
Recently, a class of dynamic processes exhibiting a functional time-varying spectral representation was established
\citep{van-delft_eichler_2018_electro_j_stat_local-statio-spectr-repres}. In particular, this class includes time-varying {\it functional} {\sc arma} processes which have been shown to be functional locally stationary. {\sc arma} processes will be described in Subsection \ref{arma}. Functional-coefficient autoregressive processes will be presented in Subsection \ref{far}.

Frequency analysis is dedicated to discover underlying periodicities in time series. This periodic behavior can be very complex. Frequency domain analysis is useful if the time series under study is known to remain invariant within a certain time interval. Issues arise if the frequency changes over time.

\noindent{\bf $\bullet$ Time domain methods}
\newline
Unit root tests represent standard methods to identify nonstationarity. To convey the intuition behind this concept, we  introduce unit roots in the case of a linear autoregressive process {\sc ar}(p): $X_t = \phi_0 + \sum_{i=1}^{p}\ \phi_i\ X_{t-i}+ \sigma \epsilon_t$ where the $\epsilon_i$'s are usually uncorrelated white noise processes. The reformulation as $(1 - \phi_1 L - \phi_2 L^2 - ... - \phi_p L^P) X_t = \phi_0 + \sigma \epsilon_t$ where $L$ represents the lag operator ($L X_t = X_{t-1},\ t > 1$) allows to exhibit the so-called characteristic equation of ${X_t}$: $1 - \phi_1 y - \phi_2 y^2 - ... - \phi_p y^P = 0$. An {\sc ar}(p) process has a unit root if one of the (complex) roots of the characteristic equation has its modulus equal to 1. Unit roots are one cause for nonstationarity. An {\sc ar}(p) process with a unit root can be transformed into a stationary process through a given number of differentiation transformations (see equation \ref{differentiation}, Subsection \ref{additive_trend}). An {\sc ar}(p) process with unit root and multiplicity $m$ for this root exactly requires $m$ successive differentiations to become stationary. Such a process is said to be integrated to order $m$.

Standard unit root tests, such as \textit{Phillip-Perron test} \citep{phillips_perron_1988_biometrika_test-unit-root-time-series-regres} and \textit{Dickey-Fuller test} \citep{dickey_fuller_1979_journ-americ-stat-assoc_-ar-times-series-unit-root}, check stationarity under one-order autocorrelation assumption and constant trend assumption for the former, linear trend for the latter. 
In the {\sc kpss} test, the time series 
under study is written as the sum of a linear trend, a random walk process RW (defined as $RW_t = RW_{t-1} + u_t$ where the $u_t$'s are i.i.d($0,\sigma^2$)) and a standard error \citep{kwiatkowski_phillips_schmidt_et_al_1992_journ-econometr_unit-root-test}. \textit{Augmented Dickey-Fuller} ({\sc adf}) test checks the stationarity around a linear trend under a $p$-order autocorrelation \citep{cheung_lai_1995_journ-busin-economi-stat_augment-dickey-fuller-test, dolado_gonzalo_mayoral_2002_econometrica_dickey-fuller-test-unit-roots,elliott_rothenberg_stock_1992_econometr_ar-unit-root-test}. 
The works developed by \citet{phillips_1987_econometrica_unit-root-test-random-walk-and-arima-model} extend unit root-based tests to the random walk and more general {\sc arima} models (to be presented in Subsection \ref{arima}).

	Several methods were proposed to deal with time series in which break points are known or suspected \citep[{\it e.g.},][]{perron_1989_econometrica_unique-known-break-point-unit-root-test}. The \textit{Zivot and Andrews test} tests the null hypothesis of unit root existence against the stationary alternative with a unique structural break point \citep{zivot_andrews_2002_journ-busin-economic-stat_one-struct-break-unit-root-test}. 

	Other methods are able to cope with more than one break point. The {\sc cusum} test is widely used to test for parameter changes in time series. The method described by \citet{lee_ha_na_et_al_2003_journ-scand-stat_cusum-param-change-time-series} implements a {\sc cusum} test, to handle the break point problem for parameters other than the mean and variance. 

	In the same category, the piece-wise locally stationary ({\sc pls}) time series model described by \citet{zhou_2013_journ-americ-stat-assoc_heterosced-autocor-struct-change} allows both abrupt and smooth changes in the temporal dynamics of the process. In this framework, a bootstrap procedure is used to approximate the null distribution for a {\sc cusum} test statistic ($H_0$: the time series is {\sc pls}). The proposed method goes beyond testing changes in mean. It can be extended to testing structural stability for multidimensional parameters, in second and higher-order nonstationary time series. Further in this line, another {\sc cusum} approach relying on bootstrapping was designed to relax assumptions \citep{dette_wu_zhou_2015_arxiv_change-point-second-order-time-series}: for instance, when testing for the stability of autocovariance, it is not required that mean and variance be constant under $H_0$; moreover, this approach allows break points at different times, for variance and autocovariance.

Table \ref{recapitulation.stationarity.tests} sums up the (non)stationarity tests that perform in time domain analysis and that were presented in this section. 

\begin{table}[ht]
\scalefont{0.75}
\centering
\begin{tabular}{|l|l|l|l|l|l|}
\hline
 \multirow{2}{*}{\textbf{Method}} & \multirow{2}{*}{\textbf{Trend $T_t$}} & \multirow{2}{*}{\textbf{Test}}  & \multirow{2}{*}{\textbf{Hypotheses}} & \multirow{2}{*}{\textbf{$H_0$}} &\multirow{2}{*}{\textbf{Reference}} \\
                    &        &                &        &             & \\
\hline
                & Constant & Phillips-Perron & \multirow{2}{*}{$X_t = T_t + AR(1)$} & \multirow{2}{*}{{\sc ns}} & \cite{phillips_perron_1988_biometrika_test-unit-root-time-series-regres}\\
\cline{2-3} \cline{6-6}
                & \multirow{5}{*}{Linear} & Dickey-Fuller  & &  & \cite{dickey_fuller_1979_journ-americ-stat-assoc_-ar-times-series-unit-root}\\
 \cline{3-6}
                &           & {\sc kpss}      & $X_t = T_t + RW_t + \epsilon_t$ & {\sc s} &\cite{kwiatkowski_phillips_schmidt_et_al_1992_journ-econometr_unit-root-test}\\
\cline{3-6}
 Unit root     &           &   \multirow{3}{*}{{\sc adf}}       &  \multirow{3}{*}{$X_t = T_t + AR(p)$} &  \multirow{3}{*}{{\sc ns}} & \cite{cheung_lai_1995_journ-busin-economi-stat_augment-dickey-fuller-test}\\
                &           &         &                     &    & \cite{dolado_gonzalo_mayoral_2002_econometrica_dickey-fuller-test-unit-roots}\\
                &           &         &                     &    & \cite{elliott_rothenberg_stock_1992_econometr_ar-unit-root-test}\\
\cline{2-3} \cline{6-6}
\cline{2-6}
		        &  Constant  & Phillips       & $X_t = T_t + RW_t + \epsilon_t $ or $ X_t = T_t + ARIMA$ & {\sc ns} & \cite{phillips_1987_econometrica_unit-root-test-random-walk-and-arima-model} \\
\hline
Unit root  & \multirow{2}{*}{Linear} &  Perron       & unique known breakpoint in trend & \multirow{2}{*}{{\sc ns}} & \cite{perron_1989_econometrica_unique-known-break-point-unit-root-test}\\
and break point &  &  Zivot and Andrews & unique unknown breakpoint in trend & & \cite{zivot_andrews_2002_journ-busin-economic-stat_one-struct-break-unit-root-test}\\
\hline
            & \multirow{4}{*}{Constant}   &  \multirow{2}{*}{Lee}    & random coefficient autoregressive model   & \multirow{4}{*}{{\sc s}} & \multirow{2}{*}{\cite{lee_ha_na_et_al_2003_journ-scand-stat_cusum-param-change-time-series}}\\
Break point &  &        & infinite-order  moving  average  process & &\\
\cline{3-4}\cline{6-6}
analysis    &  &  Zhou  &                          &   &\cite{zhou_2013_journ-americ-stat-assoc_heterosced-autocor-struct-change}\\
      &  &  Dette \textit{et al}  &           &   & \cite{dette_wu_zhou_2015_arxiv_change-point-second-order-time-series} \\      
\hline
\end{tabular}
\caption{Stationary tests that perform in time domain. {\sc adf}: augmented Dickey-Fuller test. {\sc s}: stationary. {\sc ns}: nonstationary. {\sc rw}: random walk process. {\sc arima}: autoregressive integrated moving average. The {\sc arima} model is a generalization of the autoregressive ({\sc ar}) model. These models are presented in Subsections \ref{arima} and \ref{lar}, respectively. {\sc ls}: locally stationary. In an {\sc ls} time series model, the process is assumed to be smoothly changing over time. {\sc pls}: piecewise locally stationary. In the {\sc pls} time series framework, the time series is divided into several time intervals, and the process is assumed to be stationary in each interval.}
\scalefont{1.0}
\label{recapitulation.stationarity.tests}
\end{table}{}

\section{Time Series Decomposition}
\label{decomposition}

Subsection \ref{objectives.decomposition} will describe the motivation behind the decomposition of nonstationary time series into nonstationary effects and a remaining component. Then, various decomposition schemes will be described through Subsections \ref{non.stationary.mean} to \ref{non.stationary.mean.variance}. Subsection \ref{recapitulation.decomposition} will offer a recapitulation of the most popular models and methods designed for time series decomposition. Further, Subsection \ref{forecasting} will address prediction in the decomposition scheme. 

\subsection{Objectives of Time Series Decomposition}
\label{objectives.decomposition}

The aim of time series decomposition is to decompose a nonstationary time series $\mathbf{X} = \{X_1, X_2, ...\} = \{X_t\}_{t=1}^{\infty}$ into nonstationary effects (the deterministic components) and a remaining component (the stochastic constituent) $\{Z_t\}_{t=1}^{\infty}$, to allow further tasks. Such tasks encompass the characterization of the cause for nonstationarity, together with prediction. The commonly studied nonstationary effects are trend and seasonality, which are taken into account in many time series analysis models such as {\sc arima} \citep{usman_2019_journ-applied-sci-envir-manag_arima-applied-neonatal-mortality,nyoni_2019_mpra_arima}, {\sc sarima} \citep{samal_2019_int-conf-information_techno-compt-sci_sarima-prophet-model,valipour_2015_meteo-appli_sarima-arima,martinez_2011_revista-sociedade-brasileira-medicina-tropical_sarima}, and exponential smoothing-based methods \citep{pegels_1969_manag-sci_exponential-forecasting-new-variations,taylor_2003_int-journ-forecasting_exponent-smooth-damped-multipli-trend,
aryee_essuman_djagbletey_darkwa_2019_journ-biostat-epidem_compar-sarima-holtwinters}.  

In such a decomposition scheme, the deterministic components are predictable and contribute to the prediction task through their estimations or through extrapolation. For example, the evolution of a trend can be modeled and extrapolation will thus contribute to build the predicted values. Similarly, seasonality (that is, periodic fluctuation) can be modeled, to yield the corresponding deterministic contribution to the predicted values. As a complement, attempts to improve the prediction accuracy require an analysis of the remaining stochastic component. Therefore, depending on the time series, a more or less complex decomposition scheme seeks to obtain the remaining stochastic component $\{Z_t\}$. It is important to note that the stochastic component is desired to be structured, due to autocorrelation, to allow for forecasting. In contrast, the presence of noise (or chaos), that is the effect of unknown or unmeasurable factors, is not a favourable situation for forecasting.  

Autocorrelation is the key concept that allows future prediction in time series. Informally, autocorrelation is a measure that reflects the degree of dependence between the values of a time series over successive time intervals. Otherwise stated, a high autocorrelation means that values are highly dependent on their lagged values. Thus, many models attempting to capture autocorrelation have been proposed in the literature. Stationary models are preferred when possible.

Stationarity is a central and desirable property in time series analysis. Intuitively, stationarity can be explained as follows: if the dynamic system represented by the stochastic component $\{Z_t\}$ obtained through decomposition is characterized by a given behavior at time $t$, this behavior will likely reproduce at time $t+1$. In other words, the system behavior is time-invariant. This implies stability for the relation between $Z_t$ and its lagged values. A key property inherent to stationarity is therefore exploited to model the stochastic process: the autocorrelation is invariant with time. The models thus obtained will offer a simplified framework, to predict future values, based on autocorrelation.

The link between stationarity and linearity was established in Wold's theorem (1938) \citep{wold_1954_book_stationary-time-series}. In a nutshell, Wold's theorem states that any stationary stochastic process can be expressed as an infinite weighted sum of uncorrelated white noise errors. In other words, the process may be represented through an {\sc ma}$(\infty)$ model, where {\sc ma}$(\infty)$ represents a (linear) moving average model with an infinite number of parameters (see Subsection \ref{ma}). 

Conversely, a linear model is not necessarily stationary. In Section \ref{linear.time.series.models}, we will see that the moving average ({\sc ma}) model is stationary by definition. In the same section, we will discuss conditions to guarantee stationarity for two other linear models (linear autoregressive: {\sc ar}; autoregressive moving average: {\sc arma}). 

It should also be emphasized here that numerous stochastic processes exist that show stationarity but are nonlinear. The reason lies in Wold's representation not being adapted or realistic: namely, this representation does not allow to fit a limited amount of data with a reasonable number of parameters and to provide reliable predictions. In such stationary processes, the link between $X_t$ and its lagged values is nonlinear; in other cases, $X_t$ is subject to regime changes.

The phenomenon that governs a stationary time series may be linear or nonlinear. However, if a series is nonstationary, the underlying phenomenon is generally nonlinear. For instance, a nonlinear model may be envisaged in the case when a stochastic process switches from regime to regime. Regime switching is an acknowledged cause for nonstationarity. A compromise is nonetheless proposed by some models that describe jumps between locally stationary processes \citep{zhou_2013_journ-americ-stat-assoc_heterosced-autocor-struct-change}. Other situations may require still more complex modeling, such as in financial markets, where periods of low variations are interspersed with periods of high variations. Essentially, wherever there is time-varying variance (or so-called "volatility"), the time series do not conform to a linear model (see for instance the {\sc garch} model, Subsection \ref{garch}). But the border is not so clear-cut in the case of time-varying variance: nonlinearity does not necessarily imply nonstationarity. Several works have established conditions to guarantee the stationarity for specific variants of the {\sc garch} model (see for example \citealp{bollerslev_1986_journ-economet_garch-conditions-for-stationarity,bougerol_picard_1992_journ-of-economet_stationarity-garch-processes-and-some_nonneg_time-series,panorska_mittnik_rachev_1995_appl-math-letter_garch-stationarity-stable-paretian-alpha-stable-conditional-distrib,mohammadi_2017_journ-of-forecasting_alpha-stable-garch-and-arma-garch-m-models}).

Decomposition aims at preprocessing a raw time series to yield a remaining stochastic component. The latter is further modeled as a linear or nonlinear process, if possible. Moreover, stationarity is a desirable situation as it simplifies modeling and prediction tasks. The risk remains that neither linear nor nonlinear process fits the data well, which would translate in poor prediction performance. Section \ref{time.series.model.evaluation} will be devoted to the evaluation of models dedicated to time series.

Whereas the trend and seasonal components are deterministic and time-dependent, the remaining stochastic component (stationary or not, linear or not) is expected to display no apparent trend and seasonal variations and to contain the information about the autocorrelation of observations. In this favourable situation, the aims pursued by time series decomposition are then to describe the autocorrelation structure and to perform accurate prediction based on the autocorrelation inherent to this remaining part. To achieve the second aim, the remaining component is obtained by filtering out the deterministic components from the raw series. In particular, $\mathbf{X}$ is stationary if the remaining stochastic component is stationary while the other components are null; otherwise, it is nonstationary.

Subsection \ref{weak.stationarity.def} described three classes of nonstationarity: mean, variance, mean and variance nonstationarity. The next three Subsections \ref{non.stationary.mean} to \ref{non.stationary.mean.variance} will exhibit the connection between these former descriptions and various decomposition schemes combining trend, seasonality and the remaining stochastic component. From weakest to strongest level of weak nonstationarity, we can enumerate \textit{mean nonstationarity}, \textit{variance nonstationarity} and \textit{mean and variance nonstationarity}. 

\subsection{Additive Decomposition - Nonstationary Mean}
\label{non.stationary.mean}
$\mathbf{X}$ is a nonstationary mean process if its mean varies over time, that is $\mathbb{E}[X_t] = \mu(t)$, while its variance stays constant. This situation is described by an additive decomposition model defined as
\begin{equation}
\label{additive.model}
    X_t = \mu(t)+ Z_t,
\end{equation}
where $Z_t$ is the remaining stochastic component of $\mathbf{X}$. $Z_t$ is hypothesized to be stationary, with mean and variance respectively equal to $0$ and $\sigma^2$.

It has to be noted that if an estimator of $\mu(t)$ is known, that of the stationary component can be found as follows: $\hat{Z}_t = X_t - \hat{\mu}(t)$. Generally, $\mu(t)$ is either a trend $T_t$ or a seasonal component $S_t$ or any combination of both. In what follows, we will present the classical models widely studied in the literature: {\bf additive trend (}$\bm{\mu(t) = T_t}${\bf )}, {\bf additive seasonality (}$\bm{\mu(t) = S_t}${\bf )} and {\bf additive trend and seasonality (}$\bm{\mu(t) = T_t + S_t}${\bf )}. Figure \ref{additive.decomposition} shows examples of real-life time series that can be modeled through additive decomposition.

\subsubsection{Additive Trend Scheme: $\mu(t) = T_t$} 
\label{additive_trend}
The trend $T_t$ represents the long-term evolution of $\mathbf{X}$. The trend can be increasing, decreasing, polynomial, {\it etc}. It can be either estimated then removed from $X_t$, or (much less easily) modeled then removed from $X_t$. Otherwise, $T_t$ can be directly removed from $X_t$ by transformation.

~\\
\noindent $\bullet$ {\bf Estimation of} $\bm{T_t}$ {\bf in the additive trend scheme}\\
\noindent In this line, we will mention three categories of methods.

~\\
\noindent \textbf{\emph{- Moving average (rolling mean)}}\\
\noindent Generally speaking, smoothing methods take an important place in time series analysis, particularly, in time series trend estimation.
In this context, the \textit{moving average} method, also referred to as \textit{rolling mean} method, can adjust to a large range of trends.
\textit{Moving average} is a very simple \textit{scatterplot smoother} for which the initial data points $X_t$ are replaced with the means computed from successive intervals $I_t = \{X_{t'}, t' \in [a_t, b_t]\}$, with $a_t = \max(1, t-w)$, $b_t = t-1$ and $w$ the \textit{span size}. The trend estimate is then defined as
\begin{equation}
\label{moving.average}
\begin{split}
    \hat{T}_t = \frac{1}{|I_t|}\sum_{i \in I_t}^{} X_i,
\end{split}{}
\end{equation}{}
where $|I_t|$ stands for the cardinal of $I_t$. The smoothness of the estimated trend depends on the span size $w$: the larger $w$, the smoother the trend. It has to be noted that this method is particularly sensitive to outliers.

In this scheme, if the series is observed till time step $t$, at horizon $h > 0$, prediction is made as follows: the trend $\hat{T}_{t+h}$ is extrapolated according to equation (\ref{moving.average}), relying on observed values and previously extrapolated values $\hat{T}_{t+h'}, 0 < h' < h$.

~\\
\noindent \textbf{\emph{- Damped Holt additive trend method (generalized Holt method)}}\\
We first explain \textit{Simple exponential smoothing}, equivalently, \textit{exponentially weighted moving average}, a well-known smoothing method that also relies on a sliding window \citep{perry_2010_book_chapter_exponentially_weithted_ma_time_series}. This method is suited to noisy time series without trend. It is defined by the following recursion scheme:

\begin{equation}
\label{exponential.smoothing}
\begin{split}
	\tilde{X}_t = \alpha X_t + (1 - \alpha) \tilde{X}_{t-1},
\end{split}
\end{equation} 
with $\tilde{X}_t$ the smoothed value (or {\it level}) at time $t$, and $\alpha$ the \textit{smoothing parameter} comprised between $0$ and $1$. 

	It has to be noted that the lower the $\alpha$ parameter, the stronger the smoothing. No general method exits that allows to choose the best value for $\alpha$. The $\alpha$ parameter is generally adjusted thanks to expert knowledge that indicates how much weight should be given to the current {\it versus} past observations. Other data-driven selection methods were proposed \citep{gelper_fried_croux_2010_journ-forecasting_exponent-and-holt-winters-smooth-param-select}.

	The difference with the previously seen \textit{moving average} method is that each data point is smoothed based on weights that are exponentially decreasing as the observations are older. Thanks to the exponential decay of weights, \textit{exponential smoothing} is less sensitive to outliers than \textit{moving average}.

The \textit{damped Holt additive trend method}, also referred to as \textit{generalized Holt method}, is suited to time series that exhibit an increasing or decreasing trend \citep{gardner_mckenzie_1985_int-journ-forecasting_exponent-smooth-paramet-damped_holt_addit_trend,taylor_2003_int-journ-forecasting_exponent-smooth-damped-multipli-trend}. This method intertwines two exponential smoothings:
\begin{equation}
\label{damped.holt.method}
\begin{split}
	\tilde{X}_t &= \alpha X_t + (1 - \alpha) (\tilde{X}_{t-1} + \phi\ \hat{T}_{t-1}),\\
	\hat{T}_{t} &= \beta (\tilde{X}_t - \tilde{X}_{t-1}) + (1 - \beta)\ \phi\ \hat{T}_{t-1},
\end{split}
\end{equation} 
where $\tilde{X}_t$ is the smoothed series, $\hat{T}_t$ is the estimated trend, $\alpha$ and $\beta$ are the smoothing parameters comprised between $0$ and $1$, and $\phi$ is the dampening parameter that gives more control over trend exploration. The first \textit{exponential smoothing} removes random variations (noise), to produce the level of the series at time step $t$, and the second one smoothes the trend. 

If $\phi = 1$, we obtain the standard \textbf{Holt additive trend} method \citep{holt_2004_int-journ-forecasting_reprint_from_1957_report_trends_season_exponent_weighted_ma}. If $\phi = 0$, we obtain the \textit{simple exponential smoothing} method (no trend). If $0 < \phi < 1$, the trend extrapolation is damped and approaches an horizontal asymptote given by $\hat{T}_t\ \phi/(1-\phi)$. If $\phi > 1$, $\hat{T}_{t+h}$ has an exponential growth; this setting seems to be suited to series with exponential trends.

At horizon $h > 0$, the trend is extrapolated as
\begin{equation}
\hat{T}_{t+h} = \sum_{i=1}^h \phi^i \hat{T}_t.
\end{equation}

~\\
\noindent \textbf{\emph{- Local polynomial smoothing, locally estimated scatterplot smoothing}}\\
\noindent More flexible smoothing methods based on \textit{moving regression} have been introduced in the literature. These methods decompose the time series into trend and remaining component $Z_t$, following a scheme alternative to exponential smoothing. In this category, the \textbf{local polynomial} ({\sc lp}) smoothing method fits a polynomial (usually a straight line or parabola) on datapoints within time windows centered at each time step $t$ \citep{fan_gijbels_1996_book_local-polynomial-modeling-and-appli}. A simple or weighted \textit{least squares} method is used for this purpose. Then, the smoothed value at time $t$ equals the value of the corresponding polynomial at that time. The smoothing becomes stronger as the sliding window size increases. By removing the smoothed value from $X_t$, one obtains the stochastic component $Z_t = X_t - T_t$.
                                                                     
In the same category, the \textit{locally estimated scatterplot smoothing} ({\sc loess}) method introduces robustness to outliers into {\sc lp} smoothing \citep{cleveland_grosse_shyu_2017_chapter_local-regres-models}. This method operates iteratively. At each iteration, a weighted {\sc lp} smoothing is performed on the raw series (local polynomials are fitted using weighted least squares), and the weights are corrected as decribed hereafter.

At iteration $n$, the weight assignment to datapoints is governed by two aspects: (i) their distances to the datapoint under consideration (the one recorded at time step $t$); (ii) the residuals of the smoothing achieved at iteration $n-1$ (that is the difference between each datapoint and its smoothed version calculated at previous iteration). Following (i), close neighbors are assigned large weights. According to (ii), large residuals yield low weights. In this way, outliers characterized by very large residuals are assigned very low weights and will have a negligible effect in the next iteration.

~\\
\noindent $\bullet$ {\bf Parametric modeling of} $\bm{T_t}$ {\bf in the additive trend scheme}\\
\noindent The previously mentioned methods were designed to estimate the trend, to be further removed from the raw process $\mathbf{X}$. Alternatively, $T_t$ can be modeled by a parametric function of the time, that is $T_t = F(t;\theta)$, whose parameters $\theta$ are usually estimated by the \textit{least squares} method. However, in the overwhelming majority of time series, the trend model is unknown, which impedes the wide application of such parametric methods.

~\\
\noindent $\bullet$ {\bf Transformation of} $\bm{X_t}$ {\bf in the additive trend scheme}

\noindent In the above presented methods, the trend is either estimated or modeled, to be further removed from the raw process $\mathbf{X}$. An alternative is to transform $\mathbf{X}$ through {\bf differentiation}. Differentiation is an iterative process that ends up removing $T_t$ from the raw series. However, this method is suitable when the trend is regular and has a slow variation (as for polynomial functions). The differentiation operator is described as
\begin{equation}
\label{differentiation}
\begin{split}
    \Delta^1 X_t = \Delta X_t = (1 - L)X_t, \\ 
    \Delta^k X_t = \Delta^{k-1} (\Delta X_t) = (1 - L)^k X_t,
\end{split}{}
\end{equation}
where $\Delta X_t$ is the first-order difference of $X_t$, $\Delta^k X_t$ the $k^{th}$ order difference and $L$ the backward shift operator with $L X_t = X_{t-1}$ and $L^j X_t = X_{t-j}$.

In practice, successive differentiations are performed until a series without trend is obtained. $\mathbf{X}$ is said to be integrated to order $k$ if it becomes stationary after the $k^{th}$ application of first-order differentiation operator. Moreover, the removed trend $T_t$ is a $k$-degree polynomial function. $X_t$ can be recovered from $\Delta^k X_t$ and $\{X_{t-j}\}_{j=1}^{k}$ by inverse differencing using binomial expansion of $(1 - L)^k$:
\begin{equation}
\label{inverse.transformation} 
\begin{split}
    \Delta^k X_t = (1 - L)^k X_t = \left[ \sum_{j=0}^k \binom{k}{j} (-1)^j L^j \right] X_t  \\
    \implies X_{t} = \Delta^k X_t - \sum_{j=1}^{k} \binom{k}{j} (-1)^j\ X_{t-j}.
\end{split}{}
\end{equation}

The most popular additive trend-based model that uses differentiation is the {\sc arima} (Autoregressive Integrated Moving Average) model \citep{usman_2019_journ-applied-sci-envir-manag_arima-applied-neonatal-mortality,nyoni_2019_mpra_arima}. This model will be presented in Subsection \ref{arima}.

\subsubsection{Additive Seasonality Scheme: $\mu(t) = S_t$} 
\label{additive_seasonality}

\noindent In time series analysis, a seasonal phenomenon is characterized by variations that repeat after a fixed duration $m$ called period. As with additive trend, $S_t$ can be estimated, modeled or directly removed from $\mathbf{X}$ by differentiation.

~\\
\noindent $\bullet$ {\bf Estimation of} $\bm{S_t}$ {\bf in the additive seasonal scheme}\\
\noindent \cite*{pegels_1969_manag-sci_exponential-forecasting-new-variations} proposed an exponential smoothing method that copes with additive seasonal effect. The estimates are calculated as
\begin{equation}
\label{pegels.additive.seasonality}
\begin{split}
\tilde{X}_t &= \alpha(X_t + \hat{S}_{t-m} ) + (1 - \alpha) \tilde{X}_{t-1}, \\ 
\hat{S}_t   &= \gamma(\tilde{X}_t - X_t) + (1 - \gamma) \hat{S}_{t-m},\\
\end{split}
\end{equation}
with $\tilde{X}_t$ the smoothed series, $\hat{S}_t$ the $m$-periodic seasonal component estimate, $\alpha$ and $\gamma$ the smoothing parameters comprised between $0$ and $1$.

At period $t+h, h>0$, the seasonal effect is computed by periodicity, that is $\hat{S}_{t+h} = \hat{S}_{t+h-m}$.

~\\
\noindent $\bullet$ {\bf Parametric modeling of} $\bm{S_t}$ {\bf in the additive seasonal scheme}\\
\noindent In a parametric setting, $S_t$ can be modeled by an $m$-periodic parametric function (trigonometric polynomial function), $S_t = G(t;\theta)$, whose parameters are estimated by \textit{least squares}.

~\\
\noindent $\bullet$ {\bf Transformation of} $\bm{S_t}$ {\bf in the additive seasonal scheme}\\
If $S_t$ is an $m$-seasonal component, it can be removed from $X_t$ by successive $m$-order differentiations. This operation is called seasonal differentiation and is defined as
\begin{equation}
\label{m_order_differentiation}
\begin{split}
    \Delta_{m} X_t = \Delta_{m}^1 X_t = X_t - X_{t-m}, \;\ \Delta_m^k X_t = \Delta_m (\Delta_m^{k-1} X_t).
\end{split}{}
\end{equation}

\subsubsection{Additive Trend and Seasonality Scheme: $\mu(t) = T_t + S_t$} 
\noindent In this decomposition scheme, one has to correct for both additive trend and seasonality. Hereafter we present four categories of methods.

~\\
\noindent \textbf{\emph{- Sequential correction}}\\
\noindent $S_t$ and $T_t$ can be removed from the original time series one after another: a first correction is made with respect to the seasonal component $S_t$; then $T_t$ is removed. This procedure is implemented by the specific case {\sc sarima}$(p,d,q,P=0,D,Q=0)$ of the Seasonal Autoregressive Integrated Moving Average model \citep{samal_2019_int-conf-information_techno-compt-sci_sarima-prophet-model,valipour_2015_meteo-appli_sarima-arima,martinez_2011_revista-sociedade-brasileira-medicina-tropical_sarima}. In this specific case, $d$ denotes the number of classical differentiations (equation \ref{differentiation}, Subsection \ref{additive_trend}) and $D$ is the number of seasonal differentiations (equation \ref{m_order_differentiation}, Subsection \ref{additive_seasonality}). The {\sc sarima} model will be depicted in Section \ref{sarima}.

~\\
\noindent \textbf{\emph{- Simultaneous correction in additive Holt-Winters method}}\\
\noindent In contrast to the previous case, the \textit{additive Holt-Winters method} performs the two corrections simultaneously, using triple exponential smoothing \citep{aryee_essuman_djagbletey_darkwa_2019_journ-biostat-epidem_compar-sarima-holtwinters}. This extension of the standard \textit{Holt additive trend} method was specifically designed to cope with additive seasonal components. Herein, three \textit{exponential smoothing} processes are interwoven, that respectively smooth the series level ($\tilde{X}_t$), the trend and the seasonal components. To note, this method only supports increasing or decreasing trends. This method involves the three equations below:

\begin{equation}
\label{additive.holt.winter}
\begin{split}
	\tilde{X}_t &= \alpha (X_t - \hat{S}_{t-m}) + (1 - \alpha) (\tilde{X}_{t-1} + \hat{T}_{t-1}),\\ 
	\hat{T}_{t} &= \beta (\tilde{X}_t - \tilde{X}_{t-1}) + (1 - \beta) \hat{T}_{t-1},\\ 
	\hat{S}_t   &= \gamma (\tilde{X}_t - X_t) + (1 - \gamma) \hat{S}_{t-m}, 
\end{split}
\end{equation}    
with $\tilde{X}_t$ the smoothed series, $\hat{T}_t$ the trend estimate, $\hat{S}_t$ the $m$-periodic seasonal component estimate, $\alpha, \beta$ and $\gamma$ the smoothing parameters comprised between $0$ and $1$.

At horizon $h > 0$, the trend is extrapolated and the seasonal effect is computed by periodicity: 
\begin{equation}
\hat{T}_{t+h} = h\ \hat{T}_{t}, \qquad \hat{S}_{t+h} = \hat{S}_{t+h-m}. 
\end{equation}

~\\
\noindent \textbf{\emph{- Decomposition dedicated to quaterly and monthly data}}\\
\noindent \textbf{{\textsc x}11} and \textbf{{\textsc seat}s (Seasonal Extraction in \textsc{arima} Time Series)} are two popular methods that are limited to decomposing quaterly and monthly data. The reader interested in these specific methods is referred to the book of \citet{dagum_bianconcini_2016_book_season_time_series}. To note, {\textsc x}11 can handle both additive and multiplicative decompositions (see Subsection \ref{non.stationary.variance}).

~\\
\noindent \textbf{\emph{- Flexible Seasonal Trend decomposition using \textsc{loess}}}\\
\noindent In contrast to the above mentioned {\textsc x}11 and {\textsc seat}s methods, Seasonal Trend decomposition using {\sc loess} ({\sc stl}) is one of the most widely-used decomposition methods that can accommodate whatever trend and whatever periodicity \citep{cleveland_cleveland_mc-rae_terpenning_1990_journ-offi-stat_stl-season-trend-decomp,bergmeir_hyndman_benitez_2016_int-j-of-forecast_bagging-exponen-smooth-methods}. Thus daily, weekly, monthly, yearly, \textit{etc} time series may be processed.

The {\sc stl} iterative procedure essentially relies on {\sc loess} smoothing (previously introduced). {\sc stl} decomposition runs two embedded loops: (i) the inner loop iterates a user-specified number of iterations in which seasonal smoothing followed by trend smoothing respectively update the seasonal and trend components; (ii) in each outer loop iteration, a complete run of the inner loop is followed by the computation of robustness weights: for any datapoint, the neighborhood weight used throughout whole $(k+1)^{th}$ run of inner loop is that of $k^{th}$ run multiplied by the robustness weight. These robustness weights are used to control for aberrant behavior. We now briefly describe an inner loop iteration. At iteration {\it k}, the detrended series $X_t - T^{(k)}_t$ is first obtained. Cycle-subseries are distinguished (for example, in a monthly series with a yearly periodicity, cycle-subseries are the series related to January observations, February observations, {\it etc.}). Each subseries of the detrended series is then smoothed using {\sc loess}. The collection of all smoothed cycle-subseries values, $\{C^{(k+1)}_t\}$, is obtained. A succession of moving average steps followed by {\sc loess} is applied to $\{C^{(k+1)}_t\}$, to obtain $\{L^{(k+1)}_t\}$. The seasonal component obtained at $(k+1)^{th}$ iteration is then computed as follows: $\{S^{(k+1)}_t\} = \{C^{(k+1)}_t\} - L^{(k+1)}_t$. Finally, trend smoothing is applied on the deseasonalized series $X_t - \{S^{(k+1)}_t\}$, using {\sc loess}. These smoothed values constitute the trend component at $(k+1)^{th}$ iteration.

\begin{figure}[t]
\begin{tabular}{cc}
	\begin{subfigure}[b]{0.48\textwidth}
        \includegraphics[width=\textwidth]{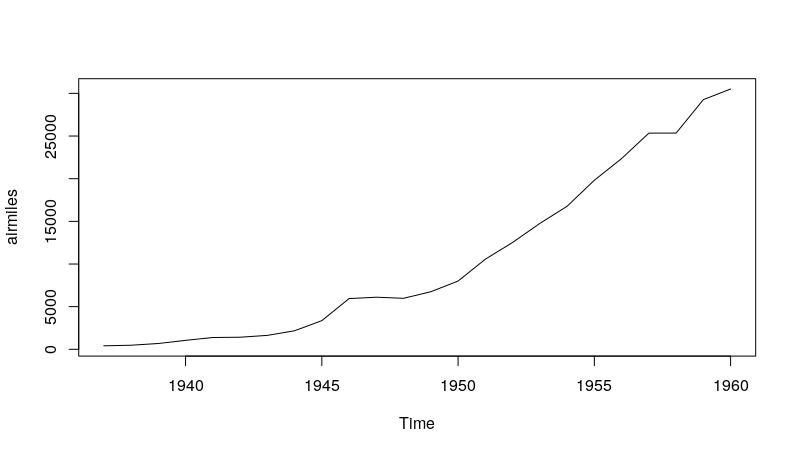}
        \caption{Nonstationary mean with exponentially increasing trend. Revenue passenger miles flown by commercial airlines in the United States.}
        \label{}
    \end{subfigure}
    &\begin{subfigure}[b]{0.48\textwidth}
        \includegraphics[width=\textwidth]{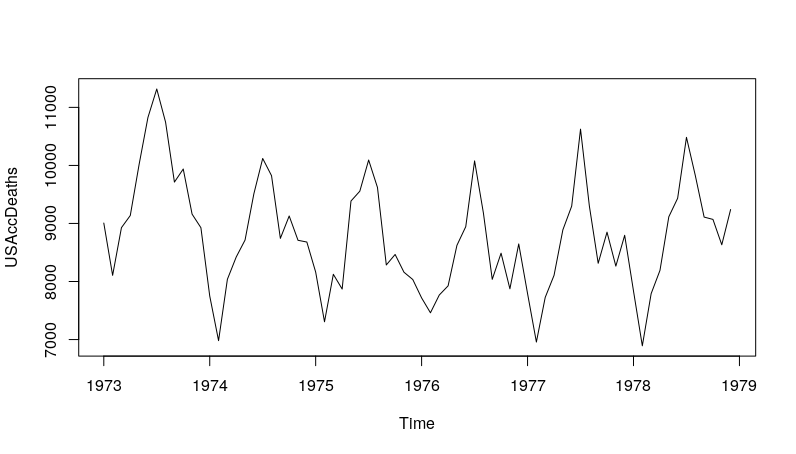}
        \caption{Nonstationary mean with additive seasonality. Monthly totals of accidental deaths in the United States.\\}
        \label{}
    \end{subfigure}\\
    \begin{subfigure}[b]{0.48\textwidth}
        \includegraphics[width=\textwidth]{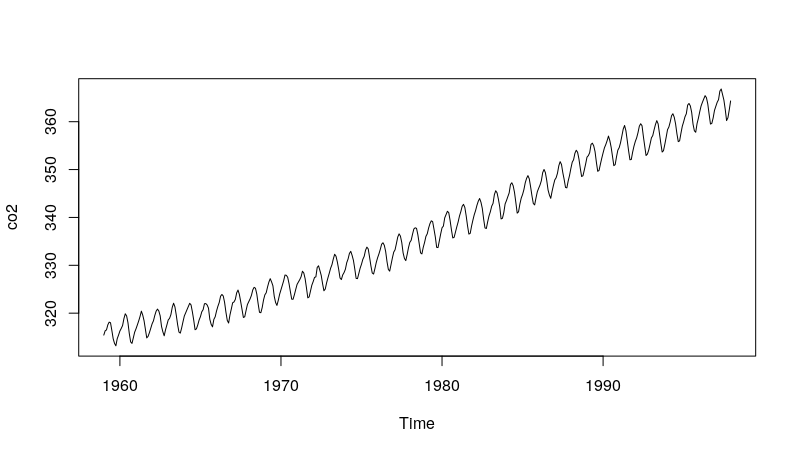}
        \caption{Nonstationary mean with additive trend and seasonality. Mauna Loa (Hawaii) atmospheric concentration of $CO_2$, expressed in parts per million (ppm); monthly observation from 1959 to 1997.}
        \label{co2.concentration.hawaii}
    \end{subfigure}
    &\begin{subfigure}[b]{0.48\textwidth}
        \includegraphics[width=\textwidth]{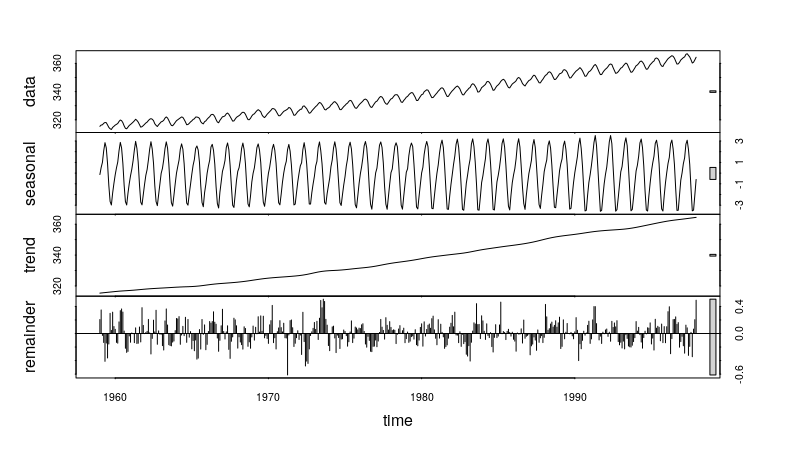}
        \caption{Seasonal Trend Decomposition ({\sc stl}) of time series shown in (c).\\\\}
        \label{stl.decomposition.of.co2.concentration.hawaii}
    \end{subfigure}
\end{tabular}
\newline
\caption[Square plots]{Illustration of nonstationary-mean real-life time series and of additive decomposition. }
\label{additive.decomposition}
\end{figure}

Figures \ref{co2.concentration.hawaii} and \ref{stl.decomposition.of.co2.concentration.hawaii} show the {\sc stl} decomposition of "Mauna Loa" volcano $CO_2$ concentration time series. 

\subsection{Multiplicative Decomposition - Nonstationary Variance}
\label{non.stationary.variance}

$\mathbf{X}$ is a nonstationary variance process if its variance is time-dependent, that is $\text{Var}(X_t) = \sigma^2(t)$. 
This situation is modeled as
\begin{equation}
\label{multiplicative.model}
  X_t = \sigma(t) \times Z_t,
\end{equation}{}

\noindent where $Z_t$ is the remaining stochastic component which is assumed to be stationary, and has a zero mean and a unit variance.

As for the $\mu(t)$ component in additive decomposition, three schemes may be described. $\sigma(t)$ is generally either a trend $T_t$ {\bf (multiplicative trend)} or a seasonal phenomenon $S_t$ {\bf (multiplicative seasonality)}, or any combination of both {\bf (multiplicative trend and seasonality)}. Multiplicative seasonality is illustrated with the following example: if in the summer months, 1, 000 more property transactions are performed than in the winter months, the seasonality is additive. In contrast, multiplicative seasonality corresponds to the situation where 10\% more transactions are performed in summer months than in winter months. In the multiplicative trend and seasonality case, a classical model consists in considering the product $T_t \times S_t$. In other words, under the additive (respectively multiplicative) seasonality assumption, the amplitude of the seasonal pattern does not depend (respectively depends) on the level of the series. Again, as for additive decomposition, the $\sigma(t)$ component in multiplicative decomposition can be estimated, then removed (in this case, $\hat{Z}_t = \frac{X_t}{\hat{\sigma}_t}$), or $\sigma(t)$ can be directly extracted from $\mathbf{X}$ by transformation.

~\\
\noindent $\bullet$ {\bf Estimation of} $\bm{\sigma_t}$ {\bf in the multiplicative scheme}\\
\noindent {\bf Multiplicative trend}, {\bf multiplicative seasonality} and {\bf multiplicative trend and seasonality schemes} were described by \citet{pegels_1969_manag-sci_exponential-forecasting-new-variations} in his comprehensive presentation of all nine combinations of trend and seasonal effects in additive form, multiplicative form or absence of effect. Therein, exponential smoothing methods were customized to deal with these three schemes:

\begin{equation}
\label{pegels.multiplicative_level_in_multi_trend}
\tilde{X}_t = \alpha X_t + (1 - \alpha) \tilde{X}_{t-1}\ \hat{T}_{t-1} \quad \text{(multiplicative trend),}
\end{equation}

\begin{equation}
\label{pegels.multiplicative_level_in_multi_season}
\tilde{X}_t = \alpha \frac{X_t}{\hat{S}_{t-m}} + (1 - \alpha)\tilde{X}_{t-1} \quad \text{(multiplicative seasonality),}
\end{equation}

\begin{equation}
\label{pegels.multiplicative_level_in_multi_trend_season}
\tilde{X}_t = \alpha \frac{X_t}{\hat{S}_{t-m}} + (1 - \alpha) \tilde{X}_{t-1}\ \hat{T}_{t-1} \quad \text{(multiplicative trend and seasonality)},
\end{equation}
where $\tilde{X}_t$ is the smoothed series, $\hat{T}_t$ the trend estimate, $\hat{S}_t$ the $m$-periodic seasonal component estimate, $\alpha$ is a smoothing parameter taking its value within $[0,1]$, and the multiplicative trend and seasonal components are derived from the formulas below: 

\begin{equation}
\label{pegels.multiplication.trend.seasonality.smoothing_multi_trend}
\hat{T}_t = \beta \frac{\tilde{X}_t}{\tilde{X}_{t-1}} + (1 - \beta) \hat{T}_{t-1},
\end{equation}

\begin{equation}
\label{pegels.multiplication.trend.seasonality.smoothing_multi_seasonal}
\hat{S}_t = \gamma \frac{X_t}{\tilde{X}_t} + (1 - \gamma) \hat{S}_{t-m},
\end{equation}

%
with $\beta$ and $\gamma$ smoothing parameters taking their values in $[0,1]$, and specified initial values $\tilde{X}_1$, $\hat{T}_1$, $\hat{S}_1, \cdots \hat{S}_m$.

At horizon $h > 0$, depending on the decomposition scheme, the trend is extrapolated and the seasonal effect is computed by periodicity: 
\begin{align}
\label{pegels.multiplication.trend.seasonality.estimation}
\hat{T}_{t+h} &= (\hat{T}_{t})^h\\
\hat{S}_{t+h} &= \hat{S}_{t+h-m}.
\end{align}\\
The multiplicative trend assumption involves equations (\ref{pegels.multiplicative_level_in_multi_trend}) and (\ref{pegels.multiplication.trend.seasonality.smoothing_multi_trend}), the multiplicative seasonality assumption relies on equations (\ref{pegels.multiplicative_level_in_multi_season}) and (\ref{pegels.multiplication.trend.seasonality.smoothing_multi_seasonal}), whereas the multiplicative trend and seasonal hypothesis builds on equations (\ref{pegels.multiplicative_level_in_multi_trend_season}), (\ref{pegels.multiplication.trend.seasonality.smoothing_multi_trend}) and (\ref{pegels.multiplication.trend.seasonality.smoothing_multi_seasonal}). 

Pegels suggests that a multiplicative trend "appears more probable in real-life applications" than an additive trend. However, additive trend is most used in practice. The reason lies in that the additive trend scheme yields a more conservative trend extrapolation (that is a linear extrapolation, in contrast to an exponential extrapolation for the multiplicative trend). Therefore, the additive trend assumption may be more robust when applied to a large variety of time series data. Following this observation, \cite*{taylor_2003_int-journ-forecasting_exponent-smooth-damped-multipli-trend} proposed a damped version of Pegels multiplicative trend model, in which the trend projection is damped by an extra parameter, in an analogous fashion to the damped Holt additive trend model (equation \ref{damped.holt.method}). This model is defined below:
\begin{align}
\label{damped.multiplicative.trend}
\begin{split}
\tilde{X}_t &= \alpha X_t + (1 - \alpha) \tilde{X}_{t-1} \hat{T}_{t-1}^{\phi},\\
\hat{T}_t &= \beta \frac{\tilde{X}_t}{\tilde{X}_{t-1}} + (1 - \beta) \hat{T}_{t-1}^{\phi},
\end{split}
\end{align}
where $\tilde{X}_t$, $\hat{T}_t$, $\alpha$ and $\beta$ have the same meaning as previously.

At horizon $h > 0$, we obtain 
\begin{align}
\hat{T}_{t+h} = \hat{T}_{t}^{\sum_{i=1}^h \phi^i}.
\end{align}

If $\phi=1$, we obtain the Pegels multiplicative trend model (equations \ref{pegels.multiplicative_level_in_multi_trend} and \ref{pegels.multiplication.trend.seasonality.smoothing_multi_trend}). If $\phi=0$, the method is identical to simple exponential smoothing (equation \ref{exponential.smoothing}). If $0 < \phi < 1$, the multiplicative trend is damped and extrapolations approach an horizontal asymptote given by $\hat{T}_{t}^{\phi(1 - \phi)}$. If $\phi > 1$, the trend extrapolation exponentially increases over time. 

Empirical studies on $1,428$ time series from various domains pointed out that damped Pegels multiplicative trend outperforms standard Pegels multiplicative trend \citep{taylor_2003_int-journ-forecasting_exponent-smooth-damped-multipli-trend}. To note, the same observation is done for damped Holt additive trend and standard Holt additive trend.

~\\
\noindent $\bullet$ {\bf Transformation of} $\bm{X_t}$ {\bf in the multiplicative scheme}\\
\noindent Log transformations and power transformations may stabilize the variance, prior to the use of an additive model. In the power transformation framework, parameter adjustment motivated multiple works. 

\begin{figure}[t]
    \begin{subfigure}[b]{0.5\textwidth}
        \includegraphics[width=\textwidth]{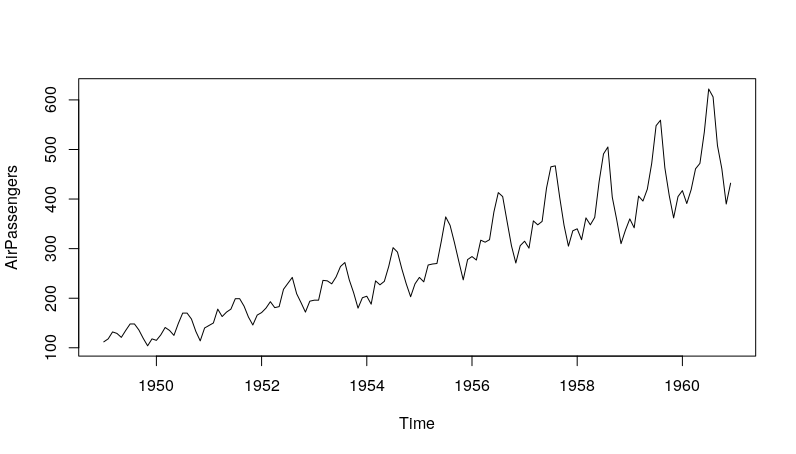}
        \caption{Initial time series.}
        \label{}
    \end{subfigure}
    \begin{subfigure}[b]{0.5\textwidth}
        \includegraphics[width=\textwidth]{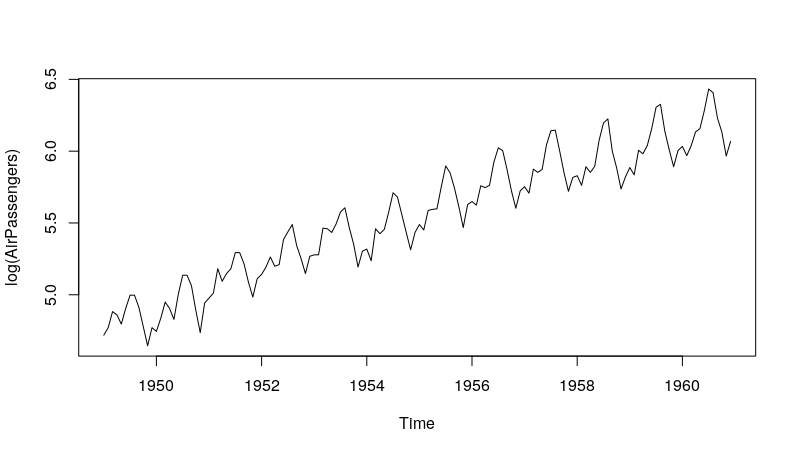}
        \caption{Logarithmic decomposition of the time series in (a).}
        \label{}
\end{subfigure}

\caption[Square plots]{Illustration of logarithmic transformation applied in the purely multiplicative decomposition framework. Monthly airline numbers of passengers from 1949 to 1960 in the United States, in thousands. $\mathbf{X}$: initial time series. $\mathbf{Y}$: time series obtained by transformation of $\mathbf{X}$. {\bf(a)} Initial real time series $\mathbf{X}$. In this example, the trend is increasing and a period of $12$ seems to exist. {\bf (b)} Multiplicative decomposition: $X_t = \sigma(t) \times Z_t$. In the purely multiplicative decomposition, $\sigma(t) = T_t \times S_t$. Thus, $Y_t = \ln(X_t) = \ln(T_t) + \ln(S_t) + \ln(Z_t)$, which results in an additive decomposition.
}
\label{mutiplicative.decomposition}
\end{figure}

~\\
\noindent \textbf{\emph{- Logarithmic transformation}}\\
\noindent In purely multiplicative decomposition, $\sigma(t) = T_t \times S_t$, the logarithmic transformation results in an additive decomposition (see time series in Figure \ref{mutiplicative.decomposition}), that is $Y_t = \ln(X_t) = \ln(T_t) + \ln(S_t) + \ln(Z_t)$, which has a constant variance. However, the logarithmic transformation is limited to time series with strictly positive values.

~\\
\noindent \textbf{\emph{- Box-Cox parametric power transformations}}\\
\noindent Box and Cox proposed parametric power transformations, to stabilize the variance, obtain more linear resulting series, as well as render the data more normal distribution-like \citep{box_cox_1964_journ-royal-stat-society_box-cox-transfo, bergmeir_hyndman_benitez_2016_int-j-of-forecast_bagging-exponen-smooth-methods}.

We illustrate below this category of transformations with the two following widely-used variants:
\begin{equation}
\label{box.cox}
Y_t = \left\{
    \begin{array}{ll}
        \frac{X_t^{\lambda_1}-1}{\lambda_1}  \quad \text{ if } \lambda_1 \neq 0\\
        \ln(X_t) \quad \text{ if } \lambda_1 = 0
    \end{array}
\quad \quad ; \quad \quad
Y_t = \right\{
    \begin{array}{ll}
         \frac{(X_t + \lambda_2)^{\lambda_1} - 1}{\lambda_1}  \quad  \text{ if } \lambda_1 \neq 0\\
         \ln(X_t + \lambda_2) \quad \text{ if } \lambda_1 = 0.
    \end{array}{}
\end{equation}{}

The left transformation is defined for strictly positive processes only, while the right one allows negativity through the appropriate choice of $\lambda_2$'s (real) value ($X_t > - \lambda_2$). Parameter $\lambda_1$ is real-valued and has be to calibrated. Many extensions of the original Box-Cox transformation have been introduced in the literature, in order to accommodate for such properties as bimodality and asymmetry \citep{sakia_1992_journ-royal-stat-soc_box-cox-transfo-review,hossain_2011_journ-emerging-trends-econom-manag_box-cox-transfo}.

~\\
\noindent \textbf{\emph{- Parameter adjustment in parametric power transformations}}\\
\noindent In their seminal paper, \citet{box_cox_1964_journ-royal-stat-society_box-cox-transfo} proposed maximum likelihood and Bayesian estimates for $\lambda_1$. 

Alternatively, the procedure introduced by Guerrero (1993) allows to estimate the $\lambda_1$ parameter in equation (\ref{box.cox}). We now shortly describe the principle underlying this procedure. If $\mathcal{T}$ denotes the power transformation function ($\mathcal{T}(X_t) = Y_t$), and $\mathcal{T}'$ denotes the derivative of $\mathcal{T}$, the Taylor expansion of $\mathcal{T}(X_t)$ about $\mathbb{E}[X_t]$ yields a linear approximation: $\text{Var}(\mathcal{T}(X_t)) = \mathcal{T}'(\mathbb{E}(X_t))^2\ \text{Var}(X_t)$. A variance-stabilizing power transformation for $X$ must then satisfy: $[\text{Var}(X_t)]^{(1/2)} / [\mathbb{E}(X_t)]^{1-\lambda_1} = a$ for some constant $a > 0$. In practice, when only one observation is available at each time step $t$, it is impossible to estimate the variance. To solve this problem, the time series under consideration can be divided into $H$ subseries; thus, a local estimate of variance and mean can now be computed within each subseries. The objective is then to stabilize the variance between subseries.

\cite*{osborne_2010_practic_assess_resea_eval_practical-box-cox} introduced a graphical protocol, to select $\lambda_1$ by hand. This procedure chains four steps: (i) to start with, the time series under consideration is divided into at least $10$ subseries; (ii) within each subseries, the mean and standard deviation are estimated; (iii) then, the curve $log(standard\ deviation)$ $versus$ $log(mean)$ is plotted; (iv) $\lambda_1$ is estimated as $1-b$, where $b$ is the slope of the previous curve.

Finally, a selection procedure by visual inspection is widely used by experienced analysts. It consists in applying power transformation using different values of $\lambda_1$ and comparing the plots obtained. The drawbacks of this procedure are its inaccuracy and the sujectivity involved in the visual inspection. Moreover, it can be time-consuming, especially if the tested $\lambda_1$ values are arbitrarily chosen.

\subsection{Mixed Decomposition - Nonstationary Mean and Variance}
\label{non.stationary.mean.variance}
To cope with $\mathbf{X}$'s both nonstationary mean and variance, we can use a model combining additive and multiplicative decompositions:
\begin{equation}
\label{mixed.model}
\begin{split}
    X_t = \mu(t) + \sigma(t) \times Z_t,
\end{split}{}
\end{equation}{}
where $Z_t$ is the remaining stochastic component which is supposed to be stationary and has a zero mean and a unit variance, and where $\mu(t)$ and $\sigma^2(t)$, respectively the mean and variance of $X_t$, have the same meaning as in previous Subsections \ref{non.stationary.mean} and \ref{non.stationary.variance}.

To note, the stationary component $Z_t$ can be extracted by successively applying the procedures previously presented for nonstationary mean (Subsection \ref{non.stationary.mean}) and nonstationary variance (Subsection \ref{non.stationary.variance}). On the other hand, further refined models resorting to the mixed decomposition framework were proposed, that exhibit trend and seasonality. In the remainder of this subsection, we mention two of these models.

\subsubsection{Additive Trend and Multiplicative Seasonal Component, Multiplicative Holt-Winters: $X_t = (T_t + Z_t) S_t$}

In \textit{multiplicative Holt-Winters}, $X_t = (T_t + Z_t)S_t$, that is $\mu(t) = T_t \times S_t$ and $\sigma(t) = S_t$, which means that additive trend and multiplicative seasonal components are considered \citep{holt_1957_onr-memorandum_expon-weight-aver-addit-trend-mult-season,holt_2004_int-journ-forecasting_reprint_from_1957_report_trends_season_exponent_weighted_ma,winters_1960_manag-sci_winters-meth_expo_weight_ma,aryee_essuman_djagbletey_darkwa_2019_journ-biostat-epidem_compar-sarima-holtwinters}. As in \textit{additive Holt-Winters} method (dedicated to additive trend and seasonality, $X_t = (T_t + S_t) + Z_t$)), this method performs a triple exponential smoothing defined by the following equations:

\begin{equation}
\label{multiplicative.holt.winter}
\begin{split}
	\tilde{X}_t &= \alpha \frac{X_t}{\hat{S}_{t-m}} + (1 - \alpha) (\tilde{X}_{t-1} + \hat{T}_{t-1}),\\
	\hat{T}_{t} &= \beta (\tilde{X}_t - \tilde{X}_{t-1}) + (1 - \beta) \hat{T}_{t-1},\\
	\hat{S}_t   &= \gamma \frac{X_t}{\tilde{X}_{t}} + (1 - \gamma) \hat{S}_{t-m},
\end{split}
\end{equation}
with $\tilde{X}_t$ the smoothed series, $\hat{T}_t$ the additive linear trend estimate, $\hat{S}_t$ the multiplicative $m$-periodic seasonal component estimate, $\alpha, \beta$ and $\gamma$ the smoothing parameters chosen from $[0,1]$.

At period $t+h, h > 0$, the trend is extrapolated and the seasonal component is determined thanks to periodicity:
\begin{equation}
\hat{T}_{t+h} = h\ \hat{T}_{t}, \qquad \hat{S}_{t+h} = \hat{S}_{t+h-m}.
\end{equation}

\subsubsection{Multiplicative Trend and Additive Seasonal Component, Pegels Mixed Model: $X_t = S_t + T_t\ Z_t$}
Not only did Pegels propose multiplicative model schemes. To tackle mixed decomposition, in contrast to multiplicative Holt-Winters, the model introduced by \citet{pegels_1969_manag-sci_exponential-forecasting-new-variations} considers a multiplicative trend and additive seasonal effect. This model also relies on triple exponential smoothing:

\begin{equation}
\label{pegels.multiplicative.trend.additive.seasonal}
\begin{split}
        \tilde{X}_t &= \alpha (X_t + \hat{S}_{t-L}) + (1 - \alpha) \tilde{X}_{t-1}\ \hat{T}_{t-1},\\
	\hat{T}_t   &= \beta \frac{\tilde{X}_t}{\tilde{X}_{t-1}} + (1 - \beta) \hat{T}_{t-1},\\
	\hat{S}_t   &= \gamma(\tilde{X}_t - X_t) + (1 - \gamma) \hat{S}_{t-m},
\end{split}
\end{equation}
with $\tilde{X}_t$ the smoothed series, $\hat{T}_t$ the multiplicative trend estimate, $\hat{S}_t$ the additive $m$-periodic seasonal component estimate, $\alpha, \beta$ and $\gamma$ the smoothing parameters which take values within $[0,1]$. 

At horizon $h > 0$, the trend is extrapolated and the seasonal effect is computed by periodicity: 
\begin{equation}
\hat{T}_{t+h} = (\hat{T}_{t})^h, \qquad \hat{S}_{t+h} = \hat{S}_{t+h-m}.
\end{equation}

\subsection{Recapitulation for Time Series Decomposition models}
\label{recapitulation.decomposition}

Table \ref{recapitulation.times.series.decompositions} provides a quick overwiew of the main categories of approaches dedicated to time series decomposition. 

\begin{table}[ht]
\scalefont{0.65} 
\centering
\begin{tabular}{|l|l|l|l|l|}
\hline
 \textbf{Decomposition} & \textbf{Characteristics} & \textbf{Trend and seasonal} & \textbf{Models} & \textbf{References}\\
                        &                          & \textbf{components}         &                 &\\
\hline
 & &                                                 & \multirow{2}{*}{differentiation ({\sc arima})$(p,d,q)$ ${}^{(1)}$}   & \cite{usman_2019_journ-applied-sci-envir-manag_arima-applied-neonatal-mortality}\\

 & &                                                &              & \cite{nyoni_2019_mpra_arima}\\
\cline{4-5}
                 &                 &            & Damped Holt additive trend method & \cite{taylor_2003_int-journ-forecasting_exponent-smooth-damped-multipli-trend}\\
                       &           &  Additive trend:& (generalized Holt method)   & \\
                       &           &  $\mu(t) = T_t$                               & Holt additive trend method & \cite{holt_2004_int-journ-forecasting_reprint_from_1957_report_trends_season_exponent_weighted_ma}\\
\cline{4-5}
                       &                &            & Local polynomial smoothing     & \cite{fan_gijbels_1996_book_local-polynomial-modeling-and-appli}\\
\cline{4-5}
Additive               &  Nonstationary &            & Locally estimated scatterplot & \cite{cleveland_grosse_shyu_2017_chapter_local-regres-models}\\
decomposition          &  mean          &             & smoothing ({\sc loess})             &\\
\cline{3-5}
$X_t = \mu(t) + Z_t$   &              &             & Pegels additive seasonality & \cite{pegels_1969_manag-sci_exponential-forecasting-new-variations}\\
                       &                & Additive seasonality:             & (double exponential smoothing)      &\\
\cline{4-5}
                       &                & $\mu(t) = S_t$                    & $m${-}order differentiation  &\\ 
                       &                &                                      & ($m$: seasonality period)  &\\
\cline{3-5}
    &          &  & {\sc sarima}$(p,d,q)(P=0,D,Q=0)_m$ ${}^{(2)}$& \cite{samal_2019_int-conf-information_techno-compt-sci_sarima-prophet-model}\\
    &          &  & ($m$: seasonality period) & \cite{valipour_2015_meteo-appli_sarima-arima}\\
    &          &  &        & \cite{martinez_2011_revista-sociedade-brasileira-medicina-tropical_sarima}\\
\cline{4-5}
    &              &    & Additive Holt-Winters method & \cite{holt_2004_int-journ-forecasting_reprint_from_1957_report_trends_season_exponent_weighted_ma}\\
    &              & Additive trend and          &                              & \cite{aryee_essuman_djagbletey_darkwa_2019_journ-biostat-epidem_compar-sarima-holtwinters}\\
\cline{4-5}
    &              & seasonality:                & Quaterly and monthly data    & \cite{dagum_bianconcini_2016_book_season_time_series}\\
    &              & $\mu(t) = T_t + S_t$        & {\sc x}11,                   & \\
    &              &                             & {\sc s}easonal {\sc e}xtraction in {\sc arima} {\sc t}ime & \cite{dagum_bianconcini_2016_book_season_time_series}\\
    &              &                       & {\sc s}eries ({\sc seat}s)          &\\
\cline{4-5}
    &              &                        & Flexible Seasonal Trend      & \cite{cleveland_cleveland_mc-rae_terpenning_1990_journ-offi-stat_stl-season-trend-decomp}\\
    &              &                        & decomposition using {\sc loess} ({\sc stl})   &\cite{bergmeir_hyndman_benitez_2016_int-j-of-forecast_bagging-exponen-smooth-methods}\\
\hline
 &    &     Multiplicative trend:  & Damped multiplicative trend & \cite{taylor_2003_int-journ-forecasting_exponent-smooth-damped-multipli-trend}\\
 &    &     $\sigma(t) = T_t$      & model                       &\\
\cline{4-5}
 &                  &  &  &\\
\cline{3-3}
               &                & Multiplicative seasonality:      &                                    & \\
Multiplicative &  Nonstationary & $\sigma(t) = S_t$                & Pegels multiplicative models       & \cite{pegels_1969_manag-sci_exponential-forecasting-new-variations}\\

\cline{3-3}
decomposition                & variance       & Multiplicative trend         & (exponential smoothing)  &\\
$X_t = \sigma(t) \times Z_t$ &                & and seasonality:             & &\\
                             &                & $\sigma(t) = T_t \times S_t$ ${}^{(3)}$ &  &\\ 
\cline{4-5}
                              &                & $\sigma(t) = T_t \times S_t$ ${}^{(4)}$ & Logarithmic transformation &\\
\cline{4-5}
                              &                & $\sigma(t) = function(T_t, S_t)$ ${}^{(5)}$ &  Box-Cox transformations & \cite{box_cox_1964_journ-royal-stat-society_box-cox-transfo}\\
                              &                &                                   & (power transformations) &\\
\hline
                    &                & Additive trend and  &  & \\
                    &                & multiplicative                                          & Multiplicative Holt-Winters             & \cite{holt_1957_onr-memorandum_expon-weight-aver-addit-trend-mult-season,holt_2004_int-journ-forecasting_reprint_from_1957_report_trends_season_exponent_weighted_ma}\\
                    &                & seasonality:                                            & method        & \cite{winters_1960_manag-sci_winters-meth_expo_weight_ma}\\
Mixed               &                & $\mu(t) = T_t \times S_t, \sigma(t) = S_t$              &         & \cite{aryee_essuman_djagbletey_darkwa_2019_journ-biostat-epidem_compar-sarima-holtwinters}\\
decomposition       & Nonstationary & $X_t = (T_t + Z_t)\ S_t$                                  &         &\\
\cline{3-5}
$X_t = \mu(t) + \sigma(t) \times Z_t$ &  mean and       & Multiplicative trend &  & \\
                                      & variance        & and additive         & Pegels mixed model & \cite{pegels_1969_manag-sci_exponential-forecasting-new-variations}\\
                                      &                 &  seasonality:        & (triple exponential smoothing)&\\
                                      &                 &  $\mu(t) = S_t$, $\sigma(t) = T_t$ &   &\\
                                      &                 &  $X_t = S_t + T_t\ Z_t $ & &\\
\hline
\end{tabular}
\caption{Time series decomposition and examples of corresponding models. $X_t$: initial time series. $Z_t$: the remaining component assumed to be stationary, with a zero mean and a unit variance. On the one hand, $X_t$ can be characterized through its moments: $\mu(t)$ (mean), $\sigma(t)$ (variance). On the other hand, in the decomposition scheme, $X_t$ can be characterized as the combination of a trend $T_t$, a seasonality $S_t$ and $Z_t$. ${}^{(1)}$ {\sc arima}$(p,d,q)$ is described in Subsection \ref{arima}. ${}^{(2)}$ {\sc sarima}$(p,d,q)(P=0,D,Q=0)_m$ is a specific instantiation of the {\sc sarima} model (Subsection \ref{sarima} ; Table \ref{special.cases.of.sarima.model}) for which the seasonal component reduces to $D$ $m$-order differentiations, with $m$ the seasonality period. ${}^{(3)}$ Pegels multiplicative models. ${}^{(4)}$ Logarithmic transformation. ${}^{(5)}$ Box-Cox transformations.}
\label{recapitulation.times.series.decompositions} 


\scalefont{1.0}
\end{table}

\subsection{Time Series Decomposition and Prediction}
\label{forecasting}

Except for exponential smoothing-based decomposition methods, the aim of time series decomposition is to filter out the deterministic features from a raw time series, and capture the remaining stochastic component. The latter is expected to show no apparent trend and seasonal variations. One step further this preprocessing task, modeling this remaining stochastic component is the basis to design a forecasting procedure. To achieve this aim, one attempts to fit some model on $\bm{Z}$: linear and nonlinear models  will be presented in Sections \ref{linear.time.series.models} and \ref{non.linear.time.series.models}. Taking into account $\bm{Z}$'s autocorrelation structure, when it exists, allows to increase prediction accuracy for the initial time series $\bm{X}$. Autocorrelation, the key to forecasting, is captured in such linear and nonlinear models through the modeling of the conditional mean and conditional variance. These latter notions will be detailed in Sections \ref{linear.time.series.models} and \ref{non.linear.time.series.models}. To note, the autocorrelation function is known for simple linear models, whereas it is very complicated to derive it for nonlinear models. The \textbf{Ljung-Box statistical test} allows to assess the absence of {\it residual} autocorrelation  at lag $r \ge 1$, relying on the squared sample autocorrelation function \citep{ljung-box_1978_biometrika_arma-model-valid-non-correl-test}. When this test is applied to residuals obtained from some fitted model, it allows to determine whether the model has captured the dependencies within the data. One expects that the more dependencies the model has captured, the more accurate the prediction relying on this model will be. 

In contrast, the autocorrelation in the initial time series is implicitly modeled in exponential smoothing-based decomposition methods. Indeed, the very aim of these specific decomposition methods is prediction. Therefore, the two next subsections will describe prediction in general when $\bm{Z}$ has been isolated by decomposition, and prediction in the specific case of exponential smoothing-based decomposition.

\subsubsection{Prediction focused on isolated stochastic component}

If autocorrelation can be assessed and modeled for the process $\bm{Z}$, the forecast for $\mathbf{X}$ at period $h>0$, denoted by $\hat{X}_{t+h}$, depends on the decomposition method used. If a transformation was used for multiplicative decomposition, $\hat{X}_{t+h}$ is computed from $\hat{Z}_{t+h}$ through inverse transformation: in a nutshell, the prediction is made for the transformed series ($\bm{Z}$); then the inverse transformation yields the prediction for the initial series; in this case, $\mu(t)$ and $\sigma(t)$ are directly filtered out from the original series. Apart from this specific transformation case, the deterministic components $\mu(t)$ and $\sigma(t)$ are estimated and $\hat{X}_{t+h}$ is obtained as a function of $\hat{\mu}(t+h)$ estimate, $\hat{\sigma}(t+h)$ estimate and $\hat{Z}_{t+h}$ forecast, following the decomposition scheme.

When the processus $\bm{Z}$ is stationary, the relationship existing between $Z_t$ and its lagged values is time-invariant, which simplifies the prediction. When $\bm{Z}$ is not stationary, another decomposition scheme should be tried for the initial series $\bm{X}$.

In the context of power-transformed time series (Box-Cox transformation, equation \ref{box.cox}, Subsection \ref{non.stationary.variance}), it has been shown that the forecasts obtained for the transformed series become biaised after the inverse transformation has been applied. The reason lies in the difference in scales between transformed and initial time series. Methods devoted to correct the resulting biais have been proposed in the literature \citep{guerrero_1993_journ-forecasting_analys-power-transfo-box-cox}.

\subsubsection{Prediction in the case of exponential smoothing-based decomposition}

\noindent In the decomposition framework, exponential smoothing-based methods hold a place apart. Indeed, no attention is paid therein to capture a remaining stochastic component. The reason lies in that these methods were designed for a prediction purpose, from the outset. Therefore, autocorrelation is implicitly acknowledged in $\bm{X}$'s modeling: applying exponential smoothing amounts to use an autoregressive model (see Subsection \ref{lar}) in which the coefficients decrease exponentially with time.

To note, exponential smoothing parameters can be chosen by minimizing the mean squared ahead-forecast errors \citep{gelper_fried_croux_2010_journ-forecasting_exponent-and-holt-winters-smooth-param-select}. The same procedure can be used in order to select the \textit{damping parameter} in generalized Holt method and damped multiplicative trend model \citep{taylor_2003_int-journ-forecasting_exponent-smooth-damped-multipli-trend}. On the other hand, initial values for the smoothed level, trend and seasonal components  must also be chosen. To this end, methods respectively based on linear regression, averaging and maximum likelihood have been described in the literature 
\citep{gardner_2006_int-journ-forecasting_exponent-smooth-paramet-initial-val-select}. It is important to note that the smaller the smoothing parameters, the more sensitive forecasts will be to the initial values (since a strong weight is attributed to past values).

Table \ref{recapitulation.exponential.smoothing.based.method.forecast} summarizes forecasting functions for the exponential smoothing-based methods introduced in Subsections \ref{non.stationary.mean} to \ref{non.stationary.mean.variance}.

If a significant autocorrelation is found in the residuals of an exponential smoothing method, defined  by $\epsilon_t = X_t - \hat{X}_{t}$ ($\hat{X}_{t}$ is computed using Table 3: $\hat{X}_{t} = \hat{X}_{(t-1)+h}$ with $h=1$), this indicates that the corresponding forecasts are suboptimal. In this case, prediction accuracy can be improved by adjusting a zero-mean first-order autoregressive {\sc ar}(1) model (see Subsection \ref{lar}) to the terms $\epsilon_t$'s \citep{gardner_1985_journ-of-forecasting_exponential-residuals-ajustment, taylor_2003_journ-operational-resear-soci_ar-1-residuals-in-exp-smooth}. Thus, at period $h$, the forecasts provided in Table \ref{recapitulation.exponential.smoothing.based.method.forecast} are modified by adding the term $(\phi_1)^h\ \epsilon_t$ where $\phi_1$ is the autoregressive coefficient of the {\sc ar}(1) model.\\\\

\begin{table}[ht]
\scalefont{0.85}
\centering
\begin{tabular}{|l|l|l|l|l|}
\hline
 \multirow{2}{*}{\textbf{Model}} &\multirow{2}{*}{\textbf{Decomposition}} &\multirow{2}{*}{\textbf{Trend $\hat{T}_{t+h}$}} & \multirow{2}{*}{\textbf{Seasonality $\hat{S}_{t+h}$}} & \multirow{2}{*}{\textbf{Forecast $\hat{X}_{t+h}$}}\\
   & &  & &\\
\hline
Damped Holt additive trend & $X_t = T_t + Z_t$ & $\sum_{i=1}^h \phi^i \hat{T}_t$ & - & $\tilde{X}_t + \hat{T}_{t+h}$\\ 
\hline
Pegels additive seasonality & $X_t = S_t + Z_t$ & - & $\hat{S}_{t+h-m}$ & $\tilde{X}_t + \hat{S}_{t+h}$\\ 
\hline
Additive Holt-Winters & $X_t = T_t + S_t + Z_t$ & $h \hat{T}_{t}$ & $\hat{S}_{t+h-m}$ & $\tilde{X}_t + \hat{T}_{t+h} + \hat{S}_{t+h}$\\ 
\hline
Taylor damped multiplicative trend  & $X_t = T_t\  Z_t$ & $\hat{T}_t^{\sum_{i=1}^h \phi^i}$ & - & $\tilde{X}_t\ \hat{T}_{t+h}$\\ 
\hline
Pegels multiplicative seasonality & $X_t = S_t\ Z_t$ & - & $\hat{S}_{t+h-m}$ &  $\tilde{X}_t\ \hat{S}_{t+h}$\\ 
\hline
Pegels multiplicative trend and seasonality & $X_t = T_t\ S_t\ Z_t$ & $(\hat{T}_t)^h$ & $\hat{S}_{t+h-m}$ & $\tilde{X}_t\ \hat{T}_{t+h}\ \hat{S}_{t+h}$\\ 
\hline
Multiplicative Holt-Winters & $X_t = (T_t + Z_t)\ S_t$ & $h \hat{T}_{t}$ & $\hat{S}_{t+h-m}$ & $(\tilde{X}_t + \hat{T}_{t+h})\ \hat{S}_{t+h}$\\ 
\hline
Pegels mixed model & $X_t = S_t + T_t\ Z_t$ & $(\hat{T}_t)^h$ & $\hat{S}_{t+h-m}$ & $\tilde{X}_t\ \hat{T}_{t+h} + \hat{S}_{t+h}$\\
\hline
\end{tabular}
\caption{Forecasting function for methods based on exponential smoothing.}
\label{recapitulation.exponential.smoothing.based.method.forecast}
\scalefont{1.0}
\end{table}

\section{Linear Time Series Models}
\label{linear.time.series.models}
As seen in section \ref{decomposition}, time series are generally decomposed into trend and seasonal effects, and a remaining stochastic component, following a decomposition model: additive, multiplicative, or mixed (equations \ref{additive.model}, \ref{multiplicative.model}, \ref{mixed.model}, respectively). In this section, we will focus on how the remaining stochastic component denoted by $\mathbf{Z} = \{Z_t\}_{t=1}^{\infty}$ can be modeled in order to make accurate predictions on $Z_t$. This prediction task requires the modeling of the relationship between past and future values of $\mathbf{Z}$: 

%
%
%


\begin{equation}
\label{time.series.model.specification_bis}
	Z_t = f(Z_{t-1}, Z_{t-2}, ...) + g(Z_{t-1}, Z_{t-2},...)\ \epsilon_t,
\end{equation}{}

where $\epsilon_t$'s are i.i.d.$(0,1)$ white noise processes, $f$ and $g^2$ are respectively the conditional mean and conditional variance of $Z_t$, that is $Z_t | Z_{t-1}, Z_{t-2},.. \sim \mathcal{D}(f, g^2)$ with $\mathcal{D}$ the law of residuals. Given information from the past, $f$ is the optimal forecast of $Z_t$ and $g^2$ is the forecast variance. This yields a forecast interval $[f-g, f+g]$. 

$f$ and $g$ can be modeled in a nonparametric way. This approach is outside the scope of the present survey. Further details can be found in the interesting review of nonparametric time series models presented in \cite{hardle_lutkepohl_chen_1997_int_stat_review_review-nonparametric-time-series}. The present survey will focus on the parametric modeling of $f$ and $g$.

When $f$ and $g$ are linear, that is when the relationship between $Z_t$ and its past values is linear, the model is said to be \textbf{linear}. Otherwise, the model is said to be \textbf{nonlinear} (see Section \ref{non.linear.time.series.models}). Linear time series models have been widely studied and applied in the literature \citep{ge-kerrigan_2016__arma-appli-ocean-wave-forecasting,gomes-castro_2012_joun-sustainable-energy_appli-wind-speed-power} because of their simplicity and strong theoretical foundation. These models rely on the stationarity hypothesis. As already highlighted in Subsection \ref{objectives.decomposition}, a relationship exists between stationarity and linearity: Wold's theorem (1938) shows that every stationary process can be represented as an infinite weighted sum of error terms. 


The most popular linear models dedicated to time series analysis are presented in the remainder of this section: linear autoregressive ({\sc ar}), moving average ({\sc ma}) and autoregressive moving average ({\sc arma}). These models differ from each other in the design of the conditional mean $f$. In contrast, they all assume a constant conditional variance $g^2$. These models are mainly analyzed under three aspects: autocorrelation function structure, model fitting methods and forecasting procedure. In the last subsection, {\sc arma}-based models such as {\sc arima} and {\sc sarima} (autoregressive integrated moving average and seasonal {\sc arima}, respectively) are briefly introduced. For presentation convenience, the {\sc sarima} variant is mentioned in the same Section as {\sc arma}, but it should be highlighted that {\sc sarima} is not a linear model.


In practice, there is no indication as to which linear model should be preferred over others. Therefore, model diagnosis and performance metrics should be considered in order to chose the best model for a given time series (see Section \ref{time.series.model.evaluation}).


\subsection{Linear Autoregressive Model - \textsc{ar}$(p)$}
\label{lar}
The linear autoregressive model has been introduced in the seminal paper of Yule in 1927. It is a regression model in which regressors are the $p$ lagged values of the response variable $Z_t$, where $p$ is called {\sc ar} process order. The {\sc ar} process is described as
\begin{equation}
\label{lar.process}
    Z_t = \phi_0 + \sum_{i=1}^{p} \phi_i\ Z_{t-i} + \sigma\ \epsilon_t,
\end{equation}
where the terms $\epsilon_t$ are i.i.d.$(0,1)$ white noise processes (usually, Gaussian white noise is used), $p$ is a hyper-parameter to be fixed and $\{\phi_0, \phi_1, ..., \phi_p, \sigma^2\}$ are the model parameters.

The {\sc ar} framework models the conditional mean $f$ of $Z_t$ given its past values as a linear regression of its $p$ past values, whereas its conditional variance $g^2$ is constant and equal to $\sigma^2$. It is well known that an {\sc ar} process is stationary if and only if the roots of its characteristic equation ($1 - \phi_1 y - \phi_2 y^2 - ... - \phi_p y^p = 0$) are all inside the unit circle, that is $ \mid y_i \mid < 1, \, i=1, \dots, p$.

It has to be noted that $Z_t$ is independent of future error terms $\{\epsilon_{t'}, t' > t\}$ and is correlated to past ones $\{\epsilon_{t'}, t' < t\}$ through its past values. For instance, in the {\sc ar}$(1)$ model, $Z_t$ depends on $Z_{t-1}$, which itself depends on $\epsilon_{t-1}$ and $Z_{t-2}$, which depends on $\epsilon_{t-2}$ and $Z_{t-3}$ and so on. In the end, $Z_t$ depends on all past error terms. This is the substantial difference between {\sc ar} and \textbf{moving average} processes introduced in Subsection \ref{ma}.

\subsubsection{Autocovariance Function}
The first and second order moments of the {\sc ar} process write
\begin{align}
\label{lar.1.order.moments}
    \mathbb{E}[Z_t] &= \phi_0 + \sum_{i=1}^{p} \phi_i\ \mathbb{E}[Z_{t-i}] \; \implies \; \mathbb{E}[Z_t] = \frac{\phi_0}{1-\sum_{i=1}^{p} \phi_i},\\
\label{lar.2.order.moments}
\gamma(h) &= \text{Cov}(Z_t, Z_{t+h}) = \text{Cov}(Z_t, \sum_{i=1}^{p} \phi_i\ Z_{t+h-i} + \sigma\ \epsilon_{t+h}) = \sum_{i=1}^{p} \phi_i\ \gamma(h-i), \;\ h = 0, 1, 2, \dots
\end{align}

Note that if $\phi_0$ is null, then $\mathbf{Z}$ is a zero-mean process. Moreover, an {\sc ar} process has a finite mean if and only if $\sum_{i=1}^{p} \phi_i \neq 1$.

	
When an {\sc ar} process is stationary, its autocovariance function $\gamma(h)$ has an infinite scope with an exponential decay, that is $Z_t$ is more correlated to nearby values in time than high-order lagged values. Let us show this result for the {\sc ar}$(1)$ process. It is straightforward to show that $\gamma(h) = \gamma(0)\ (\phi_1)^{h}$. Following the stationarity property (equation \ref{weak.stationarity}, Subsection \ref{weak.stationarity.def}), 
$\gamma(h)$ should go to zero when $h$ goes to infinity, which means $|\phi_1| < 1$. This result can be generalized to the {\sc ar}$(p)$ model, that is $|\phi_i| < 1, \; i=1, ..., p$.



\subsubsection{Partial Autocorrelation Function}
The definition of the autocorrelation function was provided in equation \ref{autocorrelation_formula} (Subsection \ref{stationary_test_graphical_methods}). The partial autocorrelation ({\sc pac}) function at lag $h$ is the autocorrelation between $Z_t$ and $Z_{t+h}$ in which the dependencies of $Z_t$ on $Z_{t+1}$ through $Z_{t+h-1}$ have been removed. Mathematically, {\sc pac} does not supply any new information on the studied process. In practice, a {\sc pac} plot is commonly used to identify the order of the {\sc ar} process since the partial autocorrelation of an {\sc ar}$(p)$ process is null from lag $p+1$. Estimation based on sample {\sc pac} is sensitive to outliers (see for instance \citealp{maronna_martin_yohai_2006_book_robust-statistics-partial-autocor-estimation}, pp. 247--257). Several robust algorithms are reviewed by \citet{durre_fried_liboschik_2015_report_partial-autocor-estimation-review}.

\subsubsection{Parameter Learning Algorithms}
\label{lar.parameter.learning}
An {\sc ar}$(p)$ model has $p+2$ parameters, $\mathbf{\phi} = \{\phi_0, \phi_1, ..., \phi_p\}$ and $\sigma^2$, which have to be estimated from observations $\{z_t\}_{t=1}^{T}$. The three methods exposed thereafter can be used to this end.\newline

\noindent $\bullet$ {\bf Maximum likelihood}
The conditional likelihood function $\mathcal{L}_c$ of a $p$-order autoregressive model is the probability of observing sample $\{z_t\}_{t=p}^{T}$ knowing the first $p$ values and parameters $(\phi, \sigma^2)$. It writes as 
\begin{align}
	\mathcal{L}_c(\phi, \sigma^2) = P(Z_p^T = z_p^T |Z_1^p=z_1^p; \phi, \sigma^2) = \prod_{t=p}^{T} P(Z_t=z_t|Z_{t-p}^{t-1}=z_{t-p}^{t-1}; \phi, \sigma^2),
\end{align}
where $Z_t^{t\prime} = (Z_t, ..., Z_{t'})$, with $t \le t'$, and the conditional probability $P(Z_t=z_t|Z_{t-p}^{t-1}=z_{t-p}^{t-1}; \phi, \sigma^2)$ is given by the law of residuals {$\epsilon_t$} {\it via} equation (\ref{lar.process}). For instance, when a Gaussian white noise is considered, the conditional probability is Gaussian too, with mean $ \phi_0 + \sum_{i=1}^p \phi_i\ z_{t-i}$ and variance $\sigma^2$.

A maximum likelihood estimator is obtained by maximizing the logarithm of $\mathcal{L}_c$ with respect to parameters $(\phi, \sigma^2)$. This maximization is generally driven numerically, but, in the case of a Gaussian white noise, analytical expressions can be derived.\newline

\noindent $\bullet$ {\bf Ordinary least squares (\textsc{ols})}
{\sc ar}$(p)$ parameters can be estimated by the least squares method: 
\begin{align}
\label{lar.ols.estimator}
	\hat{\phi}_{LS} = (D^t D)^{-1} Z D,
\end{align}
where $D = \{D_{i,j} = z_{i-j}\}_{i=p+1,...,T, j=1,...,p}$ is the $(T-p)\times (p+1)$ design matrix. Computing the {\sc ols} estimator requires a $p\times p$ matrix inversion at the cost of $\mathcal{O}(p^3)$. However, if $D^t D$ is singular, pseudo-inverse methods can be considered at the price of poor precision.\newline 
%
%
%


\noindent $\bullet$ {\bf Method of moments}
This method chains three steps: (i) establish a relationship between the moments of $Z_t$ and the model parameters, (ii) estimate these moments empirically, and (iii) solve the equation obtained {\it via} (i) with the estimates obtained through (ii). Equation (\ref{lar.2.order.moments}) establishes a relationship between the autocovariance function of $Z_t$ and parameters $\mathbf{\phi}$. After dividing by $\text{Var}(Z_t)$ in (\ref{lar.2.order.moments}), we obtain the same relationship in which autocovariance function $\gamma$ has been replaced with autocorrelation function $\rho$:
\begin{align}
\label{yule.walker.equation}
	\rho(h) &=  \sum_{i=1}^{p} \phi_i\ \rho(h-i), \quad h = 0,1,2, \dots 
\end{align}
The matricial reformulation of equation (\ref{yule.walker.equation}) yields
\begin{align}
\label{yule.walker.equation.matricial.formulation}
	\rho &= R\ \phi \quad \text{with} \quad \rho = (\rho(1), ..., \rho(p))^t, \;\ R = \{R_{i,j} = \rho(i-j)\}_{i,j=1,...,p},
\end{align}
where $\rho$ is a column vector and $R$ is a $p\times p$ symmetric, semi-definite positive matrix. Equation (\ref{yule.walker.equation.matricial.formulation}) is referred to as the \textbf{Yule-Walker} equation.

Once $\rho$ and $R$ have been estimated from sample autocorrelations, Yule-Walker equation can be solved either by matrix inversion ($\hat{\phi}_{YW} = \hat{R}^{-1}\hat{\rho}$) at the cost of $\mathcal{O}(p^3)$ operations, or using \textbf{Levinson-Durbin algorithm} \citep{durbin_1960_Revue-Institut-International-Statistique_AR-MA-ARMA-model-fitting} that requires a number of operations proportional to $p^2$ only. The latter method is an iterative procedure that solves a series of {\sc ar}$(p')$ truncated problems, with $0 \le p' \le p$. At each step, the size of the problem $p'$ is incremented.
	

\subsubsection{Forecasting}
\label{lar.forecasting}
Once the {\sc ar} model parameters have been estimated using observations $\{Z_t = z_t\}_{t=1}^T$, one-step ahead prevision is performed as follows:
\begin{equation}
\label{lar.process.forecasting}
    \hat{Z}_{T+1} =  \hat\phi_0 + \sum_{i=1}^{p} \hat\phi_i\ Z_{T+1-i},
\end{equation}
where $\{\hat{\phi}_0, \hat{\phi}_1, ..., \hat{\phi}_p\}$ are autoregressive coefficient estimates.

When performing $h$-step ahead previsions ($h > 1$), previous forecasts are used as predictors. For instance, in an {\sc ar}$(2)$ process, $3$-step ahead prevision writes
\begin{align*}
	\hat{Z}_{T+1} &= \hat{\phi}_0 + \hat{\phi}_1\ Z_T   + \hat{\phi}_2\ Z_{T-1},\\
	\hat{Z}_{T+2} &= \hat{\phi}_0 + \hat{\phi}_1\ \hat{Z}_{T+1} +  \hat{\phi}_2\ Z_{T}, \\
	\hat{Z}_{T+3} &= \hat{\phi}_0 + \hat{\phi}_1\ \hat{Z}_{T+2} +  \hat{\phi}_2\ \hat{Z}_{T+1},
\end{align*}	
where the unobserved terms $Z_{T+1}$ and $Z_{T+2}$ have been substituted with their predictions $\hat{Z}_{T+1}$ and $\hat{Z}_{T+2}$. It can be shown that these substitutions increase the variance of forecast errors. Following the previous example, the forecast errors write
\begin{align*}
	Z_{T+1} - \hat{Z}_{T+1} &= \epsilon_{T+1}, \\
	Z_{T+2} - \hat{Z}_{T+2} &= \epsilon_{T+2} + \hat{\phi}_1(Z_{T+1} - \hat{Z}_{T+1}) = \epsilon_{T+2} + \hat{\phi}_1\ \epsilon_{T+1},\\
	Z_{T+3} - \hat{Z}_{T+3} &= \epsilon_{T+3} + \hat{\phi}_1 (Z_{T+2} - \hat{Z}_{T+2}) + \hat{\phi}_2(Z_{T+1} - \hat{Z}_{T+1}) = \epsilon_{T+3} + \hat{\phi}_1\ \epsilon_{T+2} + (\hat{\phi}_1^2 + \hat{\phi}_2)\ \epsilon_{T+1}.
\end{align*}	
As $\epsilon_t$'s are i.i.d.($0, \hat{\sigma}^2$), the means of forecast errors are equal to zero and their variances write
\begin{align*}
	\text{Var}(Z_{T+1} - \hat{Z}_{T+1}) &= \hat{\sigma}^2, \\
	\text{Var}(Z_{T+2} - \hat{Z}_{T+2}) &= \hat{\sigma}^2(1 + \hat{\phi}_1^2),\\
	\text{Var}(Z_{T+2} - \hat{Z}_{T+2}) &= \hat{\sigma}^2(1 + \hat{\phi}_1^2 + (\hat{\phi}_1^2 + \hat{\phi}_2)^2).
\end{align*}

\subsection{Moving Average Model - \textsc{ma}$(q)$} 
\label{ma}

Moving average is a linear regression model in which the regressors are the $q$ prediction error terms. This model is depicted as
\begin{equation}
\label{ma.process}
    Z_t = \alpha_0 + \sum_{i=1}^{q} \alpha_i\ \epsilon_{t-i} + \epsilon_t,
\end{equation}
where $\epsilon_t$'s are error terms that are i.i.d.$(0, \sigma^2)$ (usually, the normal law is used), $q$ is the {\sc ma} process order and $\{\alpha_0, \alpha_1, \alpha_2, ..., \alpha_q, \sigma^2\}$ are the model parameters.

In the {\sc ma} model, the conditional mean $f$ of $Z_t$ given its past values is a linear function of the past $q$ prediction errors, whereas its conditional variance $g^2$ is constant and equal to $\sigma^2$.

\subsubsection{Autocovariance Function}
An {\sc ma} process oscillates around a long term equilibrium defined by its mean, where oscillation amplitude depends on error variance $\sigma^2$. Thus, by construction, this process is stationary. Its first and second order moments write
\begin{align}
\label{ma.1.order.moments}
    \mathbb{E}[Z_t] &= \alpha_0, \;\ \text{Var}(Z_t) = \sigma^2(1 + \sum_{i=1}^{q} \alpha_i^2),
\end{align}
\begin{equation}
\begin{split}
\label{ma.2.order.moments}
    \gamma(h) = \text{Cov}(Z_t, Z_{t+h}) &= \mathbb{E} \left[ \left(\alpha_0 + \sum_{i=1}^{q} \alpha_i\ \epsilon_{t-i} + \epsilon_t \right) \left(\alpha_0 + \sum_{j=1}^{q} \alpha_j\ \epsilon_{t+h-j} + \epsilon_{t+h} \right) \right] - \alpha_0^2 \\
    			 &= \begin{cases}
    			 	\sigma^2 (\alpha_h + \sum_{i=1}^{q-h} \alpha_i\ \alpha_{i+h}),     \quad 0 \le h \le q \\
    			 	0     \qquad\qquad\qquad\qquad\qquad            \;\                     h > q.
    			 \end{cases}
\end{split}
\end{equation}

Note that the autocovariance function $\gamma(h)$ is null from lag $q+1$, that is $Z_t$ and $Z_{t+\tau}$ are uncorrelated for all lags $\tau > q$.
This property characterizes the {\sc ma} process and is used to select hyper-parameter $q$. Indeed, $q$ is fixed at $h$ value such that $\hat{\gamma}(h+1)$ is null, where $\hat{\gamma}(h)$'s are sample autocovariances. 

\subsubsection{Infinite Autoregressive Representation}
It is well known that the {\sc ma}$(q)$ process has an infinite autoregressive representation in which the autoregressive coefficients are defined by a recursion scheme \citep{galbraith_zinde-walsh_1994_biometrika_ma-mode-simple-noniterative-estimator}. This representation is obtained through successive substitutions of the error terms in equation (\ref{ma.process}):
\begin{align}
\label{ma.infinite.lar.representation}
	Z_t &= \epsilon_t + \sum_{i=1}^{\infty} \phi_i\ Z_{t-i},
\end{align}
where the coefficients $\phi_i$ verify the following relation:
\begin{equation}
\label{relation.between.ar.and.ma.coefficients}
\begin{split}
	\phi_0  &= \alpha_0 = 0\\
	\phi_1  &= \alpha_1 \\
	\phi_2  &= -\alpha_1\ \phi_1 + \alpha_2 \\
        \vdots\\
	 \phi_q &= -\alpha_1\ \phi_{q-1} - \alpha_2\ \phi_{q-2} - \dots - \alpha_{q-1}\ \phi_1 + \alpha_q \\
	 \phi_j  &= \sum_{i=1}^q -\alpha_i\ \phi_{j-i} \quad (j = q+1, \dots),
\end{split}
\end{equation}
with $\{\alpha_i\}$ the regression coefficients of the {\sc ma} process.

In the case when the {\sc ma} process has a nonzero mean ($\alpha_0 \neq 0$), it can be centered prior using represention (equation \ref{ma.infinite.lar.representation}).

	It is important to emphasize that an {\sc ma}$(q)$ process can be well approximated by a finite autoregressive process of order $p$, with $p > q$ chosen sufficiently large, if and only if $|\alpha_i| < 1, \;\ i=1, \dots, q$ \citep{durbin_1959_biometrika_efficient-estimation-of-ma-models}. Indeed, this condition guarantees that the variance of the remaining terms from $p+1$ goes to zero when $p$ goes to infinity. Therefore, the remaining terms are neglectible for large values of $p$.  

\subsubsection{Parameter Learning Algorithms}
\label{ma.parameter.learning}
In the {\sc ma}$(q)$ process, $q+2$ parameters $\{\alpha_0, \alpha_1, \alpha_2, ..., \alpha_q, \sigma^2\}$ have to be estimated from the training dataset $\{z_t\}_{t=1}^{T}$. In practice, this is a difficult task because of the randomness of the regressors, which are the error terms $\epsilon_t$.


On the one hand, moment estimators have been proven statistically inefficient \citep{whittle_1953_arkiv-for-matematik_estimat-information-station-timeseries}. On the other hand, even though the likelihood function can be written in terms of the sample covariances \citep{whittle_1953_arkiv-for-matematik_estimat-information-station-timeseries}, its maximization requires solving a high-order nonlinear equation. \cite*{murthy_kronauer_1973_indian-journ-of-stat_moving-average-model-max-likeli-estim} proposed an approximation method that reduces data storage and computational effort, without substantial loss in the efficiency of the maximum likelihood.

An asymptotically efficient estimation procedure was proposed by \cite*{durbin_1959_biometrika_efficient-estimation-of-ma-models}. Durbin's method is one of the most widely-used {\sc ma} fitting techniques. This method relies on the ordinary least squares method ({\sc ols}) and implements two steps:

	\textbf{Step 1.} Following the infinite autoregressive representation of the moving average model (equation \ref{ma.infinite.lar.representation}), {\sc ma}$(q)$ is approximated by a $p$-order autoregressive model, with $p > q$. So, an {\sc ar}$(p)$ model is fitted to the observations $\{z_t\}$ {\it via} an {\sc ols} method (equation \ref{lar.ols.estimator}, Subsection \ref{lar.parameter.learning}). Let $\{ \hat{\phi}_i\}_{i=1}^p$ be the {\sc ols} estimation of the autoregressive coefficients $\{\phi_i\}_{i=1}^p$.
	
	\textbf{Step 2.} Using the relations (\ref{relation.between.ar.and.ma.coefficients}), a second autoregressive model is formed, in which $\{ \hat{\phi}_i\}_{i=1}^p$ are observed data and {\sc ma}$(q)$ parameters $(\alpha_1, \dots, \alpha_q)$ are coefficients. Then, {\sc ols} estimation provides a set of estimates  $\{\hat{\alpha}_i\}_{i=1}^q$. \\
Generally, $p$ is set at $2q$ or is selected via Akaike's information criterion ({\sc aic}) or Bayesian information criterion ({\sc bic}).

	\citet{sandgren_stoica_babu_2012_euro-signal-processing-conf_ma-param-estimate-method-comparison} proposed a procedure that simplifies Durbin's method by removing its second step. Thus, after the first autoregression fitting (first step of Durbin's method), an estimation of the $q$ moving average parameters is directly computed from the first $q$ relations in (\ref{relation.between.ar.and.ma.coefficients}). This simple estimator appears to outperform Durbin's estimator in small samples and seems more robust to model misspecification \citep{sandgren_stoica_babu_2012_euro-signal-processing-conf_ma-param-estimate-method-comparison}.
	
	Finally, it is well known that Durbin's estimator accuracy is degraded when coefficients $\alpha_i$'s are close to the unit circle (that is when $\mid \alpha_i \mid \longrightarrow 1$) for a fixed value of $p$. In this case, alternative methods with higher performances have been proposed \citep{sandgren_stoica_babu_2012_euro-signal-processing-conf_ma-param-estimate-method-comparison}. However, the performance of Durbin's estimator can be improved by increasing the value of $p$ at the cost of a greater computational complexity \citep{sandgren_stoica_babu_2012_euro-signal-processing-conf_ma-param-estimate-method-comparison}. 

\subsubsection{Forecasting}
\label{ma.forecasting}
Once {\sc ma} model parameters have been estimated using observations $\{Z_t = z_t\}_{t=1}^T$, one-step ahead prediction is performed as follows:
\begin{equation}
\label{ma.process.forecasting}
\begin{split}
    \hat{Z}_{T+1} | \epsilon_{T+1-q}^{T} &= \hat{\alpha}_0 + \sum_{i=1}^q \hat{\alpha}_i\ \epsilon_{T+1-i},
\end{split}{}
\end{equation}{}
where $\{ \hat{\alpha}_0, \hat{\alpha}_1, \hat{\alpha}_2, ..., \hat{\alpha}_q, \hat{\sigma}^2 \}$ are the parameter estimates and $\epsilon_{t} = Z_{t} - \hat{Z}_{t}$ are the forecast errors.


New predictions are adjusted with respect to the $q$ last forecast errors. For instance, in an {\sc ma}$(2)$ process, three-step ahead prediction writes
\begin{align*}
	\hat{Z}_{T+1} &= \hat{\alpha}_0 + \hat{\alpha}_1\ \epsilon_{T}   + \hat{\alpha}_2\ \epsilon_{T-1},\\
	\hat{Z}_{T+2} &= \hat{\alpha}_0 + \hat{\alpha}_1\ \epsilon_{T+1} +  \hat{\alpha}_2\ \epsilon_{T},\\
	\hat{Z}_{T+3} &= \hat{\alpha}_0 + \hat{\alpha}_1\ \epsilon_{T+2} +  \hat{\alpha}_2\ \epsilon_{T+1}.
\end{align*}
%
Since the series has only been observed till time step $T$, $\epsilon_{T+1}$ and $\epsilon_{T+2}$ are unknown forecast errors. In multi-step ahead forecasts, the unknown forecast errors are commonly omitted at the price of a variance increase for forecast errors. Thus, the above example yields
%
\begin{align*}
	\hat{Z}_{T+1} &= \hat{\alpha}_0 + \hat{\alpha}_1\ \epsilon_{T} + \hat{\alpha}_2\ \epsilon_{T-1} \hspace{-15mm} &\text{ and } \quad  Z_{T+1} - \hat{Z}_{T+1} &= \epsilon_{T+1},\\
	\hat{Z}_{T+2} &= \hat{\alpha}_0 + \hat{\alpha}_2\ \epsilon_{T} \hspace{-15mm} &\text{ and } \quad   Z_{T+2} - \hat{Z}_{T+2} &= \epsilon_{T+2} + \hat{\alpha}_1\ \epsilon_{T+1},\\
	\hat{Z}_{T+3} &= \hat{\alpha}_0 \hspace{-15mm} &\text{ and } \quad Z_{T+3} - \hat{Z}_{T+3} &= \epsilon_{T+3} + \hat{\alpha}_1\ \epsilon_{T+2} + \hat{\alpha}_2\ \epsilon_{T+1}.
\end{align*}
As $\epsilon_t$'s are i.i.d.($0, \hat{\sigma}^2$), the forecast errors have a zero mean and their variances write 
\begin{align*}
	\text{Var}(Z_{T+1} - \hat{Z}_{T+1}) &= \hat{\sigma}^2,\\
	\text{Var}(Z_{T+2} - \hat{Z}_{T+2}) &= \hat{\sigma}^2(1 + \hat{\alpha}_1^2),\\
	\text{Var}(Z_{T+3} - \hat{Z}_{T+3}) &= \sigma^2(1 + \hat{\alpha}_1^2 + \hat{\alpha}_2^2).
\end{align*}
It has to be noticed that the variance of forecast error grows as $\sigma^2(1 + \hat{\alpha}_1^2 + \hat{\alpha}_2^2 + ...)$.\\

Alternatively, the unknown forecast errors can be simulated as i.i.d.($0, \hat{\sigma}^2$) processes from the law chosen for error terms. 	

\subsection{Autoregressive Moving Average Model - \textsc{arma}$(p,q)$} 
\label{arma}
Introduced by Box and Jenkins in 1970, the {\sc arma}$(p,q)$ model (also called Box-Jenkins model) is the combination of the autoregressive {\sc ar}$(p)$ and moving average {\sc ma}$(q)$ models \citep{box_jenkins_reinsel_et_al_2015_book_time_series_analysis}. The {\sc arma} model is described as below:
\begin{equation}
\label{arma.process}
Z_t = \phi_0 + \sum_{i=1}^{p} \phi_i\ Z_{t-i} + \sum_{j=1}^{q} \alpha_j\ \epsilon_{t-j} + \epsilon_t,
\end{equation}
where the $\epsilon_t$'s are error terms that are i.i.d.$(0,\sigma^2)$ (usually, a normal law is used), $(p,q)$ are hyper-parameters to be fixed and $\{\phi_0, \phi_1, ..., \phi_p, \alpha_1, \alpha_2, ..., \alpha_q, \sigma^2\}$ are the model parameters.

	When $q=0$, equation (\ref{arma.process}) gives an {\sc ar}$(p)$ process and when $p=0$, it defines an {\sc ma}$(q)$ model. We emphasize that the {\sc arma}$(p,q)$ process is stationary if and only if its {\sc ar} component is stationary. If $\phi_0 = 0$, the mean of the {\sc arma} process will be zero.

\subsubsection{Autocovariance Function}

As seen in Subsection \ref{ma} (respectively \ref{lar}), the {\sc ma}$(q)$ model (resp. the {\sc ar}$(p)$ model) is suited to stationary processes for which the autocovariance function $\gamma(h)$ has a scope $q$ (resp. has an exponential decrease). As a sum of {\sc ar}$(p)$ and {\sc ma}$(q)$ processes, {\sc arma}$(p,q)$ model can handle a wider range of autocovariance functions using few parameters. Fitting an {\sc ar} or {\sc ma} model to data dynamics may require a high-order model with many parameters. Mixing {\sc ar} and {\sc ma} models allows a more parsimonious description.


It is well known that any stationary {\sc arma} process has an infinite-order moving average {\sc ma}$(\infty)$ representation:
\begin{align}
\label{arma.ma.infinity.representation}
	 Z_t = \phi_0 +  \sum_{j=0}^{\infty} \beta_j\ \epsilon_{t-j} \quad \text{with} \quad \sum_{j=0}^{\infty} |\beta_j| < \infty.
\end{align}
Then, the autocovariance function of the {\sc arma} process writes
\begin{align}
\label{arma.second.order.moment}
	\gamma(h)  &= \sigma^2 \sum_{j=1}^{\infty} \beta_j \beta_{j+h}.
\end{align} \\
It can be shown that $\gamma(h)$ shows an exponential decrease when $h$ goes to infinity, as for the autocovariance function in {\sc ar} processes. For instance, in zero-mean {\sc arma}$(1,1)$ process
 $$Z_t = \phi_1\ Z_{t-1} + \alpha_1\ \epsilon_{t-1} + \epsilon_t,$$ 
after incorporating $Z_{t-1} = \phi_1\ Z_{t-2} + \alpha_1\ \epsilon_{t-2} + \epsilon_{t-1} $ into the previous equation, we obtain: 
$$ Z_t = \phi_1^2\ Z_{t-2} + \phi_1\ \alpha_1\ \epsilon_{t-2} + (\phi_1 + \alpha_1)\ \epsilon_{t-1} + \epsilon_t.$$
Finally, after substituting all past values $Z_{t-2}, Z_{t-3}, ...$, we obtain representation (\ref{arma.ma.infinity.representation}), where $\beta_j = \phi_1^{j-1}(\phi_1 + \alpha_1)$ and $\beta_0 = 1$. So, since $|\phi_1| < 1 $, $\beta_j \beta_{j+h} = \phi_1^{h+2j-2} (\phi_1 + \alpha_1)^2 \longrightarrow 0$ as $h \longrightarrow \infty$, whatever the fixed value of $j$. Therefore, $\gamma(h) \longrightarrow 0$ as $h \longrightarrow \infty$.
		
The orders $(p,q)$ of an {\sc arma} process can be chosen relying on the partial autocorrelation function and the autocorrelation function, as  is done for {\sc ar} and {\sc ma} processes \citep{box_jenkins_reinsel_et_al_2015_book_time_series_analysis}.

\subsubsection{Parameter Learning Algorithms}
\label{arma.parameter.learning}
{\sc arma}$(p,q)$ process fitting consists in estimating its parameters $\{\phi_0, \phi_1, ..., \phi_p, \alpha_1, \alpha_2, ..., \alpha_q, \sigma^2\}$ from observations $\{z_t\}_{t=1}^{T}$. In this subsection, some methods proposed in the literature are presented.

\citet{jones_1980_journal-technometrics_arma-model-ml-estim} proposed a maximum likelihood estimation of the {\sc arma} model in the case of Gaussian white noise. This estimation procedure relies on the Markovian representation, an information interface between the future and the past of a discrete-time stochastic process, whose existence was established for {\sc arma} processes \citep{akaike_1998_book_selected-articles-akaike_markovian-represent-likeli_estim}. The Markovian representation provides a minimal state-space representation for the recursive calculation of the likelihood function, under the Gaussian white noise assumption.

On the other hand, a three-step method was designed for {\sc arma} model fitting, including the estimation of the autoregressive and moving average orders $(p,q)$ \citep{hannan_rissanen_1982_biometrika_arma-fitting-and-order-selection,hannan_kavalieris_1984_biometrika_arma-fitting-and-order-selection_improvement}. The three steps are as follows:

	\textbf{Step 1.} The error terms $\hat{\epsilon}_t$ are obtained by fitting an {\sc ar}$(n)$ model to the data, for a large $n$:
	$$ \hat{\epsilon}_t = z_t + \sum_{i=1}^n \hat{\phi}_i\ z_{t-i}, $$
with $\{\hat{\phi}_i\}_{i=1}^n$ the autoregressive coefficients estimated via the Yule-Walker equation (\ref{yule.walker.equation}) (see Subsection \ref{lar}). 


		\textbf{Step 2.} Using $\{z_t\}$ and $\{\hat{\epsilon}_t\}$ as regressors, {\sc arma}$(p, q)$ parameters are estimated by the least squares method. Then, the orders $(p,q)$ are estimated by minimizing the following criterion:
					$$\log(\hat{\sigma}^2) + \frac{(p + q)}{T} \log(T),$$
where $\hat{\sigma}^2$ is the least squares estimate of $\sigma^2$ and $T$ is the number of observations (that is, the number of time steps). This procedure can be costly when a wide grid of $(p,q)$ values is tested. In the scenario where $p=q$, an efficient algorithm that recursively computes the sequence of {\sc arma}$(p, q)$ regressions has been proposed \citep{hannan_rissanen_1982_biometrika_arma-fitting-and-order-selection}.


\textbf{Step 3.} Once $(p,q)$ are determined, {\sc arma}$(p,q)$ parameters are estimated through an iterative optimization procedure of the likelihood function. This function is initialized with the {\sc arma} parameters corresponding to $(p,q)$ identified by step 2.
\newline

	\citet{franke_1985_biometrika_generaliz-levinson-durbin-recurs-arma} generalized the \textbf{Levinson-Durbin algorithm} used for {\sc ar} parameter learning to {\sc arma} model fitting.

	The \textbf{Innovation algorithm} is a recursive method used to compute {\sc arma}$(p,q)$ model parameters \citep{brockwell_davis_fienberg_1991_book_arma-fitting-innovation-algo}. Innovation is defined as the difference between the observed value $z_t$ at time step $t$ and the optimal forecast of that value, based on past information. In a parameter estimation framework, the motivation behind the Innovation algorithm lies in that in the innovation time series, the successive terms are uncorrelated with each other, thus yielding a white noise time series. Evaluating the likelihood directly for $\bm{Z}$ involves inverting a nondiagonal covariance matrix which may also be a cumbersome function of the model parameters. Instead, the (white noise) innovation series has a diagonal covariance matrix, which is much easier to invert. \citet{sreenivasan_1998_journ-comput-applied-math_arma-fitting-innovation-algo-order-select} extended the Innovation algorithm in order to include the selection of the {\sc arma} process orders $(p,q)$.

%



\subsubsection{Forecasting}
\label{arma.forecasting}


Once {\sc arma}$(p,q)$ model parameters have been estimated from sample $\{Z_t = z_t \}_{t=1}^T$, forecast at horizon $h$, $\hat{Z}_{T+h}$, is the sum of the forecasts for {\sc ar} (equation \ref{lar.process.forecasting}) and {\sc ma} (equation \ref{ma.process.forecasting}) components at the same horizon $h$:

\begin{align}
\label{arma.process.forecasting}
	\hat{Z}_{T+h} | \epsilon_{T+h-q}, \dots, \epsilon_{T+h-1}  &= \hat{\phi}_0 + \sum_{i=1}^{p} \hat\phi_i\ Z_{T+h-i} + \sum_{i=1}^q \hat{\alpha}_i\ \epsilon_{T+h-i},
\end{align}
where $Z_{T+h-i} = \hat{Z}_{T+h-i}$ when $h-i > 0$. 


As for the {\sc ma} model (see Subsection \ref{ma.forecasting}), the error terms which are unknown are omitted.

\subsection{Autoregressive Moving Average-based Models}
As seen in Subsection \ref{arma}, the \textit{autoregressive moving average} ({\sc arma}) model is the most general linear stationary model that allows to specify the conditional mean. Since trend and seasonal components are observed in most real-world time series, the {\sc arma} model has been extended to include these two deterministic components. The {\sc arima} and {\sc sarima} models are described in this subsection.

\subsubsection{Autoregressive Integrated Moving Average Model - \textsc{arima}$(p,d,q)$}
\label{arima}


The \textit{autoregressive integrated moving average} ({\sc arima}) model falls within the additive trend decomposition scheme (Subsection \ref{additive_trend}). The {\sc arima}$(p,d,q)$ process has a nonstationary mean and is suited to trended time series whose trends can be removed after $d$ successive differentiations (equation \ref{differentiation}, Subsection \ref{additive_trend}). Then, the resulting stationary component is an \textit{autoregressive moving average} process of order $(p,q)$. A time series $\{X_t\}$ follows an {\sc arima}$(p,d,q)$ process if 
\begin{align}
\label{arima.process}
 \Phi_p (L)\ (1 - L)^d\ X_t &=  \Theta_q (L)\ \epsilon_t,
\end{align}
where $\{\epsilon_t \}$ is a white noise series; $p, d, q$ are integers; $L$ is the backward shift operator ($L X_t = X_{t-1},$ $L^k X_t = X_{t-k}$), $\Phi_p$ and $\Theta_q$ are polynomials in $L$, of orders $p$ and $q$, respectively:
\begin{align*}
 \Phi_p (L)        &=  1 - \phi_1 L - \phi_2 L^2 - \dots - \phi_p L^p, \\
 \Theta_q (L)      &=  1 - \alpha_1 L - \alpha_2 L^2 - \dots - \alpha_q L^q.
\end{align*}

The most known specific instantiations of the {\sc arima} model are summed up in Table \ref{special.cases.of.arima.model}.
%

\begin{table}[ht]
\scalefont{0.85}
\centering
\begin{tabular}{|l|l|}
\hline
Instantiation                                   & {\sc arima}$(p,d,q)$\\
\hline
White noise               \qquad\qquad\qquad    & {\sc arima}$(0,0,0)$      \\
Random walk process       \qquad\qquad\qquad    & {\sc arima}$(0,1,0)$        \\
Autoregression 	          \qquad\qquad\qquad    & {\sc arima}$(p,0,0)$         \\
Moving average            \qquad\qquad\qquad    & {\sc arima}$(0,0,q)$          \\
{\sc arma}                \qquad\qquad\qquad    & {\sc arima}$(p,0,q)$          \\
\hline
\end{tabular}
\caption{Most popular instantiations of the {\sc arima} model. $p$ denotes the order of the {\sc ar} process (see Subsection \ref{lar}), $d$ is the number of differentiations required to detrend the raw series, $q$ is the order of the {\sc ma} process (see Subsection \ref{ma}).}
\label{special.cases.of.arima.model}
\scalefont{1.0}
\end{table}

It has to be underlined that there exists no automatic method to identify the number $d$ of differentiations required to detrend the time series. In practice, different values are tested $(1, 2, 3, \dots)$. Then two strategies can be applied: (i) $d$ is set as the first value for which the $d^{th}$ differentiation $(1 - L)^d X_t$ is stationary (see Subsection \ref{stationarity.test}); (ii) $d$ is chosen to minimize Akaike's information criterion ({\sc aic}) or Bayesian information criterion ({\sc bic}).

Let $\{X_t\}$ an {\sc arima}$(p,d,q)$ process and $\{X_t = x_t\}_{t=1}^T$ the observed data. The stationary component $Z_t = (1 - L)^d X_t$ $=\Delta^d X_t$, with $\Delta$ the differentiation operator (equation \ref{differentiation}, Subsection \ref{additive_trend}), is an {\sc arma}$(p,q)$ process. At horizon $h$, $Z_{T+h}$ is predicted through formula (\ref{arma.process.forecasting}). Then, the forecast $\hat{X}_{T+h}$ for the observed series is obtained by inverse transformation (equation \ref{inverse.transformation}, Subsection \ref{additive_trend}):
\begin{align}
\label{arima.forecasting}
 	\hat{X}_{T+h} = \hat{Z}_{T+h} - \sum_{j=1}^{d} \binom{d}{j} (-1)^j\ X_{T+h-j},
\end{align}
where $X_{T+h-j} = \hat{X}_{T+h-j}$ when $h-j > 0$.

\subsubsection{Seasonal Autoregressive Integrated Moving Average Model - \textsc{sarima}$(p,d,q)(P,D,Q)_m$} 
\label{sarima}

As seen in Subsection \ref{arima}, the {\sc arima} model handles additive trend but does not account for seasonal episodes. Seasonal {\sc arima} overcomes this limitation. It is the most popular model in the field of seasonal time series forecasting. For presentation fluency, the {\sc sarima} extension is mentioned in the same Section as {\sc arima}, but it must be kept in mind that {\sc sarima} is not a linear model. A time series $\{ X_t \}$ is generated by a {\sc sarima}$(p,d,q)(P,D,Q)_m$ process if 


%
\begin{align}
\label{sarima.process}
 \Phi_p (L)\ \Phi_P (L^m)\ (1 - L)^d\ (1 - L^m)^D\ X_t &=  \Theta_q (L)\ \Theta_Q (L^m)\ \epsilon_t,
\end{align}
where $\{\epsilon_t \}$ is a white noise series, $p, d, q, P, D, Q$ and $m$ are integers, $L$ is the backward shift operator ($L^k X_t = X_{t-k}$), and
\begin{align*}
 \Phi_p (L)        &=  1 - \phi_1 L - \phi_2 L^2 - \dots - \phi_p L^p, \\
 \Phi_P (L^m)      &=  1 - \phi_m L^m - \phi_{2m} L^{2m} - \dots - \phi_{Pm} L^{Pm}, \\
 \Theta_q (L)      &=  1 - \alpha_1 L - \alpha_2 L^2 - \dots - \alpha_q L^q, \\
 \Theta_Q (L^m)    &=  1 - \alpha_m L^{m} - \alpha_{2m} L^{2m} - \dots - \alpha_{Qm} L^{Qm}
\end{align*}
are polynomials of degrees  $p,P,q$ and $Q$, respectively, $m$ is the seasonality period, $d$ is the number of classical differentiations (equation \ref{differentiation}, Subsection \ref{additive_trend}) and $D$ is the number of seasonal differentiations (equation \ref{m_order_differentiation}, Subsection \ref{additive_seasonality}). 

Note that the {\sc sarima} process has one seasonal component ($\Phi_P,\Theta_Q$) and one nonseasonal component ($\Phi_p, \Theta_q$). Each component is composed of an autoregressive part and of a moving average part. The autoregressive parts $\Phi_P$ and $\Phi_p$ are multiplied together; similarly, the moving average parts $\Theta_Q$ and $\Theta_q$ are multiplied together. For this reason, {\sc sarima} is often referred to as \textbf{multiplicative seasonal {\sc arima}}. In table \ref{special.cases.of.sarima.model}, some specific instantiations of the {\sc sarima} model are presented.
%

\begin{table}[ht]
\scalefont{0.85}
\centering
\begin{tabular}{|l|l|}
\hline
Instantiation                                          & {\sc sarima}$(p,d,q)(P,D,Q)_m$\\
\hline
Seasonal {\sc arma}                          \qquad    &  {\sc sarima}$(0,0,0)(P,0,Q)_m$      \\
{\sc arima}                                  \qquad    &  {\sc sarima}$(p,d,q)(0,0,0)$        \\
Additive trend-seasonality model  	     \qquad    & {\sc sarima}$(p,d,q)(0,D,0)_m$       \\
\hline
\end{tabular}
\caption{Specific instantiations of the {\sc sarima} model. $p$, $q$ and $d$ characterize the nonseasonal component ({\sc nsc}) of the {\sc sarima} model; they respectively denote the {\sc ar} process order for the {\sc nsc} (see Subsection \ref{lar}), the {\sc ma} process order for the {\sc nsc} (see Subsection \ref{ma}) and the number of differentiations required to detrend the {\sc nsc}. $P$, $Q$, $D$ and $m$ characterize the seasonal component ({\sc sc}) of the {\sc sarima} model; $P$, $Q$, $D$ represent for the seasonal component what $p$, $q$ and $d$ represent for the nonseasonal component, except that $m$-order differentiation is considered for the seasonal component, where $m$ stands for the seasonality period.}
\label{special.cases.of.sarima.model}
\scalefont{1.0}
\end{table}


\section{Nonlinear Time Series Models}
\label{non.linear.time.series.models}
Many real-life processes display nonlinear features, such as irregular behavior switching. For such data, linear models (introduced in Section \ref{linear.time.series.models}) unlikely provide an adequate fit to data when a small number of parameters is used, and/or unlikely yield accurate forecasts. Thus, it seems realistic to consider nonlinear models. However, it must be highlighted that underlying nonlinear structures are not necessarily detectable by visual inspection of the raw series. In this case, preprocessing the raw data series is essential to help decipher its structure (see Section \ref{decomposition}). Following the time series model specification already seen in Section \ref{linear.time.series.models} (equation \ref{time.series.model.specification_bis}) and recalled hereafter, adopting a nonlinear framework for the remaining stochastic component $\bm{Z}$ consists in modeling the conditional mean $f$ or/and the conditional standard deviation $g$ as nonlinear functions. We remind the reader that $f$ is the best forecast for $Z_t$ and that $g^2$ is the forecast variance: 


\begin{equation*} 
	Z_t = f(Z_{t-1}, Z_{t-2}, ...) + g(Z_{t-1}, Z_{t-2},...)\ \epsilon_t,
\end{equation*} 

with $\epsilon_t$'s i.i.d.$(0,1)$ white noise processes, $f$ and $g^2$ respectively the conditional mean and conditional variance of $X_t$, meaning that $Z_t | Z_{t-1}, Z_{t-2},.. \sim \mathcal{D}(f, g^2)$ with $\mathcal{D}$ the law of residuals  $\epsilon_t$'s.

In contrast to linear models, autocorrelation is difficult to characterize in nonlinear models. Besides, adopting a nonlinear framework increases the complexity of the model learning process.

The present section will present five categories of parametric nonlinear models proposed in the literature. These models are the following:
\begin{itemize}
\item Polynomial Autoregressive Model ({\sc par})
\item Functional-coefficient Autoregressive Model ({\sc far})
\item Markov Switching Autoregressive Model ({\sc msar})
\item Smooth Transition Autoregressive Model ({\sc star})
\item Autoregressive Conditional Heteroscedasticity ({\sc arch}).
\end{itemize}	
%

\subsection{Polynomial Autoregressive Model - \textsc{par}$(q,p)$}
\label{par}
In the standard linear autoregressive model (equation \ref{lar.process}, Subsection \ref{lar}), the conditional mean $f$ is linear with respect to both model parameters and past values. This latter assumption is released by polynomial autoregressive models in which nonlinear dependencies are specified through polynomials \citep{karakucs-kuruoglu-altinkaya_2017_jour-wind-speed-power-prediction_poly-autoregress-model}. Thus, in the {\sc par}$(q,p)$ model, $f$ is a $q$-degree polynomial function of the $p$ past values, and the conditional variance $g^2$ is constant and equal to $\sigma^2$. A stochastic process $\{X_t\}$ follows a {\sc par}$(q,p)$ model if and only if
\begin{align}
\label{par.process}
Z_t = \mu + \sum_{i_1 = 1}^{p} \phi_{i_1}^{(1)} Z_{t - i_1} + \sum_{i_1 = 1, i_2=1}^{p} \phi_{i_1, i_2}^{(2)} Z_{t - i_1} Z_{t - i_2} + ... + \sum_{i_1=1, ..., i_q=1}^p \phi_{i_1, .., i_q}^{(q)} Z_{t- i_1} ... \,\ Z_{t- i_q} + \sigma\ \epsilon_t,
\end{align}
%
where the error terms $\epsilon_t$'s are i.i.d.$(0, 1)$ such that $\epsilon_t$ is independent of $X_{t-i}$ for $i > 0$, $q$ is the degree of nonlinearity, $p$ is the autoregressive order and $\phi$'s are autoregressive coefficients.

The {\sc par}$(q=1,p)$ model is identical to the {\sc ar}$(p)$ model, and for $p=q=0$, we obtain a simple white noise process.

 
As suggested by equation (\ref{par.process}), a {\sc par} process relies on a \textbf{Volterra series expansion}, the equivalent of Taylor series functional expansion in the nonlinear framework: the Volterra series differs from the Taylor series in its ability to capture memory effects. {\sc par} models have been successfully used in real-life phenomenon modeling such as in industry \citep{gruber_bordons_bars_et_al_2010_journ-robust-nonlinear-control_volterra-model-app-pilot-plant}, short-term wind speed prediction \citep{lee_2011_ieee-power-energy_volterra-model-appl-wind-farm-output-predic}, biological systems \citep{lahaye_poline_flaudin_et_al_2003_journ-neuroimage_volterra-model-app-biological-system}, seismology \citep{bekleric_2008_master-thesis_volterra-model-app-seismology} and communications \citep{fernandes_mota_favier_2010_journ-learning-nonlinear-models_volterra-model-app-communication}. 

\subsubsection{Model identification and parameter learning}


So far, we have seen that the autocorrelation function is used for the identification of linear models. For instance, in the Moving Average Model ({\sc ma}(q), Section \ref{ma}), the autocorrelation function is null as from lagged value $q+1$. The identification of nonlinear models requires other strategies, which are more complex, since the autocorrelation function is not characterizable in such models.

Regarding {\sc par} model dimension estimation, several values of $(q,p)$ can be considered. Then model selection criteria can be used to identify the final values. This procedure can be costly when a large grid of $(q,p)$ values is tested. \citet{karakucs_kuruo_altinkaya_2015_conf-european-signal-proc_estimat-nonlinearty-degree-polyn-autoregress-model} described an alternative method. In this method, $(q,p)$ as well as {\sc par} parameters ($\phi$'s and $\sigma$) are random variables for which prior distributions are defined. The posterior densities of these variables are derived through the well-known Bayes theorem. Then a Reversible Jump Markov Chain Monte Carlo ({\sc rjmcmc }) sampler is used to obtain an estimation of $(q,p)$ and of the model parameters that maximizes the posterior joint density of the variables. 

An interesting and useful feature of {\sc par} models is their linearity with respect to their parameters. Thus, for fixed $(q,p)$ hyper-parameters, model parameters can be estimated by the \textit{nonlinear least squares} ({\sc nls}) method \citep{kuruoglu_2002_journ-digital-signal-proc_polynomial-ar-nonlinear-least-lp-estimation}. {\sc nls} provides the optimal estimates in the maximum likelihood sense if Gaussian error terms are used ($\epsilon_t \sim \mathcal{N}(0,1)$). When $\epsilon_t$'s have heavier tails than in a Gaussian distribution, such as in $\alpha$-stable or generalized Gaussian distributions, the estimate of variance is not reliable. In this case, alternative techniques such as \textit{nonlinear least $L_p$-norm} estimation can be used \citep{kuruoglu_2002_journ-digital-signal-proc_polynomial-ar-nonlinear-least-lp-estimation}.
			
\subsubsection{Forecasting}

Once {\sc par}$(q,p)$ model parameters have been estimated from observations $\{z_t\}_{t=1}^T$, one-step ahead prediction is performed as follows:
%
\begin{align}
\label{star.process.forecasting}
	\hat{Z}_{T+1}  = f(z_{T}, ..., z_{T+1-p}) = &\hat{\mu} + \sum_{i_1 = 1}^{p} \hat{\phi}_{i_1}^{(1)}\ z_{T+1 - i_1} + \sum_{i_1 = 1, i_2=1}^{p} \hat{\phi}_{i_1, i_2}^{(2)}\ z_{T+1 - i_1}\ z_{T+1 - i_2} + ... \\
                       &+ \sum_{i_1=1, ..., i_q=1}^p \hat{\phi}_{i_1, .., i_q}^{(q)}\ z_{T+1- i_1} ... \,\ z_{T+1- i_q}, \nonumber
\end{align}                                                                                            
where $f$ is the conditional mean. \newline
In multi-step ahead predictions, past predicted values are used as new observations.	

\subsection{Functional-coefficient Autoregressive Model - \textsc{far}$(p,k)$}
\label{far}
The functional-coefficient autoregressive model is a direct generalization of the standard linear autoregressive model (equation \ref{lar.process}, Subsection \ref{lar}) in which the autoregressive coefficients are functions instead of constants. This class of models is devoted to exploring the nonlinear features of time series data by exploiting their local characteristics. \citet{chen_tsay_1993_journ-american-stat-assoc_functional-autoregress-model} defined the \textsc{far} model as
\begin{align}
\label{far.process}
	Z_t = \sum_{i=1}^p \phi_i(\mathbf{Z}_{t-i}^*)\ Z_{t-i} + \sigma_t\ \epsilon_t,
\end{align}
where $\{\epsilon_t\}$ is a sequence of i.i.d.$(0, 1)$ error terms such that $\epsilon_t$ is independent of $Z_{t-i}$ for $i>0$, $p$ is the autoregressive order, and the $\phi_i(\mathbf{Z}_{t-i}^*)$ terms are mesurable functions from $\mathbb{R}^k$ to $\mathbb{R}$ and
\begin{align}
\label{far.coef.def}
 \mathbf{Z}_{t-i}^* &= (Z_{t-i_1}, Z_{t-i_2}, ..., Z_{t-i_k}) \quad \text{with} \quad i_j >0 \quad \text{for} \quad j = 1, ..., k.
\end{align}
$\mathbf{Z}_{t-i}^*$ is referred to as the threshold vector, with $i_1, ..., i_k$ the threshold lags and $Z_{t-i_j}$ the threshold variables. 

When the $\phi_i$ terms are constant, the {\sc far} model is reduced to an {\sc ar} model. The {\sc far} model is very flexible and can accommodate most nonlinear features. As a matter of fact, many nonlinear models proposed in the literature are specific cases of the {\sc far} model. For instance, for $\phi_i(\mathbf{Z}_{t-i}^*) = a_i + b_i \exp(-c_i\ z_{t-d}^2)$, equation (\ref{far.process}) is reduced to an \textbf{exponential autoregressive} ({\sc expar}) model \citep{xu_ding_yang_2019_journ-nonlin-dynamics_expon-autoregress-model-expar}. With $\phi_i(\mathbf{Z}_{t-i}^*) = a_i^{(1)} \, \mathbbm{1} (Z_{t-d} \le c) + a_i^{(2)} \, \mathbbm{1} (Z_{t-d} > c)$ where $\mathbbm{1}$ is the indicator function, equation (\ref{far.process}) is reduced to a \textbf{threshold autoregressive} ({\sc tar}) model \citep{chan_yau_zhang_2015_journ-econometrics_lasso-estimation-of-threshold-ar-models,hansen_2011_stat-and-its-interface_threshold-ar-in-economics}. 

	
The definition of {\sc far} models (equation \ref{far.process}) is flexible. However, only parsimonious models are considered in real-life applications: a small number of threshold variables, for instance $k=1,2$, and a low {\sc ar} order are expected.
	
\subsubsection{Model identification and parameter learning}

A {\sc far} model is identified by the {\sc ar} order $p$, the threshold lags $i_1, ...,i_k$ and the functional forms of the $\phi_i$ terms. To set the value of $p$, we may test several values and use model selection criteria, such as the Akaike's information criterion ({\sc aic}). Threshold lags may be identified from the exploration of the nonlinear feature of the data. To this end, various threshold nonlinearity tests can be used \citep{tsay_1989_journ-americ-stat-assoc_testing-modeling-threshold-autoregres-models,so_chen_chen_2005_journ-forecasting_bayesian-threshold-nonlinearity-test,gospodinov_2005_journ-financial-economet_threshold-nonlinearity-test}. Several functional forms can be tested for each of the $\phi_i$ terms; then the best one is selected using model selection criteria, in a manner similar to {\sc ar}-order $p$ identification (see Subsection \ref{lar.parameter.learning}). For $k=1$, an alternative data-driven procedure based on \textit{arranged local regressions} has been proposed to select the functionals forms of $\phi_i$'s \citep{chen_tsay_1993_journ-american-stat-assoc_functional-autoregress-model}. To note, in the {\sc expar} and {\sc tar} models, a unique functional form is shared by all coefficients.


We remind the reader that the $\phi_i$ terms are parametric functions of shape-parameters (for instance, $a_i$, $b_i$, and $c_i$ in the case of {\sc expar}). Once the {\sc far} model has been identified, the $\phi_i$'s can be estimated by the conditional least squares method \citep{klimko_nelson_1978_annals-of-stat_conditional-least-squares}.
	
\subsubsection{Forecasting}

Once {\sc far}$(p,k)$ model parameters have been estimated from observations $\{z_t\}_{t=1}^T$, one-step ahead prediction yields
%
\begin{align}
\label{star.process.forecasting}
	\hat{Z}_{T+1}  &= f(z_{T}, ..., z_{T+1-p}) = \sum_{i=1}^p \hat{\phi}_i(\mathbf{z}_{T+1-i}^*)\ z_{T+1-i},
\end{align}
where $f$ is the conditional mean and $\mathbf{z}_{T+1-i}^* = (z_{T+1-i_1}, z_{T+1-i_2}, ..., z_{T+1-i_k}) \quad \text{with} \quad i_j >0 \quad \text{for} \quad j = 1, ..., k$.
In multi-step ahead predictions, past predicted values are used as new observations.

\subsection{Markov Switching Autoregressive Model - \textsc{msar}$(q,p)$}
\label{msar}
Many real-life phenomena such as in economic systems \citep{hamilton_1989_econometrica_nonstationary-timeseries-markov-switching-ar,hamilton_1990_journ-econometrics_time_series-regime-changes} and meteorogy \citep{ailliot_2015_jour-statist-planning-inferen_non-homog-hidden-markov-switching-models} are subject to switches in regimes. Intuitively, a stochastic process $\mathbf{Z}$ is subject to switches in regimes if the behavior of the associated dynamic system changes at each break which coincides with the beginning of a regime. The break points are not directly observed and probabilistic inference is required to determine whether and when break points occurred, based on the observed time series. 

Let $\mathbf{S}=\{S_t\}_{t=1}^{\infty}$ a homogeneous Markov process with $S_t$ the regime under which $\mathbf{Z}$ is running at time step $t$. $\mathbf{S}$ is referred to as the \textit{state process} of $\mathbf{Z}$. For example, in the economic growth model described in \citep{hamilton_1989_econometrica_nonstationary-timeseries-markov-switching-ar} $S_t$ can take two values: $1$ and $0$, respectively, for fast and slow growths. However, in practice, it is not always possible to provide a physical interpretation for the hidden states. From now on, we will indifferently use {\it regime} or {\it state}, as well as {\it regime switching} or {\it transition}.
	

Regime switching is a nonlinear feature that must be handled by dedicated time series models. In this context, the Markov switching autoregressive model has been proposed by \cite*{hamilton_1990_journ-econometrics_time_series-regime-changes}, to offer a flexible alternative to the popular {\sc arma} and {\sc arima} frameworks. In the {\sc msar} model, the regime shifts are explicitly introduced through a hidden variable. The changes in regime are governed by a homogeneous Markov process ({\sc hmp}), where the transitions are time-independent. The bivariate process is modelled as




%
\begin{align}
\label{msar.process}
	&Z_t|_{S_t=s}  = \phi_0^{(s)} + \sum_{i=1}^{p} \phi_i^{(s)} Z_{t-i}  + \sigma^{(s)}\ \epsilon_t,\\
	&S_t| S_{t-1}, ..., S_{t-q} \sim  \textsc{hmp}(q,K),
\end{align}
%
where the $\epsilon_t$'s are i.i.d. Gaussian white noise processes,  $\mathbf{S}$ is a homogeneous Markov process of order $q$ with $K$ states ($K \ge 1$), $\{\phi_0^{(s)}, ..., \phi_p^{(s)}\}$ and $\sigma^{(s)}$ are respectively the autoregressive parameters and standard deviation for each state $s$, $s = 1, ..., K$. The transition between any two successive regimes only depends on the $q$ previous regimes (\textit{Markov} process). Moreover, given the current regime/state $S_t$, $\mathbf{Z}$ follows an {\sc ar}$(p)$ model whose parameters are specific to $S_t$.

Obviously, the conditional mean $f$ and conditional variance $g^2$ of a {\sc msar} process also switch over time. Thus, given $S_t=s$ with $s=1 \in \{1,2,...,K \}$, $f_{(s)}(Z_{t-1}, ..., Z_{t-p}) = \phi_0^{(s)} + \sum_{i=1}^{p} \phi_i^{(s)} Z_{t-i}$ and $g^{2}_{(s)}(Z_{t-1}, ..., Z_{t-p}) = \sigma^{2}_{(s)}$.

\begin{figure}[t]
\begin{center}
	\includegraphics[width=7cm]{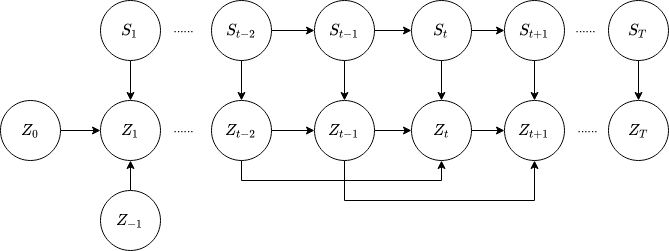}
\end{center}
\caption{The conditional independence graph of the Markov switching autoregressive model for $q=1, p=2$.}
\label{msar.conditional.ind.graph}
\end{figure}

	
When $K=1$, the {\sc msar}$(q,p)$ model is reduced to the {\sc ar}$(p)$ model. When $q=1$, $\mathbf{S}$ is a Markov chain. The conditional independence graph of the {\sc msar}($q=1,p=2$) model is shown in Figure \ref{msar.conditional.ind.graph}: therein, the state at time step $t$ only depends on the state at immediate past time step $t-1$; morever, the value of $Z_t$ depends on $Z_{t-2}$ and $Z_{t-1}$ (the autoregressive part of {\sc msar}), as well as on the state at time step $t$ (the switching part of {\sc msar}).
	
\subsubsection{Model identification and parameter learning}

In practice, $q$ is usually set to $1$ and $p$ is expected to be small. The number of states $K$ and the process order $p$ can be fixed using expert knowledge or they can be chosen using a model selection criterion such as Bayesian information criterion ({\sc bic}) or Akaike's information criterion ({\sc aic}). 

A homogeneous Markov process is parameterized by: (i) the initial state law $\{\pi_k\}_{k=1,...,K}$ with $\pi_k$ the probability for $\mathbf{S}$ to be in state $k$ at initial time step; and (ii) the transition probabilities $P(S_t=s|S_{t-1}=s_1, ..., S_{t-q}=s_q)$ which is reduced to a transition matrix $\{a_{i,j}\}_{i,j=1,...,K}$ when $q=1$. As the state process $\mathbf{S}$ is unobserved, {\sc msar} parameter learning is performed through the \textbf{Expectation-Maximization} algorithm \citep{hamilton_1990_journ-econometrics_time_series-regime-changes}.
	
\subsubsection{Forecasting}
After the parameters related to both autoregressive and switching parts of the the {\sc msar}$(q,p)$ model have been estimated from observations $\{z_t\}_{t=1}^T$, one-step ahead prediction (in the sense of mean square errors) is achieved as follows \citep{clements-krolzig_1998_the-economet-j_compar-forecast-perf-msar-thresh-ar}:
%
\begin{align}
\label{star.process.forecasting}
	\hat{Z}_{T+1}  &= \mathbb{E}[Z_{T+1}\,|\, Z_{1}, \dots, Z_T; \, \hat{\theta}] = \sum_{s=1}^K  P(S_{T+1}=s \,|\, Z_{1}, \dots, Z_T; \, \hat{\theta}) \times f_{(s)}(z_{T}, ..., z_{T+1-p}),
\end{align}
with $\hat{\theta}$ the set of estimated parameters. \\
Probabilities $P(S_{T+1}=s \,|\, Z_{1}, \dots, Z_T; \, \hat{\theta})$ are entirely determined based on the \textbf{smoothed probabilities} $\gamma_t(s') = P(S_t=s' \,|\, Z_{1}, \dots, Z_t; \, \hat{\theta})$ and the transition probabilities $\{a_{i,j}\}_{i,j=1,...,K}$:
$$ P(S_{T+1}=s \,|\, Z_{1}, \dots, Z_T; \, \hat{\theta}) = \sum_{s'=1}^K \gamma_T(s') \times a_{s',s}. $$
The \textbf{smoothed probabilities} $\gamma_t(s') = P(S_t=s' \,|\, Z_{1}, \dots, Z_t; \, \hat{\theta})$ are computed through the \textit{forward-backward} algorithm \citep{baum-petrie-soules_1970_jour-annals-math_em-for-hmc-models}.
In multi-step ahead predictions, past predicted values are considered as new observations, and the probabilities of $S_{t+h}, h > 1$ are calculated using those of $S_{t+h-1}$ and the transition probabilities between states.

\subsubsection{Variants of the {\sc msar} model}
Some variants of the {\sc msar} model have been proposed in the literature. \citet{bessac_ailliot_cattiaux_et_al_2016_advanc-stat-climat-met-ocean_hidd-obs-regime-switch-ar-model} depict a variant where the state process $\mathbf{S}$ is directly observed. This specific model is referred to as \textbf{observed Markov switching {\sc ar}}. \citet{ailliot_prevosto_soukissian_et_al_2003_conference_gamma-markov-switching-autoregress} described an {\sc msar} model in which Gamma white noise is used instead of the usual Gaussian white noise. This adaptation was motivated by a better fit to data (in this case, wind speed for a given month, across several years). Further, Ailliot and other co-authors described an {\sc msar} in which $\mathbf{S}$ is a nonhomogeneous Markov process, that is transition from state to state is time-dependent \citep{ailliot_2015_jour-statist-planning-inferen_non-homog-hidden-markov-switching-models}. More recently, variants of the {\sc msar} model proposed to integrate partial knowledge about $\mathbf{S}$. \citet{juesas-ramasso-drujont_2021_arxiv_msa-hmm-part-knowl} use \textit{belief functions} to model \textit{prior} probabilities about $\mathbf{S}$. \citet{dama-sinoquet_2021_ictai_phmc-lar-mach-health-diagn} supposed $\mathbf{S}$ known (observed) at some random time steps, and hidden at the remaining ones. 

\subsection{Smooth Transition Autoregressive Model - \textsc{star}$(p)$}
\label{star}
The smooth transition autoregressive model is another extension of the linear {\sc ar} model that allows transitions from an {\sc ar} model to another \citep{lin_terasvirta_1994_journ-econometrics_smooth-transition-autoregres,eitrheim_terasvirta_1996_journ-econometrics_smooth-transition-autoregres-adequacy-test}. It is a nonlinear state-dependent dynamic model like the Markov switching autoregressive model {\sc msar} (Subsection \ref{msar}). In other words, for both {\sc msar} and {\sc star} models, alternative {\sc ar} dynamics are allowed. The substantial difference between {\sc star} and {\sc msar} models lies in how the autoregressive parameters change over time. The switching part of an {\sc msar} process switches between a finite and relatively small number of {\sc ar} processes; in this case transitions are said to be abrupt. In contrast, in a {\sc star} model, the autoregressive parameters depend on a transition function which is generally continuous. Thus, the probability to observe exactly the same dynamics at two distinct time steps is null. Therefore, the process switches between an infinite number of {\sc ar} processes (one per time step). These transitions are said te be smooth.

The {\sc star} model is defined as
\begin{align}
\label{star.process}
	Z_t = \phi_0^{(1)} + \sum_{i=1}^p \phi_i^{(1)}\ Z_{t-i} + G(S_t) \left(\phi_0^{(2)} + \sum_{i=1}^p \phi_i^{(2)}\ Z_{t-i}\right)  + \epsilon_t,
\end{align}
where $\{\epsilon_t\}$ is a sequence of i.i.d.$(0,\sigma^2)$ error terms, $p$ is the {\sc ar} order, $\{\phi_0^{(j)}, ..., \phi_p^{(j)}, \sigma^2 \}$ with $j=1,2$ is a vector of parameters to be estimated, $S_t$ is the \textit{transition variable} and $G$ is the \textit{transition function} which is bounded between zero and one.

From equation (\ref{star.process}), it is easily seen that $\mathbf{Z}$ switches from one {\sc ar} process to another one with autoregressive coefficients varying according to transition function $G$. Thus, {\sc star} model dynamics are altered by the transition function, conditional on the transition variable, in a potentially smooth manner. The degree of smoothness depends on how $G$ is modeled. Given the transition variable $S_t$, the conditional mean $f_{(S_t)}(Z_{t-1}, ..., Z_{t-p}) = (\phi_0^{(1)} + G(S_t) \phi_0^{(2)}) + \sum_{i=1}^p Z_{t-i}(\phi_i^{(1)} + G(S_t) \phi_i^{(2)}) $ is state-dependent, whereas the conditional variance $g^2$ is constant and equal to $\sigma^2$.  

It is important to emphasize that structural changes in regimes are not incompatible with stationarity. If dynamic changes are local phenomena in the process behavior, then long-run statistics (mean and variance) of the process may remain stable. This case has been observed with large-scale medium-frequency events known as the so-called El Ni$\tilde{n}$o Southern Oscillations ({\sc enso}) \citep{ubilava_helmesr_2013_environ-modeling-softwa_smooth-transition-autoregres-model}.


\subsubsection{Choice of transition function and transition variable - parameter learning}

Transition functions define how the autoregressive parameters vary over time. The most frequently used transion functions are the following:
\begin{align}
\label{star.logis.trans.func}
	G(S_t) &= \{ 1 + \exp[ -\xi (S_t -  c) ] \}^{-1} \qquad \qquad \qquad \qquad \text{logistic transition function},\\
\label{star.exp.trans.func}
	G(S_t) &= \{ 1 - \exp[ -\xi (S_t - c)^2] \} \hspace{2mm} \qquad \qquad \qquad \qquad \text{exponential transition function},\\
\label{star.quad.trans.func}
	G(S_t) &= \{ 1 + \exp[ -\xi (S_t^3 + c_1 S_t^2 + c_2S_t + c_3 ] \}^{-1} \hspace{2,5mm} \quad \text{cubic transition function},	
\end{align}
where $\xi \ge 0$ is the smoothness parameter determining the smoothness of transitions, and $c, c_1, c_2, c_3$ are other shape parameters called location parameters. 
The transition function in (\ref{star.logis.trans.func}) allows a smooth monotonic parameter change with a single structural break for the limiting case $\xi \longrightarrow \infty$. Function (\ref{star.exp.trans.func}) mimics a nonmonotonic change which is symmetric around $c$. Equation (\ref{star.quad.trans.func}) describes the most flexible transition function which allows both monotonic and nonmonotonic changes. When $\xi \longrightarrow 0$, all transition functions (\ref{star.logis.trans.func}) to (\ref{star.quad.trans.func}) tend to a constant and the {\sc star} model converges to an {\sc ar} model.

Generally, the transition variable $S_t$ is set as a lagged value of $Z_t$, such as $S_t = Z_{t-d},$ with $d > 1$ \citep[see for example][]{ubilava_helmesr_2013_environ-modeling-softwa_smooth-transition-autoregres-model}. In this case, the {\sc star} model is a specific instantiation of the functional autoregressive model (equation \ref{far.process}, Subsection \ref{far}). Alternatively, the literature dedicated to parameter stability or structural change indicates that $S_t$ is a function of time, usually $S_t=t$. The reader interested in further details is referred to the works developed by \citet{lin_terasvirta_1994_journ-econometrics_smooth-transition-autoregres} on the one hand, and by \citet{eitrheim_terasvirta_1996_journ-econometrics_smooth-transition-autoregres-adequacy-test} on the other hand.


Once $G$, $S_t$ and $p$ have been set, {\sc star} model parameters can be estimed using a nonlinear optimization procedure. Autoregressive order $p$ can be chosen by optimizing a model selection criterion.\newline

\subsubsection{Forecasting}
After the {\sc star}$(p)$ model parameters $\{\phi_0^{(j)}, ..., \phi_p^{(j)}, \sigma^2 \}$ with $j=1,2$, and the shape parameters of the transition function $G$ have been estimated from observations $\{z_t\}_{t=1}^T$, one-step ahead prediction is performed as follows:
%
%
\begin{align}
\label{star.process.forecasting}
	\hat{Z}_{T+1}  &= f_{(S_{T+1})}(z_{T}, ..., z_{T+1-p}) = (\hat{\phi}_0^{(1)} + \hat{G}(S_{T+1}) \, \hat{\phi}_0^{(2)}) + \sum_{i=1}^p z_{T+1-i}\ (\hat{\phi}_i^{(1)} + \hat{G}(S_{T+1}) \, \hat{\phi}_i^{(2)}).
\end{align}

As usual, past predicted values are used as new observations in multi-step ahead predictions.

\subsection{Autoregressive Conditional Heteroscedasticity - \textsc{arch}$(p)$}
\label{garch}
%
In the course of a forecasting task, the ability to make accurate prediction may vary from one period to another. Therefore, the uncertainty associated with different forecast periods represented by the forecast variance $g^2$ may change over time. However, most time series models introduced in the literature assume a constant forecast variance $g^2$ (this is the case of all models presented so far in Sections \ref{linear.time.series.models} and \ref{non.linear.time.series.models}). To release this constraint and obtain reliable forecast intervals $[f-g, f+g]$, \cite*{engle_1982_econometrica_arch-model} has introduced the {\sc arch} model. 

In the standard {\sc arch} model, the conditional mean $f$ is equal to $0$ (hence a zero unconditional mean) and the forecast/conditional variance $g^2$ (equation \ref{time.series.model.specification_bis}, Section \ref{linear.time.series.models}) is modeled as a quadratic function of the past $p$ values.
A zero-mean process $\{Z_t\}$ follows a standard {\sc arch} model if and only if
\begin{align}
\label{arch.observation.eq}
	Z_t      &= \sigma_t\ \epsilon_t,\\
\label{arch.state.eq}
\sigma^2_t |Z_{t-1}, ...,Z_{t-p} &= \phi_0 + \sum_{i=1}^{p} \phi_i\ Z_{t-i}^2,
\end{align}
where $\epsilon_t$'s are i.i.d.(0,1) white noise processes (usually, the normal law is used), $\sigma_t^2$ is the conditional variance, $\{\phi_0, \phi_1, ..., \phi_p\}$ are the model parameters and $p$ is the {\sc arch} process order.

Regarding {\sc arch} process identification, \cite*{engle_bollerslev_1986_economet-reviews_persistence_condit_variance} proposed a statistical test that checks the evidence of {\sc arch}$(p)$ for a given time series data. Once the {\sc arch} order has been fixed, the model parameters can be estimated through maximum likelihood maximization \citep{engle_1982_econometrica_arch-model}. 

An {\sc arch} process is said to be \textit{volatile} since the {\it conditional variance} is time-varying while the {\it unconditional variance} $\text{Var}(Z_t)$ is assumed constant over time. Volatility is exhibited in many financial time series. When $\phi_0 > 0$, $\phi_1, ..., \phi_p \ge 0$, a large value observed for $Z_t$ will result in large \textit{volatilities} (large conditional variances) for the next $p$ time steps. Similarly, a small value observed for $Z_t$ will lead to small volatilities for the next $p$ time steps. Thus, an {\sc arch} process tends to cluster large (respectively small) conditional variances. When it is not taken into account in time series models, the {\it volatility phenomenon} can lead to inaccurate predictions.
	
In Section \ref{lar}, we have examined conditions to guarantee the stationarity for autoregressive processes. Similar conditions have been derived for the {\sc arch} model \citep[Theorem 2]{engle_1982_econometrica_arch-model}: a standard {\sc arch}$(p)$ process with $\phi_0 > 0$, $\phi_1, ..., \phi_p \ge 0$ is stationary if and only if all roots of the associated characteristic equation are outside the unit circle ({\it i.e.}, all roots have their absolute values greater than or equal to $1$).

Other formulations of the conditional variance $g^2$ may be considered, such as exponential and absolute value forms. To simplify the parameter estimation procedure, \cite*{engle_1982_econometrica_arch-model} has suggested to choose $g^2$ symmetric, strictly positive and regular. The regularity conditions can be found in \citep{engle_1982_econometrica_arch-model}.
	
\subsubsection{Generalized {\sc arch} model - {\sc garch}($q,p$)}
The {\sc garch} model is an extension of the {\sc arch} model in which the conditional variances $\sigma_t^2$'s are autocorrelated \citep{bollerslev_1986_journ-econometrics_garch-model}. In the {\sc garch} process, equation (\ref{arch.state.eq}) becomes
%
\begin{align}
\label{garch.process}
	\sigma^2_t |Z_{t-1}, ...,Z_{t-p}, \sigma_{t-1}^2, ..., \sigma_{t-q}^2 &= \phi_0 + \sum_{i=1}^{p} \phi_i\ Z_{t-i}^2 + \sum_{j=1}^{q} \alpha_j\ \sigma^2_{t-j},
\end{align}
where $(q,p)$ characterizes the {\sc garch} process orders and $\{\phi_0, \phi_1, ..., \phi_p, \alpha_1, ..., \alpha_q\}$ are parameters with $\phi_0 > 0$ and $\phi_1, ..., \phi_p, \alpha_1, ..., \alpha_q \ge 0$.
	
{\sc arch} model extension to {\sc garch} model is similar to the extension of {\sc ar} to {\sc arma}. Similarly, this extension allows a more parsimonious description of large-order {\sc arch} processes. {\sc garch}($q=0,p$) and {\sc arch}($p$) processes coincide, and {\sc garch}($p=0,q=0)$ amounts to white noise process. {\sc garch}$(p=0,q)$ and {\sc arch}$(p=0)$ processes coincide.


Regarding {\sc garch} process identification, that is the selection of $(q,p)$, \cite*{bollerslev_1986_journ-econometrics_garch-model} designed a procedure relying on the autocorrelation function and partial autocorrelation function for the squared process $\{Z_t^2\}$. To note, a similar procedure had been proposed for {\sc arma} process identification \citep{box_jenkins_reinsel_et_al_2015_book_time_series_analysis}. Besides this informal graphical procedure, a statistical test checking the evidence of {\sc garch}$(q,p)$ for given time series data has been introduced in \citep{bollerslev_1986_journ-econometrics_garch-model}. Once $(q,p)$ have been fixed, as with {\sc arch} model, {\sc garch} model parameters can be estimated through maximum likelihood maximization \citep{bollerslev_1986_journ-econometrics_garch-model}.


It has been proven that {\sc garch}$(q,p)$ process is stationary with $\mathbb{E}[Z_t] = 0, \text{Var}(Z_t) = \phi_0 (1 - \sum_{i=1}^p \phi_i - \sum_{j=1}^q \alpha_j)^{-1}$ and $\text{Cov}(Z_t, Z_{t'}) = 0$ for $t \neq t'$, if and only if $\sum_{i=1}^p \phi_i + \sum_{j=1}^q \alpha_j < 1$ \citep{bollerslev_1986_journ-econometrics_garch-model}.

Finally, the {\sc garch} model has been used in connection with other models. In practice, in models assuming a constant forecast variance, this assumption has been relaxed by incorporating residuals modeled through a {\sc garch} process. Thus, better forecast intervals are obtained. For instance, {\sc arma} (equation \ref{arma.process}) and {\sc arima} (equation \ref{arima.process}) models have been supplemented with a {\sc garch} modeling of the residuals \citep{pham_yang_2010_mechanical-systems-signal-proces_arma-garch-forecasting,xin_Zhou_yang_et_al_2018_sensors_garch-kalman-arima-prediction}.


\subsubsection{Forecasting}
Let $\{\hat{\phi}_0, \hat{\phi}_1, ..., \hat{\phi}_p, \hat{\alpha}_1, ..., \hat{\alpha}_q\}$ the {\sc garch} model parameters estimated from observations $\{z_t\}_{t=1}^T$. One-step ahead prediction is achieved as follows:
\begin{align}
	\hat{\sigma}^2_{T+1}  &= g^2(z_{T}, \dots, z_{T+1-p}) = \hat{\phi}_0 + \sum_{i=1}^{p} \hat{\phi}_i\ z_{T+1-i}^2 + \sum_{j=1}^{q} \hat{\alpha}_j\ \hat{\sigma}^2_{T+1-j},\\
	\hat{Z}_{T+1}   &\sim \mathcal{D}(0, \hat{\sigma}^2_{T+1}),\nonumber
\end{align}

where $\hat{\sigma}_{T}, \hat{\sigma}_{T-1}, ..., \hat{\sigma}_{T+1-q}$ are computed from equation (\ref{garch.process}) and $\mathcal{D}$ is the chosen distribution for white noise errors. \newline
In multi-step ahead predictions, past predicted values are used as new observations.
	
\subsubsection{Other extensions of the {\sc arch} model}
The Markov-{\sc arch} model \citep{cai_1994_journ-business-economic-stat_regime-switch-arch} combines Hamilton's switching-regime model \citep[{\sc msar}, equation \ref{msar.process},][]{hamilton_1990_journ-econometrics_time_series-regime-changes} with the {\sc arch} model. The motivation behind this extension of {\sc arch} is to address the issue of \textit{volatility persistence}, or constancy of volatility during a relatively long period of time. The Markov-{\sc arch} model handles this issue by allowing occasional shifts in the conditional variance $g^2$ governed by a Markov process. Namely, {\sc arch} movements occur within "regimes", with occasional jumps occurring between two successive "regimes" characterized by different conditional variances. In the same line, further works have described {\sc ms}-{\sc cgarch}, the Markov switching component {\sc garch} model, in which volatility is modeled through the combination of two {\sc garch} models \citep{alemohammad_rezakhah_alizadeh_2016_communic-stat-theo-meth_markov-switching-component-garch}. 
Besides, a smooth transition {\sc arch} model was proposed, that combines a smooth transition autoregressive ({\sc star}) model (equation \ref{star.process}, Section \ref{star}) with an {\sc arch} model \citep{hagerud_1996_journ_smooth-transition-arch-model}. In this model, the conditional variance $g^2$ is the one defined in equation (\ref{arch.state.eq}), in which the $\phi_i$ terms depend on the logistic (equation \ref{star.logis.trans.func}) or exponential (equation \ref{star.exp.trans.func}) transition functions and on transition variable $S_t = X_{t-d}$. Finally, stochastic volatility models provide more realistic and flexible alternatives to {\sc arch}-type models \citep{meyer_fournier_berg_2003_the-econometrics-journ_stochas-volati-bayes-automat-differentiat}. Therein, the conditional variance is stochastic, that is $g$ is a nonlinear function of the past values of $\bm{Z}$ plus random error terms.

\section{Deep Learning}
\label{deep_learning}
Amongst nonlinear models, artificial neural networks hold a place apart as deep learning has recently gained considerable attention.


 
To model time series, parametric models informed by domain expertise have been mainly used, such as autoregressive models and exponential smoothing frameworks. Not only did these established statistical models gain their popularity from their high accuracy. They are also suitable for nonspecialists as they are efficient, robust and user-friendly. 

In times series processing, deep neural networks ({\sc dnn}s) hold a distinctive place. Recently, notable achievements have open up an avenue for deep learning. Deep learning is not a restricted learning approach, but it abides various procedures and topographies, to cope with a large spectrum of complex problems \citep{raghu_schmidt_2020_arxiv_survey-deep-learn-scient-discov}. Deep learning relies on {\it deep} artificial neural networks ({\sc ann}s), that is {\sc ann}s with a high number of layers. Deep learning has become an active field of research in the next generation of time series forecasting models. {\sc dnn}s are particularly suitable for finding the appropriate complex nonlinear mathematical function to turn an input into an output. Therefore, deep learning provides  a  means  to learn  temporal dynamics in a purely data-driven manner. In the remainder of this section, we briefly highlight how different classes of {\sc dnn}s may be adapted to achieve the forecasting task in time series. The categories of models selected for illustration are the following:
\begin{itemize}
 \item Multilayer Perceptrons,
 \item Recurrent Neural Networks,
 \item Long Short-Term Memory networks,
 \item Convolutional Neural Networks,
 \item Transformers.
\end{itemize}


\subsection{Multilayer Perceptrons}\label{mlp}

The most popular Artificial Neural Network ({\sc ann}) model, the perceptron, is composed of an input layer, hidden layers and output layer. Therein, an output $y$ is computed as a weighted sum over the nodes connected to $y$ in the preceding layer $L$: $y = A\ (\sum_{i \in L}\ x_i\ w_i + b)$, where $x_i$ denotes an input from a node in $L$, $b$ is a bias term, and $A$ is a nonlinear activation function. $A$ triggers the node activation. The most widely-used activation functions are the following:

\begin{itemize}
  \item sigmoid: $\sigma(z) = \frac{1}{1 + e^{-z}}$
  \item hyperbolic tangent: $tanh(z) = \frac{e^z - e^{-z}}{e^z + e^{-z}}$
  \item Rectified Linear Unit (ReLU): $R(z) = max (0,z)$
  \item softmax: $softmax(z_i) = \frac{e^{z_i}}{\sum_j\ e^{z_j}}$, where $z$ is a vector of reals.
\end{itemize}

\subsubsection{Gradient-based learning algorithm}

\begin{figure}[t]
\begin{center}
	\includegraphics[width=7cm]{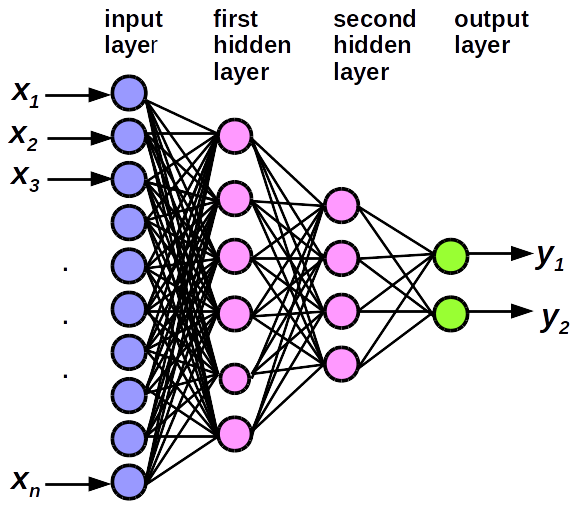}
\end{center}
\caption{Example of Multilayer Perceptron ({\sc mlp}). The number of nodes in the input layer equals the number of features in each data row. Each layer is fully connected to its subsequent layer.}
\label{figure_mlp}
\end{figure}

Beyond single-layer perceptrons, increasing the number of hidden layers in {\it multilayer} perceptrons ({\sc mlp}s) allows to tackle more complex problems (Figure \ref{figure_mlp}). This comes at a higher learning cost. Learning a perceptron is a supervised task that aims at instantiating the weights $w$ of the connections between the layers, and the biaises $b$, to minimize some cost function $C(w,b)$. The {\sc mlp} learning algorithm proceeds by successive improvements. At a current point, a modification is made in the opposite direction to the gradient, so as to decrease the contribution to the cost ($w_{i,j} \leftarrow w_{i,j} - \alpha \frac{\partial C}{\partial w_{i,j}}$; $b{i} \leftarrow b_{i} - \alpha \frac{\partial C}{\partial b_{i}}$), where $\alpha$ is the learning rate. This {\it gradient algorithm} is iterated until cost convergence. However, a neural network  potentially consists in the composition of millions of functions. Therefore, the function modeled to turn the input into the output is not simple and a difficulty lies in the calculation of the different partial derivatives $\frac{\partial C}{\partial w_{i,j}}$ and $\frac{\partial C}{\partial b_{i}}$. 

{\bf Gradient backpropagation} is the cornerstone of {\sc mlp} learning. Since we know the expected result in last layer and the calculations that took us from penultimate to last layer, it is possible to compute the error on the penultimate layer, and so on up to the first layer. Thus, potentially millions of partial derivatives may be computed efficiently through a unique forward and backward pass.

\subsubsection{Enhancing backpropagation performance for deep learning}

It soon quickly turned out that solving more and more complex problems requires an increasing number of layers in neural networks. However, the deeper a neural network is, the more computing power is needed to learn it, and the less the gradient backpropagation algorithm works satisfactorily. Technological and algorithmic advances have triggered the spectacular rise of deep learning in the last ten years: the increasing availability of massive data allows to train deeper models; neural network learning naturally lends itself to mass parallelization {\it via} {\sc gpu}s; a set of new techniques facilitated the use of gradient backpropagation. Moreover, the provision of open-source frameworks is making a significant contribution to the upturn of deep learning, to facilitate backpropagation as well as the customisation of network architectures.

\subsubsection{Forecasting}

In time series forecasting, the data fed to an {\sc mlp} (as well as to any other kind of neural network) must be prepared from a single sequence. The sequence must be splitted into multiple $\{ x=input\ /y=output\}$ patterns (or samples) from which the model can be learned. For example, for a one-step prediction purpose, if three time steps are used for inputs, the time series $[10, 15, 20, 25, 30, 35, 40, 45, 50]$ will be divided into the following samples: $\{x = [10, 15, 20], y = 25\}$, $\{x = [15, 20, 25], y = 30\}$, $\{x = [20, 25, 30], y = 35\} \cdots$. The {\sc mlp} model will map a sequence of past observations as input to an output observation. 

In time series forecasting as in other domains, the emergence of competing neural networks has relegated {\sc mlp}s to the background.
 



\subsection{Recurrent Neural Networks}\label{rrn}

In the previous subsection, we have described how a univariate time series must be preprocessed to feed an {\sc mlp} for one-step prediction purpose. However, two limitations appear when short-term predictions are made from a fixed sized window of inputs (three time steps in the illustration of Subsection \ref{mlp}): (i) the sliding window adds memory to the problem, for a contextual prediction, but defining the window size is challenging as there is no guarantee that sufficient knowledge is brought; (ii) an {\sc mlp} takes as input a feature vector of fixed size, which defeats the purpose of processing series with no prespecified size. 

%
%

\subsubsection{Context-informed prediction} Recurrent neural networks ({\sc rnn}s) were designed to handle sequential information \citep{lipton_berkowitz_elkan_2015_arxiv_review-recur-neur-netwk-seq-learn}. Typically, {\sc rnn}s are suitable to make predictions over many time steps, in time series. An {\sc rnn} achieves the same task at each step (with varying inputs): the sequential sequence ($x_1, x_2, \cdots, x_{t}, x_{t+1}\cdots $) is input to the {\sc rnn}, element by element (one step at a time). 

\begin{figure}[t]
\begin{center}
\begin{tabular}{ccc}
	\begin{subfigure}[b]{0.22\textwidth}
          \begin{center}
        \includegraphics[width=5cm]{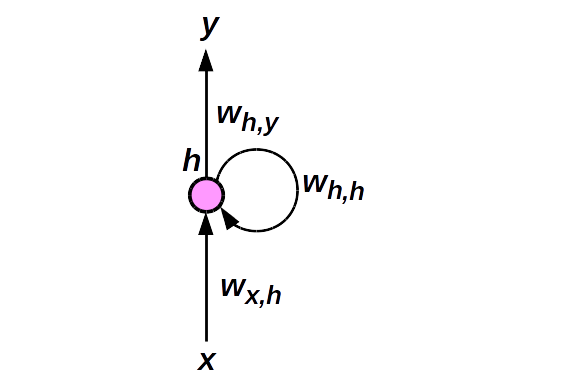}
          \end{center} 
        \end{subfigure}
    &\begin{subfigure}[b]{0.32\textwidth}
       \begin{center}
        \includegraphics[width=\textwidth]{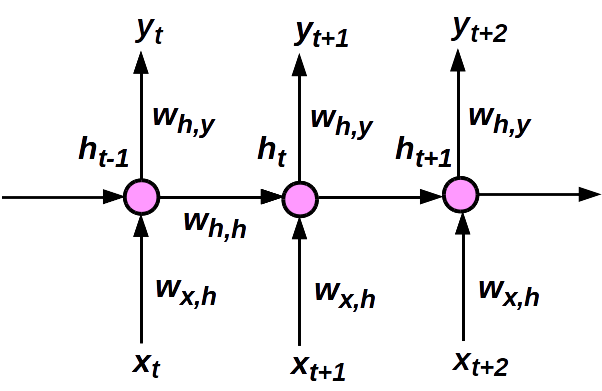}
       \end{center} 
     \end{subfigure}
    & \begin{subfigure}[b]{0.32\textwidth}
        \begin{center}
        \includegraphics[width=6.0cm]{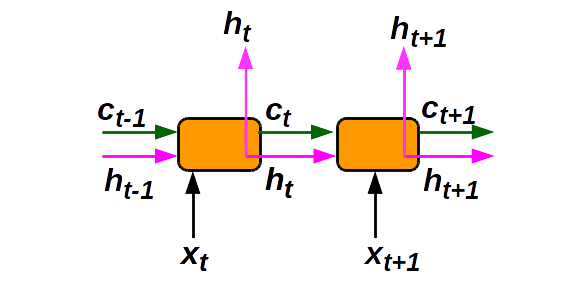}
        \end{center} 
    \end{subfigure}\\
     (a) & (b) & (c)
\end{tabular}
\end{center}
\caption{Examples of Recurrent Neural Network ({\sc rnn}) and Long-Short Term Memory ({\sc lstm}) network. {\bf (a)} {\sc rnn} architecture. $x$, $h$ and $y$ respectively denote input, hidden state (or "memory") and output. At any time step, an element $x_t$ of the time series is fed into the {\sc rnn} module. $h_t$ depends on previous hidden state $h_{t-1}$ and on current value $x_t$. {\bf (b)} Unfolding in time for the forward pass of the {\sc rnn} in subfigure \ref{figure_rnn_lstm} (a). $x_t$: $h_t = f(w_{h,h} h_{t-1} + w_{x,h}\ x_t)$. The nonlinear function $f$ is usually the $tanh$ or $ReLU$ function, and $y_t = \mathrm{softmax}(w_{h,y}\ h_t)$. {\bf (c)} Unfolding in time for an {\sc lstm}. $x_t$: input vector; $h_t$, $c_t$: hidden layer vectors. The {\sc lstm} module models a mathematical function that takes three inputs and yields two outputs: $(h_t,c_t) = f(h_{t-1},c_{t-1},x_t)$, where $h_t,h_{t-1},c_t, c_{t-1} \in [-1,1]$. The two ouputs are fed back into the {\sc lstm} module at time step $t+1$.}
\label{figure_rnn_lstm}
\end{figure}

An {\sc rnn} performs the same task for each element of a sequence, with the output being dependent on the previous computations. Conceptually, an {\sc rnn} can be seen as an {\sc mlp} architecture enriched with loops (Figure \ref{figure_rnn_lstm} (a)). In other words, an {\sc rnn} has a memory to capture information about what has been calculated so far, and its decisions are impacted by what the {\sc rnn} learned from the past. Thus, in an {\sc rnn}, not only are outputs influenced by weights associated with inputs as in standard feedforward {\sc nn}s; a {\it hidden state vector} allows contextual decisions, based on prior input(s) and output(s).

\subsubsection{Backpropagation Through Time}

For a better understanding, Figure \ref{figure_rnn_lstm} (b) provides the {\sc rnn} unfolded on the forward pass and along the input time series $\{x_t\}$. For instance, if the time series was of length $T$, the network would be unfolded into a $T$-layer neural network without cycles. The hidden states capture the dependencies that neighbor datapoints might have with each other in the series. 

We have seen previously that {\sc mlp}s use different parameters at each layer. In constrast, an {\sc rnn} shares the same parameters across all steps ($w_{h,h}$, $w_{x,h}$, $w_{h,y}$, see Figure \ref{figure_rnn_lstm} (b)). This substantially diminishes the number of parameters to learn. However, in a deep learning context, {\sc rnn} learning faces several major difficulties: (i) standard backpropagation cannot be applied as is, due to the loop connections; (ii) exploding or vanishing gradients are likely to generate unstability, thus hampering the reliability of weight updates; (iii) theoretically, {\sc rnn}s can exploit information from arbitrarily long time series, but in practice they are limited to model dependencies only within a few time steps. 

The first issue is addressed using Backpropagation Through Time ({\sc bptt}): the network is unfolded into a standard {\sc mlp} in which nodes involved in recurrent connections are replicated; backpropagation is then applied. Because the parameters are shared by all time steps in an {\sc rnn}, the gradient at each output depends not only on the calculations of the current time step, but also on those of the previous time steps.

The second and third issues are tackled through using a novel class of architectures called Long Short-Term Memory networks, usually just called {\sc lstm}s.

\subsubsection{Forecasting}

Time series forecasting through deep {\sc rnn}s is an active area (see for instance in the financial domain \citealp{chandra_and_chand_2016_appl-soft-comput_time-series-prediction-finance-deep-learn-rnn}; \citealp{hiransha_gopalakrishnan_menon_et_al_2018_procedia-comput-scienc_stock-market-prediction-deep-learn-rnn}). Recently, an extensive empirical study demonstrated that in many cases, {\sc rnn}s outperform statistical methods which are currently the state-of-the art in the community \citep{hewamalage_bergmeir_bandara_2021_int-j-forecast_recurr-neur-netw-time-series-forecast}.

\subsection{Long Short-Term Memory Networks}\label{lstm}
{\sc lstm}s are the most widely-used subclass of {\sc rnn}s, as they perform better than {\sc rnn}s in capturing long dependencies. {\sc lstm}s are intrinsically {\sc rnn}s in which changes were introduced in the computation of hidden states and outputs, using the inputs. 

\subsubsection{Keeping track of dependencies}
\begin{figure}[t]
\begin{center}
\begin{tabular}{cc}
    \begin{subfigure}[b]{0.45\textwidth}
      \begin{center}
        \includegraphics[width=5.7cm]{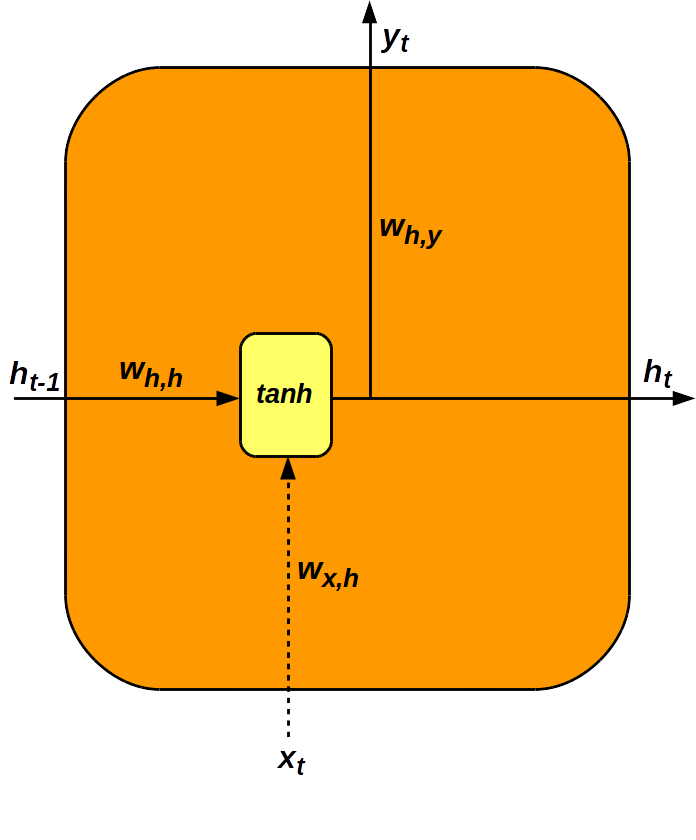} 
      \end{center}
    \end{subfigure}
    & \begin{subfigure}[b]{0.45\textwidth}
         \begin{center}
        \includegraphics[width=6cm]{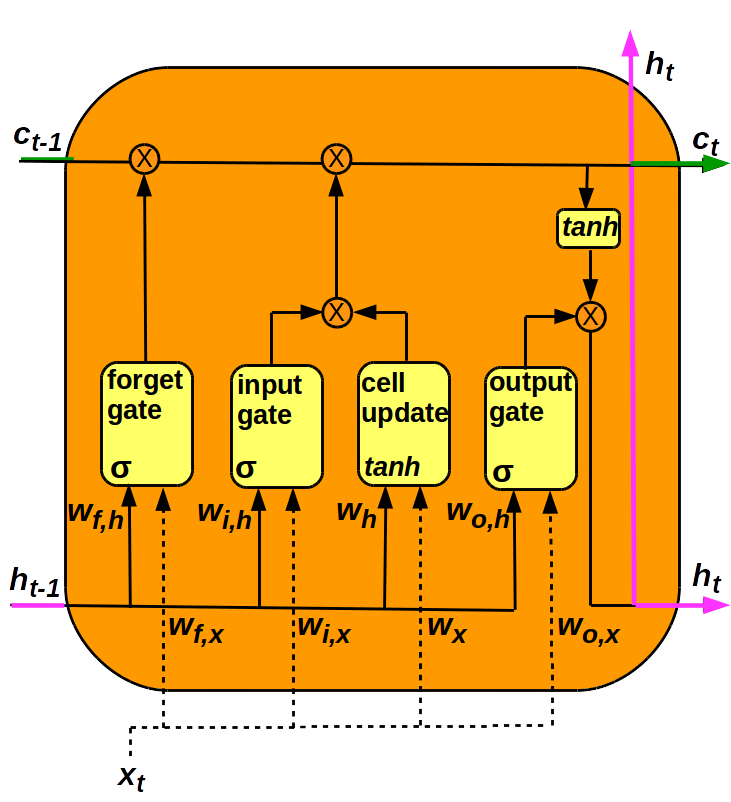}
      \end{center}
    \end{subfigure}\\
    (a) & (b)\\
\end{tabular}
\end{center}
\caption{Illustrations for the repeating modules of a Recurrent Neural Network ({\sc rnn}) and a Long-Short Term Memory ({\sc lstm}) network. {\bf (a)} Standard {\sc rnn}. $h_t = tanh(w_{h,h} h_{t-1} + w_{x,h}\ x_t)$, $y_t = \mathrm{softmax}(w_{h,y}\ h_t)$. {\bf (b)} {\sc lstm}. Notations: $x_t$: input vector; $h_t$, $c_t$: hidden layer vectors; $\sigma$, $tanh$ : activation functions. To regulate the flow of information passing through modules, gates rely on the sigmoid activation function $\sigma$ and on the pointwise multiplication operation ($\otimes$) as follows:\\
$f^g(x_t, h_{t-1}) = \sigma(w_{f,x}\ x_t + w_{f,h}\ h_{t-1} + b_f) \in [0,1]$ (forget gate)\\
$i^g(x_t, h_{t-1}) = \sigma(w_{i,x}\ x_t + w_{i,h}\ h_{t-1} + b_i) \in [0,1]$ (input gate)\\
$o^g(x_t, h_{t-1}) = \sigma(w_{o,x}\ x_t + w_{o,h}\ h_{t-1} + b_o) \in [0,1]$ (output gate)\\
$c^u(x_t, h_{t-1}) = tanh(w_x\ x_t + w_h\ h_{t-1} + b) \in [0,1]$ (cell update),\\
where $w_{f,x}, w_{i,x}, w_{o,x}, w_{f,h}, w_{i,h}, w_{o,h}, w_x, w_h$ and $b_f, b_i, b_o, b \in \mathbb{R}$ are respectively weight and biais parameters.}
\label{figure_rnn_lstm_modules}
\end{figure}

We have seen that the {\sc rnn} architecture is described as a chain of repeating modules. In standard {\sc rnn}s, this repeating module will have a basic structure, such as a single $tanh$ layer (see Figure \ref{figure_rnn_lstm_modules} (a)). Typically, in the base module of an {\sc lstm}, a so-called long short-term memory cell and several gates interact in a more or less complex way. A gate may be triggered or not, depending on the sigmoid activation function. Thus are controled the change of state and addition of information flowing through the module. 

In a nutshell, these gates fall into three categories: (i) an input gate conditionally decides which values from the input will contribute to update the memory state, (ii) a forget gate conditionally determines what information will be discarded from the module, (iii) an output gate conditionally decides what will be output, based on module input and module memory state. Besides, the influence of these gates is controled by weights, to be learned during the training process. Each module therefore amounts to a mini state machine. Thus, the {\sc lstm} memory module improves over the common recurrent module used in {\sc rnn}s by two points: (i) it allows to tune how new information is added to the previously stored information; (ii) it enables oblivion for some part of this previously stored information. 

An illustration of the architecture of a single {\sc lstm} module is provided in Figure \ref{figure_rnn_lstm_modules} (b). The unfolding in time of this single-module {\sc lstm} is shown in Figure \ref{figure_rnn_lstm} (c).

\subsubsection{Refining the Architecture}
 
Higher-order abstractions can be taken into account if we compose {\sc lstm} modules between them. First, the concatenation of $n$ {\sc lstm} modules results in a layer of $n$ {\sc lstm} modules. Further, stacking multiple such layers will increase the complexity of the function modeled by the network.

\subsubsection{Learning procedure}

Since {\sc lstm}s are a specialization of {\sc rnn}s, weight updates and optimization resort to the same techniques. 

\subsubsection{Forecasting}

Due to their ability to address gradient explosion and vanishing gradient, {\sc lstm}s are actively investigated for time series forecasting purpose. {\sc lstm}s have been used for time series prediction in a number of domains, such as traffic speed \citep{ma_tao_wang_et_al_2015_transp-res_time-series-forecast-deep-learn-lstm-traffic-predict}, electricity price \citep{peng_liu_liu_et_al_2018_energy_times-series-forecast-deep-learn-electric-price}, electric load \citep{cheng_xu_zhang_et_al_2017_conf-inform-sci-and-syst_time-series-forecast-deep-learn-lstm-electric-load}, renewable energy power \citep{gensler_henze_sick_raabe_2016_ieee-conf-syst-man-cybern_times-series-forecast-deep-learn-lstm-solar-power} and financial markets \citep{selvin_vinayakumar_gopalakrishnan_et_al_2017_conf-advan-in-comput-comm-and-inform_times-series-forecast-deep-learn-lstm-finan,fischer_krauss_2018_euro-journ-oper-res_times-series-forecast-lstm-finance}. 

Some works dedicated to times series forecasting implement an optimization procedure to identify the hyperparameters, prior to {\sc lstm} learning. For instance, \citet{chung_shin_2018_sustainability_time-series-forecast-deep-learn-lstm-genet-algo-stock-market} combined {\sc lstm} learning with a genetic algorithm in an application dedicated to stock market forecasting. A solution using the differential evolution algorithm is described by \citet{peng_liu_liu_et_al_2018_energy_times-series-forecast-deep-learn-electric-price}, for the purpose of energy-related time series forecasting. 

Interestingly, \citet{hua_zhao_li_chen_liu_et_al_2019_ieee-comm-mag_time-series-forecast-deep-learn-stochas-lstm-telecom} introduced stochastic connectivity into conventional {\sc lstm} modules. A certain level of sparsity is thus obtained in the module architecture, which alleviates the computational burden in training. 




\subsection{Convolutional Neural Networks}\label{cnn}

The event that triggered the emergence of deep learning was an image recognition and classification contest (ImageNet, 2012). Very soon, a specific kind of deep neural networks, the convolutional neural networks ({\sc cnn}s), were proposed and dedicated to image analysis. Having connections from all nodes of one layer to all nodes in the subsequent layer, as in {\sc mlp}s, is extremely inefficient. {\sc cnn}s arose from the observation that a careful pruning of the connections, based on domain knowledge, boosts performance. A {\sc cnn} is a particular kind of artificial neural network aimed at preserving spatial relationships in the data, with very few connections between the layers. Moreover, {\sc cnn}s exploit knowledge which is intrinsic to the data considered (for instance, spatial relationships in an image). The name Temporal Convolutional Networks ({\sc tcn}s) was first referred to by \citet{bai_kolter_koltun_2018_arxiv_empir-eval-convol-neur-netw-rec-neur-netw}, to emphasize the autoregressive property of {\sc cnn}s used for a forecasting task, and the ability to process sequences of arbitrary length.

\begin{figure}[t]
\begin{center}
	\includegraphics[width=14cm]{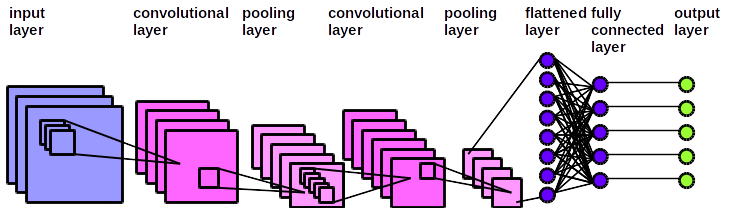}
\end{center}
\caption{Example of Convolutional Neural Network ({\sc cnn}) architecture. In {\sc cnn}s, the number of convolution and pooling layers is much larger than 2, as in this illustration. The value $C_k$ of $k^{th}$ unit in a convolutional layer is calculated as $C_k = f(x * W + b)$, where $x$ is the vector of values for the units defining the local region (in the preceding layer), associated with $k^{th}$ unit, $W$ denotes the weights characterizing the filter driving the convolution, $b$ is a bias, and $f$ is an activation function (sigmoid, tanh or {\sc r}e{\sc lu}, for instance). When the whole preceding input layer has been convolved in this way, a novel feature map (the convolutional layer) is obtained. In complement to the convolutional layer, the pooling layer is responsible for reducing the spatial size of the convolved feature. Max pooling is generally employed. Through max pooling, the maximum value from the portion of the novel feature map covered by the filter is assigned to a dedicated unit in the pooling layer. At the end of the network, the feature maps (corresponding to as many channels) obtained through the latest convolution are flattened into a single (long) one-dimensional vector. The latter vector is fully connected to an additional layer $\mathcal{L}$. The softmax function is used to obtain the final output: $softmax(z_i) = \frac{e^{z_i}}{\sum_{j \in \mathcal{L}}\ e^{z_j}}$.
}
\label{figure_cnn}
\end{figure}

In the generic scheme, the input to a {\sc cnn} is a matrix. This input is then converted into successive layers, throughout the {\sc cnn} (see Figure \ref{figure_cnn}). These layers (matrices in fact) capture relationships of increasing granularity, up to high-level details. For example, in image analysis, low-level features would be edges, corners, color or gradient orientation; a high-level feature could be a complete face. Consecutive layers of convolutions and activations, generally interspersed with pooling layers, construct the deep {\sc cnn}. 

\subsubsection{Convolution}

The two paradigms of {\sc cnn}s are local connectivity and parameter sharing. In each convolutional layer, the convolution operation in performed by a specific filter (also known as the "kernel" or "receptive field"), which is applied on the preceding layer. Each unit in a convolutional layer is connected to only one local region of the preceding layer. In other words, a small set of neighbor units of the preceding layer is processed by the kernel, to calculate the unit in the convolutional layer. When the whole preceding layer has been convolved in this way, a novel feature map (the convolution layer) is obtained. A convolution operation involves weights. Importantly, all the units in the novel feature map are processed using the same weights. Thus, it is guaranteed that all units in the novel feature map will detect exactly the same pattern. Another no less important consequence of weight sharing is the decrease in the number of learnable parameters, which translates into a more efficient learning procedure.  

\begin{figure}[!h]
\begin{center}
\begin{tabular}{ccc}
    \begin{subfigure}[b]{0.30\textwidth}
      \begin{center}
        \includegraphics[width=5cm]{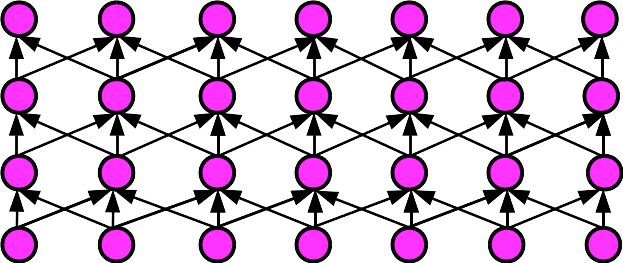} 
      \end{center}
    \end{subfigure}
    & \begin{subfigure}[b]{0.30\textwidth}
         \begin{center}
        \includegraphics[width=5cm]{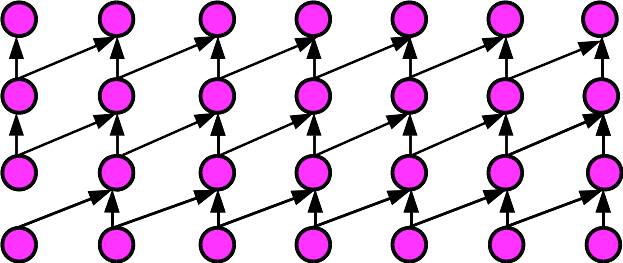}
      \end{center}
    \end{subfigure}
    & \begin{subfigure}[b]{0.30\textwidth}
         \begin{center}
        \includegraphics[width=5cm]{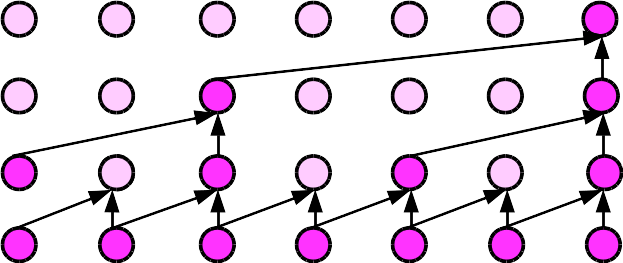}
      \end{center}
    \end{subfigure}\\
    (a) & (b) & (c)\\
\end{tabular}
\end{center}
\caption{Convolution operation. (a) Standard convolution; kernel size $= 3$: as the convolutionel kernel slides over the local region (or kernel) in the subsequent layer, the estimated value at time $t$ depends on both past and future. (b) Three-layer {\sc cnn} with causal convolution; kernel size $= 2$: the output at time $t$ does not depend on future values. (c) Dilated causal convolution: kernel size $= 2$, dilatation rate = 2: the dilatation rate defines the spacing between the values in a kernel. Dilated causal convolution provides a wider field of view at constant computational cost. Thus it allows to handle long-term dependencies within the data.}
\label{figure_std_caus_dilat_convol}
\end{figure}

In Temporal Convolutional Networks, the autoregressive characteristics implies that the value at time step $t$ must only depend on past time steps and not on future ones. To guarantee this behavior, the standard convolution operation is replaced with {\bf causal convolution} (see Figures \ref{figure_std_caus_dilat_convol} (a) and (b)). Further, {\bf dilated causal convolution} is a technique used to increase the receptive field of the {\sc tcn}, thus allowing to learn long-term dependencies within the data (see Figure \ref{figure_std_caus_dilat_convol} (c)). 

\subsubsection{Dimension reduction}

Another building block of {\sc cnn}s is the pooling layer, whose role is to subsample the feature map obtained. One aim is to reduce the spatial size of the data representation. Besides this impact on learning complexity, a second aim is to make the feature captured by the convolutional layer invariant to local anomalies (such as distorsions in image analysis, for instance). For example, max pooling returns the maximum value over neighbor units that belong to the novel feature map obtained. 


\subsubsection{Managing data seen from different perspectives}

So far, we have given a simplified description of {\sc cnn}s. The {\sc cnn} model was initially developed for an image analysis purpose. In this case, the input data, an image, comprises multiple channels (for instance Red, Green, Blue). The consequence is that kernels, together with hidden layers, have the same depth as that of the input image (the number of channels). Convolutions are applied independently on the channels. But one-dimensional data also often comes in the form of several parallel streams. For instance, electroencephalogram records are described by up to 128 channels. 

Importantly, the number of channels may not be kept constant through later layers in the {\sc cnn}. The channels (or feature maps) represent as many abstract versions of the initial data, with each channel focusing on some aspect of the information.

\subsubsection{Adaptation to time series data}
\label{adaptation_to_time_series_data}

Although initially designed to handle two-dimensional image data, {\sc cnn}s can be used to model univariate time series. A one-dimensional {\sc cnn} will just operate over a sequence instead of a matrix. 

To note, whereas handling multivariate time series requires nontrivial multivariate extensions of traditional models, {\sc cnn}s more naturally lend themselves to process multivariate time series. A multivariate time series will be fed to the {\sc cnn} as various vectors corresponding to the initial channels. For instance, in the case of economic data, one channel will correspond to the unemployment rate, another channel to the gross domestic product, a third one to the number of companies created during the year. We have seen that convolutions are applied independently (in parallel) on the dimensions (or channels) of the data. Nonetheless, putative dependencies between the data dimensions will be taken into account by the fully connected layer at the end of the {\sc cnn}.

\subsubsection{Whole \textsc{cnn} Architecture}

The first convolutional layer and pooling layer would capture low-level information from the input data. Multiple convolution, activation and pooling layers can be stacked on top of one another. Such architectures endow {\sc cnn}s the ability to extract high-level features.

Additional operations complete the description of the {\sc cnn} architecture (see Figure \ref{figure_cnn}). The multi-channel vector obtained in the last pooling layer is flattened into a single-channel vector. This (long) vector is then fully connected to an additional layer. An activation layer generates the final output.

{\sc cnn}s are trained using backpropagation as in standard artificial neural networks.

\subsubsection{Forecasting}

Notably, deep learning models have been widely investigated to address electric load forecasting \citep{gasparin_lukovic_alippi_2019_arxiv_deep-learning-time-series-forecast-electric-load}. However, in this domain, {\sc cnn}s had not been studied to a large extent until recently \citep{amarasinghe_marino_manic_2017_internat-sympos-indust-electro_convol-neur-netw-energy-load-forecast,almalaq_edwards_2017_icmla_review-deep-learn-time-series-load-forecasting,kuo_huang_2018_energies_convol-neur-netw-energy-load-forecast}. The three latter works reported that {\sc cnn}s have been proven comparable to {\sc lstm}s as regards electricity demand forecasting. Furthermore, {\sc cnn}s were shown to outperform {\sc lstm}s in works involving energy-related time series such as in photovoltaic solar power prediction \citep{koprinska_wu_wang_2018_int-joint-conf-neural-netw_time-series-forecast-deep-learn-cnn-solar-power} and in experimentations focused on power demand at charging points for electric vehicles \citep{lara-benitez_carranza-garcia_luna-romera_et_al_2020_energy_survey-time-series-forecast-deep-learn-cnn-energy-domain}. Importantly, all these works highlighted a better adequacy of {\sc cnn}s for real-time applications, compared to other neural networks. The explanation lies in their faster training and testing execution times.

Hybrid models combining convolutional and {\sc lstm} layers have been proposed. In some of these works, the feature maps generated through a {\sc cnn} are input into an {\sc lstm} which is in charge of prediction (see for instance the work of \citealp{cirstea_micu_muresan_2018_int-conf-inf-knowl-manag_multivar-time-series-forecast-deep-learn-hybrid-cnn-lstm-chem-concentr}, with its application to chemical concentration prediction). The approach developed by \citet{he_2017_int-conf-infor-techno-and-quantit-manag_convol-neur-netw-energy-load-forecast} relies on {\sc cnn}s, to extract features from multiple input sources, before an {\sc rnn} captures the temporal dependencies in the data. Other works implement further model integration by combining the features extracted in parallel from a {\sc cnn} and an {\sc lstm}. The works developed by \citet{tian_ma_zhang_et_al_2018_energies_time-series-forecast-deep-learn-hybrid-lstm-cnn-load} and \citet{shen_zhang_lu_et_al_2019_neurocomputing_time-series-forecast-deep-lean-hybrid-cnn-lstm-meteo-finan} illustrate this latter approach for the energy, meteorology and finance fields.


\subsection{Transformers}\label{transformers}  

Seq2Seq models allow to transform an input sequence into an output sequence. Sections \ref{rrn} and \ref{lstm} have spotlighted how {\sc rnn}s, and a specialized version of {\sc rnn}s, {\sc lstm}s, are suited to handle sequential data based on recurrent modules. Transformers are a class of deep learning Seq2Seq models that were recently introduced \citep{vaswani_shazeer_parmar_et_al_2017_neurips_attention_is_all_you_need_transformer}. Like {\sc rnn}s, transformers were designed to handle sequential data and tackle problems in natural language processing (for instance, translation). Transformers were repurposed to address forecasting in time series (see for instance \citealp{nino_2019_thesis_transformers-time-series-forecast}). 

The encoder and decoder form the two parts of the transformer's architecture (Figure \ref{figure_transformer}). The encoder mainly consists of an input layer, an input encoding, a positional encoding mechanism, and a stack of identical encoder layers. In the seminal work of \citet{vaswani_shazeer_parmar_et_al_2017_neurips_attention_is_all_you_need_transformer}, the decoder is composed of an input layer, an input encoding, a positional encoding mechanism, a stack of identical decoder layers and an output layer. 

\subsubsection{Transformers {\it versus} recurrent neural networks}

A first difference between {\sc rnn}s and transformers is that in the former, the sequential data is input element by element (one step at a time). In transformers, the entire sequence is input all at once while positional encoding preserves the sequential nature of the data. In natural language processing ({\sc nlp}) tasks, transformers rely on word and phrase encodings to represent the input data with a gain of performance in the transformation. More generally, encoding an input sequence results in representing each input element as a vector of identifiers (for example co-occurring word identifiers in {\sc nlp}). The encoding representation of the data is obtained by forcing the transformer to reconstruct the original sequence. The {\it encoder} is delegated this task. In time series applications, the {\it encoder} allows to summarize information and integrate temporal behavior from the time series under analysis. 

In connection with this first difference, another difference with {\sc rnn}s is that transformers do not require the input data to be processed sequentially. Namely, transformers do not need to process the beginning of the input sequence before its end. Therefore, transformers lend themselves to much further parallelization than {\sc rnn}s, which decreases training times. Moreover, since the sequence is not fed sequentially to a transformer, training a transformer for a forecasting purpose requires specific mechanisms to ensure that predicting a data point will only depend on previous data points. This property is obtained through two mechanisms, look-ahead masking and one-position shifting between the decoder input and target output (to be further described). 

\begin{figure}[t!]
\begin{center}
	\includegraphics[width=10cm]{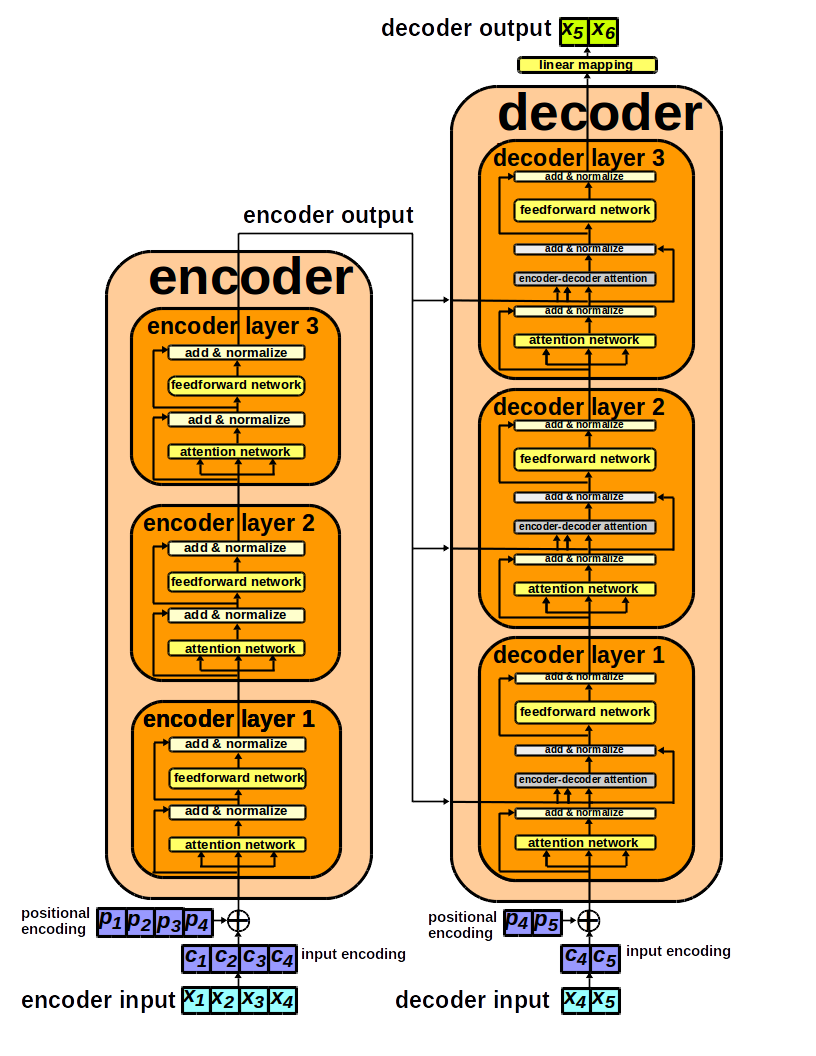}
\end{center}
\caption{Example of transformer (or encoder-decoder) architecture. The inputs and outputs (target sentences in the training task) are first each encoded as a vector of $d$-dimensional vectors. $\bullet$ The encoder mainly consists of an input layer, an input encoding, a positional encoding mechanism, and a stack of identical encoder layers. The input layer maps the input time series data to a vector of $d$-dimensional vectors (input encoding). The positional encoding mechanism integrates sequential information to the model through element-wise addition with the original input encoding. In the end, the encoder yields a vector of $d$-dimensional vectors which is fed to the decoder. $\bullet$ In this illustration inspired from the seminal work of \citet{vaswani_shazeer_parmar_et_al_2017_neurips_attention_is_all_you_need_transformer}, the decoder is composed of an input layer, an input encoding, a positional encoding, a stack of identical decoder layers and an output layer. The decoder input starts with the last data point of the encoder input ($X_4$ in this example). The decoder input is mapped to an encoded input layer. As in the first encoder layer, the input for the first decoder layer integrates positional information supplied by a positional encoding vector. Downstream the last decoder layer, an output layer maps the output of last decoder layer to the decoder output. $\bullet$ The encoder and decoder layer architectures are described in detail in Figure \ref{figure_attention}.}
\label{figure_transformer}
\end{figure}

A third difference occurs in the way {\sc rnn}s and transformers capture the dependencies within the input sequential data for a forecasting purpose. {\sc rnn} units can remember or forget parts of the previously stored information, depending on whether they find them important or not. Relying on a single {\sc rnn} unit entails that this latter be able to memorize important events of the past and use these events to predict future values. In contrast, instead of a single multi-task unit, transformers rely on two specialized units. These units define the characteristic {\it encoder-decoder} architecture shown in Figure \ref{figure_transformer}. In time series prediction, the encoder extracts the important features from the past events, whereas the decoder exploits this information to predict future values. 

In {\sc lstm}s, an issue is that there is no way to assign more importance to some parts of the input sequence compared to others while processing the sequence. The attention mechanism, a corner stone in the transformer model, emerged as an improvement to capture dependencies, especially long-range dependencies, in a sequence. In the traditional encoder-decoder model, only is the final vector produced by the encoder used to initialize the decoder. However, summarizing a long input sequence into this single vector drastically decreases the transducer performance. The key idea behind attention consists in using all intermediate encodings generated by the encoder, to enrich the information passed to the decoder.

Recurrent networks were put forward as the best way to capture timely dependencies in sequences. Specifically, {\sc lstm}s are supposed to capture long-range dependencies but they may fail to meet this purpose in practice. The introduction of the transformer model brought to light the fact that attention mechanisms in themselves are powerful and dispense a sequential recurrent processing of the data \citep{vaswani_shazeer_parmar_et_al_2017_neurips_attention_is_all_you_need_transformer}. Instead, processing all input elements at the same time and calculating {\it attention weights} between them enables to relate different positions in a sequence. Thus long-range dependencies may be captured in the representation of the sequence. 

Another difference is that {\sc rnn}s only allow one-step-ahead prediction. Therefore, multi-step-ahead is handled in {\sc rnn}s by iterating $H$ times a one-step-ahead prediction: the newly predicted series value is fed back as an input for the future series value. A major flaw of this iterated scheme lies in error propagation. Instead, transformers allow direct forecast of $x_{T+h}$ with $h \in \{1,2, \cdots H\}$.

Each encoder or decoder layer makes use of an attention mechanism to process each element of its input. Namely, the attention mechanism weights the relevance of every other element in the input sequence and extracts information accordingly, to generate the layer output. Similarly to kernels in {\sc cnn}s, multiple attention heads allow to capture various levels of relevance relations. Each decoder layer also has an additional attention mechanism which applies self-attention over the encoder output.



\subsubsection{Attention mechanism}

An attention mechanism computes weights representing the relevance of the elements in the input sequence (named keys hereafter) to a peculiar output (named query). Otherwise stated, weights indicate which elements in the input sequence are more important for predicting each output element.\newline

\noindent $\bullet$ {\bf Contextual information} 
%
Given the input sequence of $n$ elements $x_1, x_2, \dots, x_n$, each $i^{th}$ element is mapped to an associated code $c_i$ which is a, say, $d$-dimensional vector. However, this encoding neither takes into account the surrounding context nor the position in the sequence of the elements.


To address the first issue, the key is to quantify how similar $i^{th}$ element represented by $c_i$ ({\bf query}) is to each of the codes $c_1, c_2, \dots c_n$ ({\bf keys}). The inner product operation, followed by exponentiating and normalizing, yields the relative relationship $r_{i \rightarrow j}$ that exists between $i^{th}$ and $j^{th}$ elements. Thus, $r_{i \rightarrow j}$ indicates how much attention should be paid to code $c_j$ when constituting a new encoding $\tilde{c}_i$ for element $i$. $\tilde{c}_i = \sum_{j=1}^n\ r_{i \rightarrow j}\ c_j$ is therefore informed by the context in which $i^{th}$ element appears. This mechanism implements self-attention over a single sequence. 

%

If element $i$ is highly related to a particular element $j$, $r_{i \rightarrow j}$ will be large. Thus, the {\bf value} $c_j$ multiplied by relative weight $r_{i \rightarrow j}$ will contribute significantly to the revised code $\tilde{c}_i$. As a consequence, the same element $i$ observed in different contexts $(c_1, c_2, \dots, c_n)$ would be mapped to different codes $\tilde{c}_i$.

To generalize, the attention function is depicted as mapping a query and a set of key-value pairs to an output. The query, keys and values are all vectors. The output consists in a weighted sum of the values. In practice, the attention function is computed simultaneously on a set of queries which form the matrix $Q$. Similarly, keys and values are respectively stored into the matrices $K$ and $V$. For queries and keys of dimension $d$, and values of dimension $v$, dot-product-based attention provides the vector ouputs according to the following principle:
\begin{equation}
weights(Q,K,V) = softmax(\frac{QK^T}{\sqrt{d}})\ V,
\end{equation}

where $A^T$ denotes the transpose of matrix $A$ and $softmax$ allows to nonlinearly scale the weight values between $0$ and $1$. For large $d$ values, the dot-product is likely to produce very large magnitudes. Very small gradients would therefore be passed to the softmax function. To address this issue, scaling by $\sqrt{d}$ factor is applied.

%
%
%

\begin{figure}[h!]
\begin{center}
\begin{tabular}{cc}
    \begin{subfigure}[b]{0.45\textwidth}
      \begin{center}
        \includegraphics[width=5.7cm]{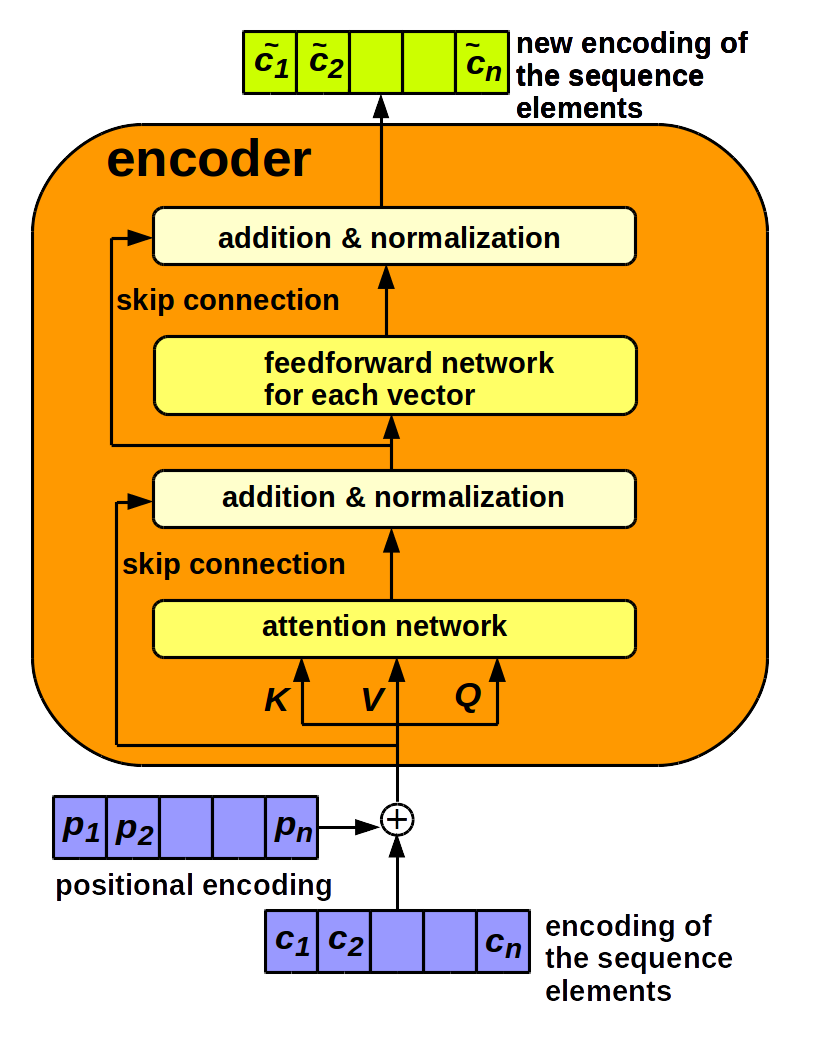} 
      \end{center}
    \end{subfigure}
    & \begin{subfigure}[b]{0.45\textwidth}
         \begin{center}
        \includegraphics[width=6cm]{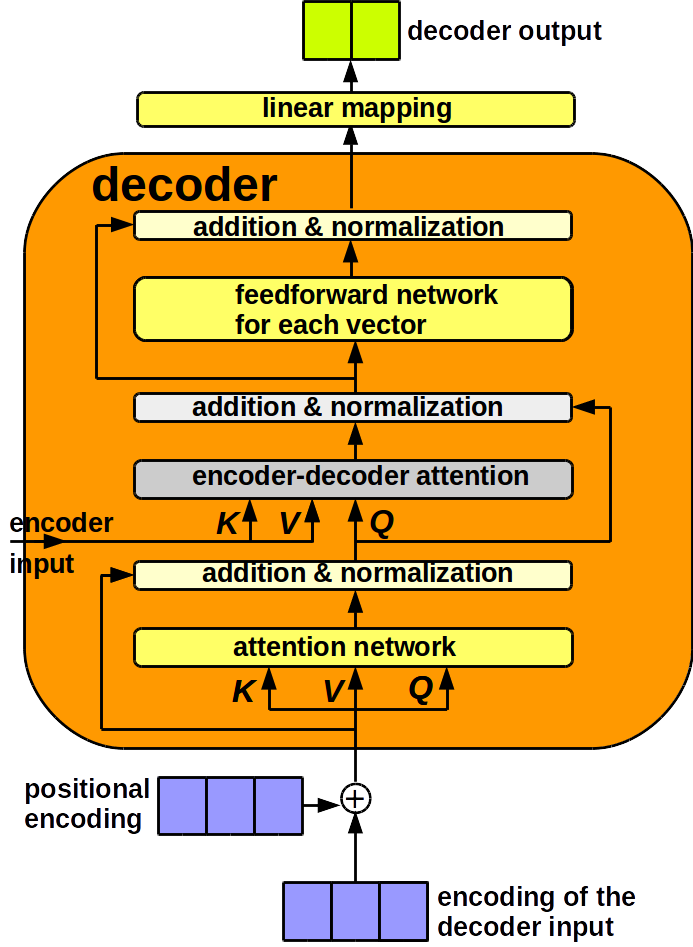}
      \end{center}
    \end{subfigure}\\
    (a) & (b)\\
\end{tabular}
\end{center}
\caption{Attention mechanisms used in a transformer. {\bf (a)} Attention mechanism in a single-layer encoder. {\bf (b)} Attention mechanisms in a single-layer decoder. $\bullet$ Each encoder layer mainly consists of two sub-layers: a self-attention mechanism and a fully-connected feed-forward neural network. Each sub-layer is followed by a complementary sub-layer, where point-wise vector addition and normalization are performed, to implement the so-called skip connection. Except for first encoder layer, the self-attention mechanism is fed with the set of encodings generated by the previous encoder layer and weights their relevance to each other to generate a set of output attended encodings. $\bullet$ Besides a self-attention mechanism and a feed-forward neural network, a decoder layer includes a third sub-layer between the two former: the role of this encoder-decoder attention mechanism is to help the decoder focus on relevant parts of the input sentence.}
\label{figure_attention}
\end{figure}

Figure \ref{figure_attention} describes in detail the architectures for the encoder and decoder layers. Besides a self-attention network present in both layers, a cross-attention mechanism is implemented through the so-called encoder-decoder attention network. Table \ref{table_q_k_v} details $Q$, $K$ and $V$ for the three separate attention mechanisms of a transformer. 

%

\begin{table}[ht]
\scalefont{0.85}
\centering
\begin{tabular}{|l|l|l|}
\hline
Encoder attention         &Q:   &vector associated with current element of input sequence\\ 
                          &K=V: &vectors corresponding to all elements of input sequence\\
Decoder attention         &Q:   &vector associated with current element of output sequence\\ 
                          &K=V: &vectors corresponding to all elements of output sequence\\
Encoder-decoder attention &Q:   &output of the decoder's masked attention (output of the previous decoder layer)\\
                          &K=V: &all the vectors output by the encoder layers (ouput of the encoder)\\
\hline
\end{tabular}
\caption{Details on queries, keys and values for the three separate attention mechanisms of a transformer. $Q$: queries, $K$: keys, $V$: values. Generically, attention is used to relate two different sequences to one another. In self-attention, different positions of the same input sequence are related to one another. Therefore, $K$ and $V$ are the same.}
\label{table_q_k_v} 
\scalefont{1.0}
\end{table}

In the self-attention mechanism of an encoder layer, all the keys, values and queries come from the previous encoder layer. In the encoder-decoder attention mechanism of a decoder layer, the queries comme from the previous decoder layer, whereas the keys and values come from the output of the encoder.\newline

\noindent $\bullet$ {\bf Positional information}
%
The attention mechanism described so far pays no attention to the order of the elements. To palliate this drawback, it has been found convenient to add a so-called {\bf skip connection} to preserve the order of the elements: each original encoding $c_i$ is added to the output of the attention network, and the codes thus modified are normalized. 

Further, this set-up is augmented with a positional encoding, to account for the positions of the elements. This positional encoding will be added to the original element embedding $c_i$. Since we have a $d$-dimensional element encoding $c_i$ for element $i$, we wish to produce a $d$-dimensional positional encoding $p_i$ for this element. One standard way to construct $p_i$ for an element $i$ located at position $pos$ is to use sinusoidal waves as follows:

\begin{equation}
\label{positional_encoding}
\begin{split}
    p_i(pos,2k)   = sin(pos/c^{2k/d}),\\
    p_i(pos,2k+1) = cos(pos/c^{2k/d}),
\end{split}
\end{equation}  

with $c$ some constant, $\omega_{2k} = 1/c^{2k/d}$ the sinusoidal wave frequency, and $2k$ and $2k+1$ even and odd indexes in $1,\ ... ,d$. As we move from 1 to $d$, the sinusoidal wave oscillates faster. The original element encoding added to this positional encoding is input into the attention network as shown in Figures \ref{figure_attention} (a) and (b). Finally, each of the $n$ vectors output by the attention network is fed through a simple (feedforward) neural network. Again, skip connection, addition and normalization are applied to this part of the framework. 

This scheme (attention, addition and normalization, [cross-attention, addition and normalization,] feedforward network, addition and normalization) can be repeated $k$ times (a standard value for $k$ is $6$). This yields a final deep sequence encoder [decoder].\newline

%


\noindent $\bullet$ {\bf Multihead attention}
%
Finally, similarly to convolutional neural networks that rely on several kernels, several self-attention sub-layers can be used in parallel in transformers.
This so-called multi-head attention mechanism uses different linear projections of $Q$, $K$ and $V$. Learning from different representations of $Q$, $K$ and $V$ is beneficial to the model. The encoding vectors obtained through these heads are concatenated, then a dimension reduction step is applied to yied the final encoding vector.

\subsubsection{Forecasting}

In the self-attention mechanism of a decoder layer, each position in the decoder only attends to positions in the decoder up to and including this position. Values corresponding to subsequent forbidden positions are masked out in the input of the softmax function.

Together with look-ahead masking, one-position shifting between the decoder input and target output (decoder output) ensures that the prediction for a given position only depends on the known outputs at positions strictly less than this position. Thus can be preserved the autoregressive property. 
%

Given a time series containing $N$ data  points $x_{t-N+1},...,x_{t-1},x_t$, H-step  ahead  prediction is formulated as a supervised machine learning task:  the  input of  the  model is $x_{t-N+1},...,x_{t-H}$, and  the output is $x_{t-H+1},x_{t-H+2},...,x_t$. During model training, one-position shifting between the decoder input and the target output (decoder output) prevents learning the decoder to merely copy its input, and contribute to ensure that the prediction for a given position only depends on the known outputs at positions strictly less than this position. This shifting mechanism is combined with look-ahead masking that restrains attention to datapoints in the past: in the self-attention mechanism of a decoder layer, each position in the decoder only attends to positions in the decoder up to and including this position. Values corresponding to subsequent forbidden positions are masked out. In the training phase, the first neural network in an encoder layer reads the input sequence one time step at a time. 

Transformer models have been applied in such various domains as influenza-like illness forecasting \citep{wu_green_ben_2020_arxiv_time-series-forecast-transformer-influenza} and prediction of vehicle trajectories \citep{park_kim_kang_et_al_2018_ieee-intell-vehic-symp_encoder-decoder-predic-vehic_traject}. To guess the future trajectory of surrounding vehicles in real time, the latter work relies on the following architecture: {\sc lstm}-based encoders analyze the patterns underlying the past trajectory data, whereas {\sc lstm}-based decoders predict the future trajectories.


The point-wise dot-product self-attention in canonical transformers is agnostic of local surrounding context. This may entail confusion regarding whether an observed point is an anomaly, a change point or part of the patterns underlying the data. To increase prediction accuracy, \citet{li_jin_xuan_2019_neurips_time-series-forecast-transformer-locality-logasparse-convol} employ causal convolution to incorporate local contextual information such as local shapes in the attention mechanism. Further, the authors proposed a log-sparse self-attention mechanism, to increase forecasting accuracy for time series with fine granularity and long-term dependencies under memory-limited budget.

\citet{phandoidaen_richter_2020_arxiv_time-series-forecast-encoder-decoder-theory-estimators} investigated transformers to forecast high-dimensional time series. In their paper, the authors impose a specific encoder-decoder structure as a reasonable approximation of the underlying autoregressive model $f_0$. They assume that their model compresses the given information of the last $r$ lags into a vector of much smaller size and afterwards expands this concentrated information to generate the observation of the next time step ahead. A theoretical analysis of the forecasting ability of an estimator $\hat{f}$ of $f_0$ was developed under various structural and sparsity assumptions. In particular, upper bounds were provided for the forecast errors. The performances of the various neural network estimators were analyzed on simulated data. 
 


\begin{figure}[t]
\begin{center}
        \includegraphics[width=8.5cm]{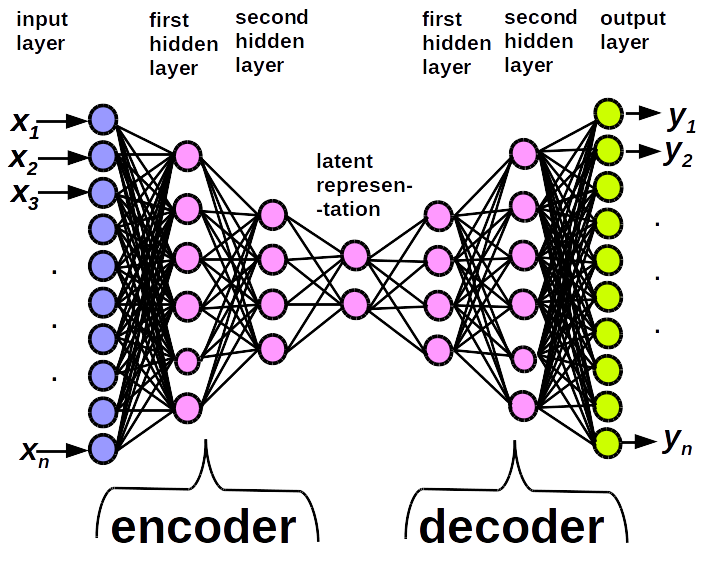}
\end{center}
\caption{Example of autoencoder ({\sc ae}) architecture.}
\label{figure_autoencoder}
\end{figure}

A specific kind of encoder-decoder architecture is the autoencoder ({\sc ae}). An {\sc ae} is a Multilayer Perceptron Network whose particular architecture is described as follows: an input layer and an output layer of same size are connected through one or more hidden layers; the hidden layers of the encoding side are mirrored to constitute the decoding side of the {\sc ae} (see Figure \ref{figure_autoencoder}). Reconstructing the input by minimizing the difference between the input and the output (predicted input) then allows to implement feature extraction in an unsupervised learning scheme.

\citet{gensler_henze_sick_raabe_2016_ieee-conf-syst-man-cybern_times-series-forecast-deep-learn-lstm-solar-power} developed an approach combining an {\sc ae} with an {\sc lstm}, to forecast power in renewable energy power plants. In the domain of financial time series forecasting, \citet{bao_yue_rao_2017_plos-one_autoencoder_lstm} considered stacked {\sc ae}s to extract features, together with an {\sc lstm} to generate the one-step-ahead output prediction. The stacked autoencoder architecture is constructed by stacking a sequence of single-hidden-layer {\sc ae}s, layer by layer. In this scheme, the hidden layer of the previous {\sc ae} serves as the input layer for the subsequent {\sc ae}.


\section{Time Series Model Evaluation}
\label{time.series.model.evaluation}
Generally, two methods are used in order to assess whether a time series model (linear or nonlinear) is suited to a given time series data: model diagnosis and out-of-sample forecast performance.

\subsection{Model Diagnosis - Goodness-of-fit}
Model diagnosis consists in testing how a model adjusts to the observed time series data. To this end, the residuals of the fitted model are analyzed. We recall that residuals are computed as the difference between the observed time series and their predictions given by the model. If the residuals are found uncorrelated, that means the model successfully captures the dependencies within the observations. Statistical tests can be used for this purpose, such as {\textbf{Durbin-Watson test} \citep{durbin_watson_1950_biometrika_serial_correlation_test_one, durbin_watson_1951_biometrika_serial_correlation_test_two} and \textbf{Ljung-Box test} \citep{ljung-box_1978_biometrika_arma-model-valid-non-correl-test}. These tests detect the presence of autocorrelation at lag 1 and lag $\ge$ 1, respectively. Such residual analysis has been used by \citet{cheng_2016_journ-comm-in-stat-theory_transitional-msar} and \citet{deschamps_2008_journ-applied-econo_comparing-star-msar} to respectively compare {\sc ar} models (subsection \ref{lar}) to {\sc msar} models (subsection \ref{msar}), and {\sc msar} models to {\sc star} models (subsection \ref{star}).

A well-known alternative to model diagnosis are goodness-of-fit tests \citep{d-agostino_stephens_1986_book_goodness-of-fit}, which address the question of model consistency (time series models or others) with the observed data. Graphical techniques rely on a qualitative ({\it i.e.}, visual) examination of the fit of the model to data, using appropriate curves \citep{mackay_2004_journ-biometrics_goodness-of-fit-hmm, willems_2009_journ-environmental-model_multi-criteria-goodness-of-fit}. Quantitative methods implement statistical hypothesis tests \citep{escanciano_2006_journ-american-stat-asso_goodness-of-fit-linear-non-lin-time-series, gonzalez-crujeiras_2013_test_goodness-of-fit-review, remillard_2017_journ-econometrics_goodness-of-fit}.
In the latter case, when tests are carried out through Monte Carlo simulations, they are referred to as Monte Carlo goodness-of-fit tests \citep{waller-smith-childs_2003_journ-ecological-mod_monte-carlo-goodness-of-fit}.

\subsection{Out-of-sample Forecast Performance}
Following time series model specification (\ref{time.series.model.specification}), forecasts are made using the conditional mean $f$ and conditional standard deviation $g$. Thus, one-step ahead prediction interval writes
$$\left[ f(Z_{T}, ..., Z_{T-p+1}) - g(Z_{T}, ...,  Z_{T-p+1}), f(Z_{T}, ..., Z_{T-p+1}) + g(Z_{T}, ...,  Z_{T-p+1}) \right],$$
where $f(Z_{T}, ..., Z_{T-p+1})$ is the best forecast at time $T+1$, $T$ is the number of observations within the training set and $p$ is the autoregressive order. In order to perform $h$-step ahead predictions, with $h>1$, the past predicted values are used as inputs, as shown in figure \ref{forecast.procedure}. For $h=2$, $Z_{T+2}$ is predicted by $f(\hat{Z}_{T+1}, Z_T, ..., Z_{T-p+2})$; and for $h=3$, $Z_{T+3}$ is predicted by $f(\hat{Z}_{T+2}, \hat{Z}_{T+1}, Z_T, ..., Z_{T-p+3})$.
%
\begin{figure}[t]
\begin{center}
	\includegraphics[width=13cm]{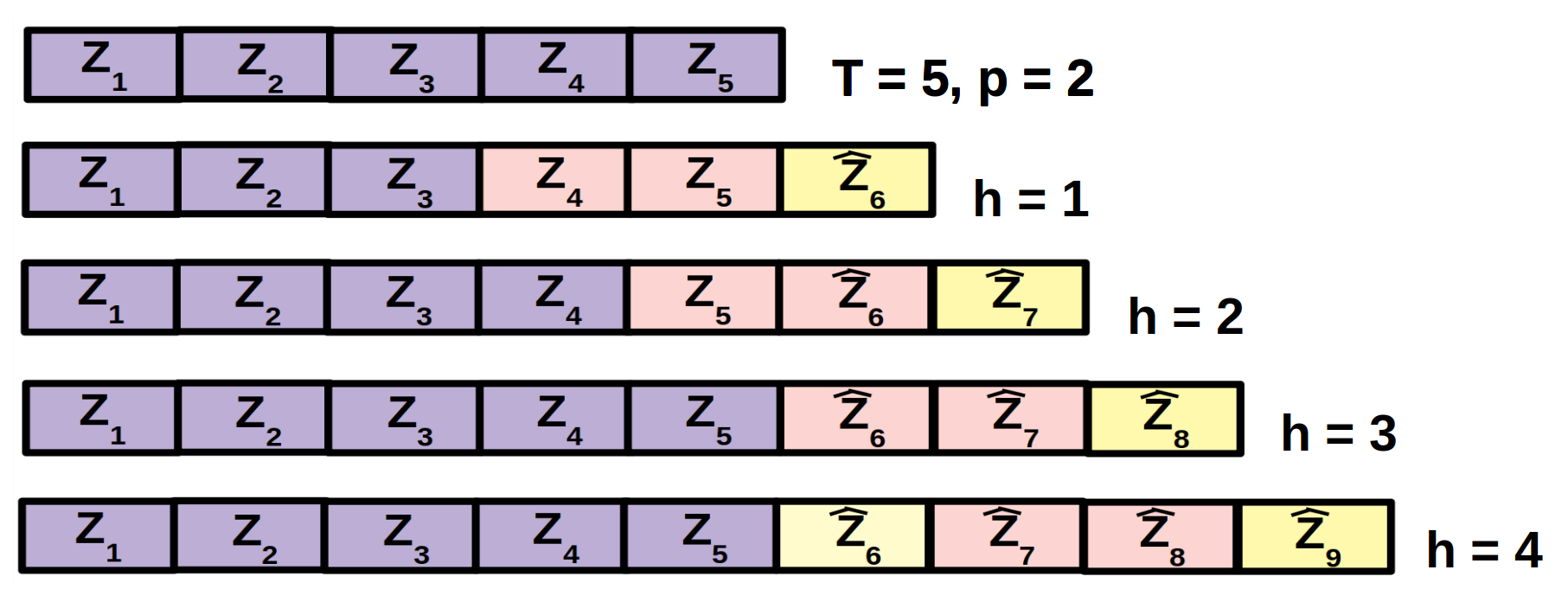}
\end{center}
\caption{Forecast procedure at horizon $h$, that is at time $T+h$ where $T$ is the number of observations in the training set. $p$ is the autoregressive order. $T = 5$, $p = 2$. Prediction at $h=1$ is a function of the $p$ latest observed values.
Prediction at $h=2$ is a function of the $p-1$ latest observed values and of prediction at $h=1$. Prediction at $h=3$ is a function of the $p-2$ latest observed values and of predictions at $h=1$ and $h=2$.\\
}
\label{forecast.procedure}
\end{figure}

Assessing prediction performance based on test-sets is crucial.} Several performance metrics have been proposed for this purpose, in the literature. For example, normalized root mean square error ({\sc nrmse}) based on $L_2$-norm provides a good measure of distorsion when Gaussian errors are considered. Normalized mean absolute percentage error ({\sc nmape}) is suited to data containing outliers. Bias metric allows to assess whether the predictions are underestimated (positive bias) or overestimated (negative bias). These metrics are defined as

\begin{align}
\text{NRMSE} &= \sqrt{\frac{1}{H} \sum_{h=1}^H \left( \frac{Z_{T+h}-\hat{Z}_{T+h}}{\max(Z_{T+1},..., Z_{T+H})} \right)^2},\\
\text{NMAPE} &= \frac{1}{H} \sum_{h=1}^H  \frac{|Z_{T+h}-\hat{Z}_{T+h}|}{\max(Z_{T+1},..., Z_{T+H})} \times 100,\\
\text{Bias}  &= \frac{1}{H} \sum_{h=1}^H (Z_{T+h}-\hat{Z}_{T+h}),
\end{align}
where $\{Z_{T+h} = z_t\}_{h=1, ..., H}$ is the test-set, $\{\hat{Z}_{T+h}\}_{h=1, ..., H}$ is the forecasted values, $T$ is the length of the training-set and $H$ is the number of ahead predictions. Best forecast performances are reached with lower values of {\sc nrmse} and {\sc nmape} and bias near to zero. 
 
	Some models are accurate for short-term forecasts (small values of $H$), and when the time horizon becomes large, their performances decrease drastically. This is generally observed when a nonlinear time series is modeled by a linear model \citep{karakucs-kuruoglu-altinkaya_2017_jour-wind-speed-power-prediction_poly-autoregress-model}. In this case, linear approximation holds for short-term predictions but when prediction horizon becomes large this approximation becomes bad. 

Readers interested in thorough content on performance estimation methods in time series forecasting models are directed to the work of \citet{cerqueira_torgo_mozetic_2020_mach-learn_emp-study-on-perform-estim-meth-time-ser-forec}.

\section{Available Implementations}
Table \ref{available.implementations} gives a recapitulation of available \textit{R} and \textit{Python} implementations for methods/models presented throughout the present review of time series model analysis.

%
\begin{table}[t!]
\scalefont{0.65}
\centering
\begin{tabular}{|l|l|ll|}
\hline
 & \multirow{2}{*}{\textbf{Method/Model}} & \multirow{2}{*}{\textbf{Language}} & \multirow{2}{*}{\textbf{Package/Module (\textit{function/class}})} \\
 &                               &                           &     \\
\hline
\multirow{8}{*}{Sationarity tests}  & \multirow{2}{*}{Phillip-Perron test}  & \textit{R}       &  tsesies (\textit{pp.test}) or urca (\textit{ur.pp}) \\
                                    &                                       & \textit{Pyhton}  &  arch.unitroot (\textit{PhillipsPerron}) \\
\cline{2-4}                                 
           & \multirow{2}{*}{(Augmented) Dickey-Fuller test} & \textit{R}   & tseries (\textit{adf.test}) or urca (\textit{ur.df})  \\
           &                                                 & \textit{Pyhton} & statsmodels.tsa.stattools (\textit{adfuller}) or arch.unitroot (\textit{ADF}) \\
\cline{2-4}
          & \multirow{2}{*}{KPSS test} & \textit{R}      & tseries (\textit{kpss.test}) or urca (\textit{ur.kpss}) \\
          &                            & \textit{Python} & statsmodels.tsa.stattools (\textit{kpss}) or arch.unitroot (\textit{KPSS})\\                                                                              
\cline{2-4}
\cline{2-4}
             & \multirow{2}{*}{Zivot-Andrew test}  & \textit{R}       & urca (\textit{ur.za}) \\
             &                                     & \textit{Pyhton}  & statsmodels.tsa.stattools (\textit{zivot\_andrews}) or arch.unitroot (\textit{ZivotAndrews})                 \\
\hline       
	                       & Seasonal-trend decomposition  & \textit{R}      & stats (\textit{stl})    \\
	                       & using Loess                   & \textit{Python} & statsmodels.tsa.seasonal (\textit{STL})  \\
\cline{2-4}
Time series                & \multirow{2}{*}{{\sc x}11 and {\sc seat}s} & \textit{R}      & seasonal (\textit{seas}) \\                            
decomposition              &                                & \textit{Python} & statsmodels.tsa.x13 (\textit{x13\_arima\_select\_order})                         \\                               
\cline{2-4}                           
                           & Locally estimated       & \textit{R}         & stats (\textit{loess})     \\
                           & scartterplot smoothing  & \textit{Python}    & statsmodels.nonparametric.smoothers\_lowess (\textit{lowess}) \\                          
\hline
Exponential                &  & \textit{R}      & forecast (\textit{ets})  \\
smoothing methods          &  & \textit{Python} & statsmodels.tsa.holtwinters  (\textit{ExponentialSmoothing}) \\
\hline
                   & \multirow{2}{*}{Differentiation operator $\Delta_m^k$}   & \textit{R}       & stats (\textit{diff})   \\
Time series        &                                                          & \textit{Python}  &                         Pandas (\textit{diff})\\
\cline{2-4}
transformation     & \multirow{3}{*}{Box-Cox transformation}   & \multirow{2}{*}{\textit{R}} & EnvStats (\textit{boxcox})  \\
                   &                                           &                             & or forecast (\textit{BoxCox, BoxCox.lambda})  \\
                   &                                           & \textit{Python}             & scipy.stats (\textit{boxcox})   \\
\hline
                           & \multirow{2}{*}{Ljung-Box test}   & \textit{R}  & stats (\textit{Box.test}) \\
						  &                                 & \textit{Python} & statsmodels.stats.diagnostic (\textit{acorr\_ljungbox})  \\
\cline{2-4}
autocorrelation          & \multirow{2}{*}{Durbin-Watson test}  & \textit{R}      & stats (\textit{dwtest})  \\
and partial               &                                      & \textit{Python} & statsmodels.stats.stattools (\textit{durbin\_watson}) \\
\cline{2-4}
autocorrelation   	     & \multirow{2}{*}{(Partial) Autocorrelation function}  & \textit{R}      & stats (\textit{(p)acf})  \\
                            &                                                      & \textit{Python} & statsmodels.tsa.stattools (\textit{(p)acf})   \\
\hline
                                     & \multirow{2}{*}{{\sc ar} model}   & \textit{R}        & tseries (\textit{arma}) with $q=0$\\
								    &                             & \textit{Python}   & statsmodels.tsa.ar\_model (\textit{AutoReg}) \\
\cline{2-4}
\multirow{2}{*}{Linear times models} & \multirow{2}{*}{{\sc ma} model} & \textit{R}          & tseries (\textit{arma}) with $p=0$ \\
                                     &                             & \textit{Python}   & statsmodels.tsa.arma\_model (\textit{ARMA}) where $p=0$ \\
\cline{2-4}
									& \multirow{2}{*}{{\sc arma} model} & \textit{R}        &  tseries (\textit{arma})\\
                                     &                             & \textit{Python}   & statsmodels.tsa.arma\_model (\textit{ARMA}) \\
\hline
                                & \multirow{2}{*}{{\sc arima} model}   & \textit{R}        & forecast (\textit{Arima}) with $P=D=Q=0$ \\
\multirow{2}{*}{{\sc arima - sarima}} &                                & \textit{Python}   & statsmodels.tsa.arima\_model (\textit{ARIMA}) with $P=D=Q=0$ \\
\cline{2-4}
                                & \multirow{2}{*}{{\sc sarima} model}  & \textit{R}        &  forecast (\textit{Arima}) \\
                                &                                & \textit{Python}   &  statsmodels.tsa.arima\_model (\textit{ARIMA})  \\
\hline
                 & \multirow{2}{*}{{\sc par}}    & \textit{R}        &       -          \\
                 &                         & \textit{Python}   &       -          \\    
\cline{2-4}                                    
                 & \multirow{2}{*}{{\sc far}}    & \textit{R}        &       -          \\
                 &                         & \textit{Python}   &       -          \\     
\cline{2-4}                                   
Nonlinear time    & \multirow{3}{*}{{\sc msar}}  & \textit{R}        & MSwM (\textit{msmFit}) \\
series models     &                        & \multirow{2}{*}{\textit{Python}}   &  statsmodels.tsa.regime\_switching.markov\_autoregression    \\ 
                  &                        &                                    & (\textit{MarkovAutoregression}) \\ 
\cline{2-4}                                       
                  & \multirow{2}{*}{sc star}  & \textit{R}        & tsDyn (\textit{tar, setar, lstar})  \\
                  &                        &  \textit{Python}  & -                \\                                        
\cline{2-4} 
                  & \multirow{2}{*}{Angle's test for {\sc arch}}     & \textit{R}        &  -\\
 &                                                             &  \textit{Python}  & statsmodels.stats.diagnostic \textit{(het\_arch)}  \\    
 \cline{2-4} 
                  & \multirow{2}{*}{({\sc (g)arch}} & \textit{R}        & tseries (\textit{garch}) \\
 &                                         &  \textit{Python}  & arch (\textit{arch\_model}) \\                                          
\hline
              & {\sc mlp},    & \multirow{2}{*}{\it{R}}  & nnet, rnn, deepnet, h2o, mxNet \\
Deep learning & {\sc rnn}, &  & tensorflow, keras (interfaces to python libraries) \\
\cline{3-4}
 & {\sc lstm}, & \multirow{2}{*}{\it{Python}} & \multirow{2}{*}{Tensorflow, Keras, PyTorch, PySpark} \\
               & {\sc cnn}, transformers   &    & \\ 
\hline
\end{tabular}
\caption{Time series analysis methods and models presented throughout the present survey. Recapitulation of available \textit{R} and \textit{Python} implementations.}
\label{available.implementations}
\scalefont{1.0}
\end{table}



\section{Future Directions of Research}

This survey is focused on modeling and forecasting in time series. Directions of research are therefore mentioned for these domains only.

\subsection{Conventional Models {\it versus} Deep Neural Models}
A vast field of possibilities has opened up for the modeling and prediction of time series with the emergence of deep learning. Deep neural networks are renowned for their ability to extract weak signals as well as complex patterns in high-dimensional contexts. Their amenability to alleviate preliminary feature engineering together with relax linearity, normality or stationarity assumptions, gives them strong assets for time series modeling. Notably, multivariate forecasting from vast amounts of related time series is an area where deep learning is proving to be effective. Such advantages should promise deep neural networks a bright future in temporal series modeling. Specialization in deep neural models is not the purpose of this survey. The recent survey of \citet{benidis_rangapuram_flunkert_2020_arxiv_survey-time-series-deep_learn-forecast} lists directions for further developments in the field of deep learning.

The historical preeminence of recurrent networks in time series modeling is now outdated. Thanks to the introduction of architectural elements such as dilated convolution, convolutional networks relegate {\sc rnn}s at a lower rank as {\sc cnn}s now offer powerful frameworks to model time series. The potential of {\sc cnn}s in this domain needs to be pushed to further boundaries. The design of new neural architectures and the development of new tricks adapted to time series domain will eventually benefit other fields.

However, deep neural networks are not a panacea in this domain. Deep learning methods need a very large amount of data; they require programming with Graphics Processing Units ({\sc gpu}s) and are time-consuming. Although {\sc gpu} programming is gaining popularity, harnessing its capabilities is far from trivial.

In contrast, established models such as the {\sc arima} are not only popular because of their high accuracy; their ease of use and robustness make them particularly suitable for non-expert users. Besides, deep neural networks suffer from a well-known lack of interpretability. Post-hoc  interpretable  models  were proposed  to help  identify  important  features  or  examples, but these models ignore any sequential dependencies between inputs. Alternatively, using strategically-placed attention layers is expected to shed light on the relative importance of features at each training time step by analyzing the attention weights. Further, attention mechanisms can be used to identify persistent temporal relationships ({\it e.g.}, seasonal patterns) in the data. However, such work is still in its infancy.

Many practical applications deal with datasets much smaller in size than the size required to train a deep neural network, even when using a pre-trained network. Time series modeling has yet to benefit from research outside the deep learning domain.

By nature, only approaches outside the deep learning field can be suitable when data are missing, and when observations arrive at random time steps. Besides, these approaches benefit from solid theoretical foundations. Therefore, any attempts to improve their forecast accuracy still remain a worthwhile endeavour. Notably, it is expected that {\bf combining such conventional methods with more general techniques of the machine learning domain} can bring promising advances in time series forecasting.

Finally, recent proposals in time series modeling and forecasting rely on the combination of deep neural models and conventional models, to find a balance between the effectivenesses of forecasting and inference tasks in presence of complex data (nonlinear, nonstationary) \citep[{\it e.g.},][]{rangapuram_seeger_gasthaus_et_al_2018_neurips_deep-learning-probab-forecasting-multivar-time-series,alaa_van-der-schaar_2019_neurIPS_combin-deep-learn-state-space,yanchenko_mukherjee_2020_arxiv_combin-state-space-deepl-learn-foreca-infer}. Such approaches will no doubt continue to receive research attention in the future.

\subsection{Multivariate Framework}

Recently, large and diverse time series datasets have been made available in a wide range of fields such as energy consumption of house-holds, demand for all products that an online shopping portal offers, server load in a data center, traffic flow, to name but a few. This increasing availability opens up new possibilities for multivariate time series forecasting, with various objectives
in mind.

Leveraging dependencies across individual time series, that is across as many variables, is expected to increase the prediction accuracy for each such variable. Besides the perpetual quest to improve forecasting accuracy, estimating forecasting uncertainty has motivated research work for decades, as such estimation may be critical for downstream decision making. Most conventional forecasting approaches have focused on predicting point estimates,
such as the mean or the median of the distribution at a future time point. Recently, the wealth of data made available in many domains makes it possible to shift the paradigm from univariate to multivariate probabilistic forecasting. 

More and more work dedicated to multivariate time series has emerged. Future directions of research will naturally fall in two categories - point forecasting in multivariate framework and probabilistic forecasting - while having to cope with high dimensionality.
A comprehensive review of the models and methods proposed to tackle multivariate time series forecasting was not the purpose of the present survey, which aims to be more general. Nonetheless, Section \ref{deep_learning} of our survey devoted to deep learning lifts the veil on time series forecasting in the context of big data. To note, research in the domain of multivariate forecasting will undoubtedly continue to explore variants around vector autoregressive models, for instance by constraining coefficients through various means such as sparsity, low-rank coefficient matrices \citep{alquier_bertin_doukhan_et_al_2020_stat-comput_high_dimension_var_low_rank_transit}. Another line of investigation is to combine low-rank matrix factorization with regularization, as pioneered by \citet{yu_rao_dhillon_2016_neurips_high-dim-time-series-matrix-factoriz-tempor_regul}. Developing hybrid models is a promising track, and a number of avenues can be explored in the line of recent proposals: combination with a classical linear autoregressive model \citep[{\it e.g.},][]{lai_chang_yang_et_al_2018_sigir_multivar-forecast-time-series-cnn-rnn-lar}, hybridization with matrix factorization \citep[{\it e.g.},][]{sen_yu_dhillon_2019_neurips_deepglo-time-series-forecast-high-dim-cnn}, modeling of each individual time series {\it via} a global component (a deep neural network) together with a local probabilistic model ({\it e.g.}, \citealp{rangapuram_seeger_gasthaus_et_al_2018_neurips_deep-learning-probab-forecasting-multivar-time-series}, \citealp{wang_smola_maddix_et_al_2019_icml_deep-factors-probab-forecasting-multivar-time-series}). The recently-proposed normalizing flow technique opens up avenues of research including its combination with an autoregressive model such as a recurrent neural network \citep{rasul_sheikh_schuster_et_al_2021_iclr_prob-conditioned-normalizing-flows-time-series-forecast}. Besides, deep ensembles ({\it i.e.}, ensembles of neural networks) seem an appealing lead to enhance predictive performance. Various techniques may be envisaged, embracing parameter sharing across ensemble networks, snapshot ensembling \citep{huang_li_pleiss_et_al_2017_iclr_snapshot-ensemb-deep-learn} and dropout \citep{srivastava_hinton_krizhevsky_2014_jmlr_dropout-neur-netw-overfit}.

\subsection{General Machine Learning Techniques to Enhance Conventional Methods}

Time series conventional modeling has to draw advantages from the classical machine learning toolkit. In return, the adaptation of these tools to a temporal data framework may motivate advanced investigations in some cases.

\subsubsection{Ensemble-based Strategies}

Ensemble-based strategies are renowned for enhancing accuracy in prediction and tackling high dimensionality. For instance, after more than 50 years of widespread use, exponential smoothing remains one of the most relevant forecasting methods in practice. The reasons for this are its simplicity and transparency, together with its amenity to handle many situations. The principle of combining bagging with exponential smoothing methods was only invastigated recently to improve point forecasts \citep{bergmeir_hyndman_benitez_2016_int-j-of-forecast_bagging-exponen-smooth-methods}. Besides, such a framework is clearly appealing to handle probabilistic forecasting and further investigations in this line are worthwhile.  


A renewed interest is shown in gradient boosting regression trees \citep{elsayed_thyssens_rashed_2021_arxiv_grad-boost-regr-trees-time-series}, and random forests \citep{goehry_yan_goude_et_al_2021_hal_rand-forest-time-series-forecast}, to forecast  time  series, with challenging issues. For instance, standard random forests cannot cope with time-dependent structure. In random forest variants focused on time series, the principle is to apply a block bootstrap, to subsample time series to take time dependence into account. Consistently, as standard permutation does not preserve the dependence structure, further work is required to extend the concept of variable importance to block variable importance. 

Complete subset regression was developed to combine forecasts \citep{elliott_gargano_timmermann_2015_j-of-econ-dyn-and-cont_compl-subset-regr-larg-dim}:  instead of selecting an optimal subset of predictors, which is infeasible in most cases, the principle is to combine predictions by averaging all possible (linear) regression models with the same fixed number of predictors. As the computational cost may nonetheless remain prohibitive, combining complete subset regression with bagging seems an interesting avenue to be further explored \citep{medeiros_vasconcelos_veiga_2021_j-of-busi-and-eco-stat_forecasting-infla}. 

Enrichment of the pool of simple baseline forecasting methods has to benefit from machine learning techniques that allow to optimally combine a large set of forecasts. More generally, as an alternative to penalized regression in baseline models, ensemble of models should be further considered to tackle high dimensionality. 

\subsubsection{Penalized regression}

Recent works \citep[{\it e.g.},][]{bessac_ailliot_cattiaux_et_al_2016_advanc-stat-climat-met-ocean_hidd-obs-regime-switch-ar-model} put forward the necessity to investigate reduced parameterizations of the autoregressive coefficients and of the matrices of covariance of residuals, to handle larger datasets. A large body of existing work in time series model-learning relies on classical penalization techniques. Beyond Ridge Regression, Lasso, Elastic Net, adaptive versions of the Lasso as well as Elastic Net and the group Lasso \citep{kock_medeiros_vasconcelos_2020_chapt_penalized-time-series-regress}, there is still much to explore to cope with high dimensionality.

\subsubsection{Clustering}

Clustering has been called upon various fields to discover hidden models underlying the data, and time series modeling is no exception to this trend. 

In Observed Regime-Switching AutoRegressive ({\sc orsar}) models, the state process is either observed or derived {\it a priori}. In the latter case, a clustering phase aims at extracting the regimes, before fitting the model. The clustering task may either rely on endogenous variables ({\it i.e.}, the variables whose dynamics is observed through the time series) or on exogenous variables assumed to drive regime-switching. Previous investigations confined to {\sc orsar} models \citep{bessac_ailliot_cattiaux_et_al_2016_advanc-stat-climat-met-ocean_hidd-obs-regime-switch-ar-model} should extend to the semi-latent Regime-Switching AutoRegressive model developed by  \citet{dama-sinoquet_2021_ictai_phmc-lar-mach-health-diagn}. These extensions would benefit from cutting-edge research in time series clustering \citep[{\it e.g.},][]{kanaan_benabdeslem_kheddouci_2020_ictai_clust-ensembl-mixt-hmms}.

Building a model for each cluster of related time series is a lead that has not been sufficiently explored. Improvement of forecast accuracy, together with reduced training cost are expected. The challenge remains to subsequently design a general model as fuzzy clusters may exist, and the time series subject to prediction may not fit well to any of the clusters identified.

\subsection{Beyond Time Series: Bivariate Processes}
 
Recent investigations around Markov-Switching AutoRegressive models have put the spotlight on these models which add a state process to time series modeling. Therein, different time scales present in the data may be taken into account: while states account for long-term evolution, the autoregressive part describes short-term fluctuations. In the state process, transitions from one state to another drive the global nonlinear dynamics of the system; local state-specific dynamics come into play to generate the time series under consideration.

As a first line of investigation, some real-world situations may require a refined model in which an additional layer would be added to simulate shorter time scales for very local features.

Besides, these models have been elaborated with Hidden Markov Model ({\sc hmm}) variants as the backbone for the state process.  The standard {\sc hmm} involves a bivariate process $\{S_t, X_t\}, t = 1, \cdots, T$, composed of observed random variables (outputs) $\{X_t\}$ and
of latent discrete random variables (latent states) $\{S_t\}$. However, in a number of situations, partial knowledge about the states ${S_t}$ is available, and a partially Hidden
Markov Model would better describe the dynamic process under analysis. Partial knowledge may be understood as uncertain knowledge on states \citep{juesas-ramasso-drujont_2021_arxiv_msa-hmm-part-knowl}, random mixture of latent and observed states \citep{dama-sinoquet_2021_arxiv_phmc-lar}, or even having the latent states depend on an observed independent Markov chain \citep{monaco_tappert_2018_pattern-recognition_partial-observ-hmm-keystroke-dynamics}. It remains to develop other modalities to best describe real-world situations.

Modeling time series comes up against the inescapable issue to capture short-term dependencies as well as long-term dependencies. So far, no work was reported in the literature to account for the long-term dependencies when (i) the state process underlying the dynamics of a system is governed by an event trace, (ii) some (if not all) of the events determine state-switching, (iii) dependencies exist between the events. An illustration of such correlated events is for instance the situation in which an action exerts a delayed effect: the administration of a treatment to a patient will only change their condition ({\it i.e.}, state), together with some monitored physiologic parameters ({\it e.g.}, blood pressure, heart rate) after a certain period of time. Examining correlation in time series is done routinely. Capturing dependencies between events is much more complex and resorts to pattern matching \citep[{\it e.g.},][]{ceci_lanotte_fumarola_2014_ic-disco-sci_seq-patt-mining-processes}, interval algebra \citep[{\it e.g.},][]{senderovich_weidlich_gal_2017_ic-on-bus-proc-man_tempo-netw-rep-event-logs}, causal inference \citep[{\it e.g.},][]{kobayashi_otomo_fukuda_2018_ieee-trans-netwk-serv-mana_mining-causa-netwk-events-time-series}, marked point process modeling using piecewise-
constant conditional intensity functions \citep[{\it e.g.},][]{parikh_gunawardana_meek_2012_uai-wks-temp-dep-event-streams, islam_shelton_casse_et_al_2017_conf-ml-health-temp-dep-event-streams}. deep neural network modeling \citep[{\it e.g.},][]{du_dai_trivedi_2018_kdd_event-traces-deep-learn}. Such works around the modeling of temporal dependencies in event streams would be likely to inspire avenues for advanced dynamic bivariate process modeling. Finally, few works have dealt with the identification of correlations between event traces and time series \citep{minaei-bidgoli_lajevardi_2008_ic-on-netw-comp-and-adv-infor-man_corr-mining-time-ser-event-logs,luo_lou_lin_et_al_2014_sigkdd_corr-events-time-series-events}. Recently, \citet{xiao_yan_yang_et_al_2017_aaai_event-trac-model-deep-learn} have designed a deep network for point process modeling, where they handle two types of recurrent neural networks ({\sc rnn}s), one with its units aligned with time series time steps, and one with its units aligned with asynchronous events to capture the long-range dynamics. Thus, time series and event sequences can be synergically modeled. Whereas the model of \citet{xiao_yan_yang_et_al_2017_aaai_event-trac-model-deep-learn} is designed for event prediction, it could inspire a novel framework dedicated to time series forecasting enhanced by trace event modeling.  

Finally, some of the events may be actions triggered by a human beeing ({\it e.g.}, administration of a drug to a patient). The design of novel {\it generative} models semi-guided by such actions while also governed by random events is highly challenging. This avenue of research has never been explored to our knowledge. 

\subsection{Dealing with Data Obsolescence}

If part of the data to train a model is obsolete, it should not be considered. Answering this question is quite tricky as it requires understanding of how the system under analysis changes, and where the causes of nonstationarity lie. For example, if a business has significantly grown since last year, the data of the same quarter of the previous year should be considered obsolete. 
%


 
\section{Glossary}
\noindent\hspace{5,5mm}$\bullet$ $\bm{X \sim f:}$ Random variable $X$ is driven by probability law $f$.

$\bullet$ {\bf autocorrelation:} Autocorrelation is the correlation between a time series variable and its lagged version in time.

$\bullet$ {\bf autocovariance:} Autocovariance is the covariance between a time series variable and its lagged version in time.

$\bullet$ {\bf characteristic equation:} A characteristic equation is a $p$-degree algebraic equation that can be associated with the $p$-order linear differential equation describing the behavior of some dynamic system of interest. The linear differential equation and characteristic equation share the same coefficients and the variables' degrees in the latter are equal to the successive differentiation orders. It is known that the solutions of a characteristic equation provide important information about the behavior of the system under analysis. For instance, when dealing with a time-differential equation, the system is said \textbf{stable} (equivalently, \textbf{stationary}) if and only if the modulus of each (complex) root of the characteristic equation is strictly less than $1$.

$\bullet$ {\bf covariance:} The covariance of two real random variables $X$ and $Y$ with expected values $\mathbb{E}[X]$ and $\mathbb{E}[Y]$ is defined as: $cov(X,Y) = \mathbb{E}[(X-\mathbb{E}[X])(Y-\mathbb{E}[Y])].$ 

$\bullet$ {\bf correlation:} The correlation coefficient of two real random variables $X$ and $Y$ with expected values $\mathbb{E}[X]$ and $\mathbb{E}[Y]$, and standard deviations $\sigma_X$ and $\sigma_Y$, is defined as: $\rho_{X,Y} = \frac{cov(X,Y)}{\sigma_X\ \sigma_Y} = \frac{\mathbb{E}[(X-\mathbb{E}[X])(Y-\mathbb{E}[Y])]}{\sigma_X\ \sigma_Y}.$ 

$\bullet$ {\bf \textsc{cusum} test:} A \textsc{cusum} test uses the cumulative sum of some quantity to investigate whether a sequence of values has drifted from some model. In time series analysis, a standard \textsc{cusum} statistics uses the sequence of residual deviations from a given model.

$\bullet$ \textbf{design matrix:} In a linear regression model, a design matrix is a data matrix in which columns are variables, lines represent data instances and a column of ones is added (at the begging or at the end), in order to take into account the intercept. This matrix allows a matricial formulation of the model. For instance, the matricial formulation of the {\sc ar}($p=2$) model learned on time series $\{z_1, z_2, \dots, z_T\}$ is the following:
$$ \mathbf{Z} = \mathbf{D} \, \Phi \, + \mathbf{E} ,$$

where 
$$ 
	\mathbf{Z} = \begin{pmatrix}
  				 z_3 \\
 				 z_4 \\
  				 \vdots \\
  				 z_{T-1} \\
  				 z_T
				\end{pmatrix}, \quad
	\mathbf{D} = \begin{pmatrix}
  					1      & z_2     & z_1 \\
  					1      & z_3     & z_2 \\
  					\vdots & \vdots  & \vdots \\
  					1      & z_{T-2} & z_{T-3} \\ 
  					1      & z_{T-1} & z_{T-2}
 				\end{pmatrix}, \quad 
 		\Phi  = \begin{pmatrix}
			 		\phi_0 \\
			 		\phi_1 \\
			 		\phi_2
				\end{pmatrix}, \quad
   \mathbf{E} = \begin{pmatrix} 
					\epsilon_3 \\
  					\epsilon_4 \\
  					\vdots \\
 					\epsilon_{T-1} \\
  					\epsilon_{T}		 
  				\end{pmatrix},
$$
and $\mathbf{D}$ is the $(T - p) \times (p + 1)$ design matrix  associated with the model.

$\bullet$ {\bf i.i.d.:} independent and identically distributed

$\bullet$ {\bf i.i.d.$\bm{(0,1)}$:} The i.i.d. random variables concerned have the same probability distribution, characterized by null mean and unit variance.

$\bullet$ {\bf heteroscedasticity:} Heteroscedasticity refers to the condition in which the variance of a stochastic process is unstable.

$\bullet$ {\bf homoscedasticity:} Homoscedasticity refers to the condition in which the variance of a stochastic process is constant over time.

$\bullet$ {\bf lag k-correlation:} Lag k-correlation is the correlation between a time series variable and its $k^{th}$ lagged version in time.

$\bullet$ {\bf level of a time series ($\bm{\tilde{X}_t$}):} The level is the value of the series at time step $t$, from which the random variations have been removed

$\bullet$ {\bf marked temporal point process:} Marked temporal point processes are a mathematical framework dedicated to the modeling of event data with covariates. 

$\bullet$ {\bf Markov chain:} A Markov chain is a discrete-time discrete-state stochastic process $\{S_t\}$ $t=1, \cdots,T$ satisfying the Markov rule: $S_t$ only depends on the preceding state, $S_{t-1}$ , given all anterior states $S_{t-2}, S_{t-3}, \cdots$.


$\bullet$ {\bf model identification and model learning:} Model identification consists in learning model hyperparameters ({\it e.g.}, number of lagged values in an autoregressive model). Relying on some criterion such as the {\sc bic} or the {\sc aic} criterion, criterion-based methods seek for a trade-off between model complexity and good fitting to training data. These widely-used methods are known to be easily applicable and to have the ability to select relevant models. It is also usual to select the model that obtains the best prediction accuracy on a test dataset. \\
Parameter estimation, also called model learning, aims at estimating the remaining parameters of the model from observed data ({\it e.g.}, the intercept, regression coefficients and standard deviation, in a linear autoregressive model). Maximum Likelihood Estimation, and Empirical Risk Minimization (whose the Ordinary Least Square method is the most known instance), are extensively used for parameter estimation, depending on the problem. 

$\bullet$ {\bf residuals:} In a time series model, the residuals are defined as what is left over after fitting the model to the data observed. When a model fits the observed data well, residuals are expected to be random variations, also called noises.

$\bullet$ {\bf unit circle:} This term refers to the circle of unit radius centered at the origin $(0,0)$, in the Cartesian coordinate system. Points located on the unit circle have a modulus equal to $1$, those situated inside (resp. outside) the circle have a modulus less (resp. greater) than $1$.
 
$\bullet$ {\bf unit roots:} In an equation expressed as an algebraic formula, solutions with a unit modulus are called unit roots ({\it i.e.}, a unit root in the field of complex numbers has its modulus equal to 1; a unit root in the field of reals has its absolute value equal to 1) .

$\bullet$ {\bf variance:} The variance of real random variable $X$ with expected value $\mathbb{E}[X]$ is defined as:  $\mathbb{V}[X]=\mathbb{E} \left[(X-\mathbb{E}[X])^{2}\right].$ 

$\bullet$ {\bf white noise process:} White noise process is defined as the standard normal distribution, that is the Gaussian distribution with zero mean and unit variance.

\bibliographystyle{spbasic}      
\bibliography{references_21_09_25_sat}   

\end{document}